\documentclass[acmlarge]{acmart}

\usepackage[capitalise]{cleveref}
\usepackage{graphicx} % Required for inserting images
\usepackage{multirow}
\usepackage{svg}
\usepackage{subcaption}     % subfigure
\usepackage{tabularx,booktabs,array}
\usepackage{longtable}
\usepackage{pdflscape}
\usepackage{threeparttable}
\newcolumntype{C}{>{\centering\arraybackslash}X} % centered, auto-stretch column
\usepackage{algorithm}
\usepackage{algpseudocode}
\usepackage{amsmath}
\usepackage{booktabs}
\newcolumntype{Y}{>{\raggedright\arraybackslash}X}
\usepackage[table]{xcolor}
\newcommand{\best}[1]{\cellcolor{green!25}#1}
\newcommand{\second}[1]{\cellcolor{yellow!25}#1}
\usepackage{listings}

\AtBeginDocument{%
  }

\newboolean{highlightversion}
\setboolean{highlightversion}{false}  % Set to false for clean version, true for highlighted

\newcommand{\revise}[1]{%
    \ifthenelse{\boolean{highlightversion}}%
    {\textcolor{blue}{#1}}  % Change the text color to red if highlighting is enabled
    {#1}                   % No highlighting if clean version
}

\begin{document}

%%
%% The "title" command has an optional parameter,
%% allowing the author to define a "short title" to be used in page headers.
\title{VSMP-IMU: Video-Grounded Semantic Motion Programs for Sensor-Aware Synthetic IMU Generation}

\author{Lala Shakti Swarup Ray}
\email{lala_shakti_swarup.ray@dfki.de}
% \orcid{1234-5678-9012}
% \authornotemark[1]
\affiliation{%
  \institution{DFKI}
  \city{Kaiserslautern}
  \country{Germany}
}

\author{Vitor Fortes Rey}
\affiliation{%
  \institution{DFKI and RPTU Kaiserslautern-Landau}
  \city{Kaiserslautern}
  \country{Germany}
}
\email{vitor.fortes_rey@dfki.de}

\author{Mengxi Liu}
\affiliation{%
  \institution{DFKI and RPTU Kaiserslautern-Landau}
  \city{Kaiserslautern}
  \country{Germany}
}
\email{mengxi.liu@dfki.de}

\author{Paul Lukowicz}
\affiliation{%
  \institution{DFKI and RPTU Kaiserslautern-Landau}
  \city{Kaiserslautern}
  \country{Germany}
}
\email{paul.lukowicz@dfki.de}

\author{Bo Zhou}
\affiliation{%
  \institution{DFKI and RPTU Kaiserslautern-Landau}
  \city{Kaiserslautern}
  \country{Germany}
}
\email{bo.zhou@dfki.de}
%%
%% The "author" command and its associated commands are used to define
%% the authors and their affiliations.
%% Of note is the shared affiliation of the first two authors, and the
%% "authornote" and "authornotemark" commands
%% used to denote shared contribution to the research.

%%
%% By default, the full list of authors will be used in the page
%% headers. Often, this list is too long, and will overlap
%% other information printed in the page headers. This command allows
%% the author to define a more concise list
%% of authors' names for this purpose.
\renewcommand{\shortauthors}{Kim et al.}

\begin{teaserfigure}
    \centering
    \includegraphics[width=\textwidth]{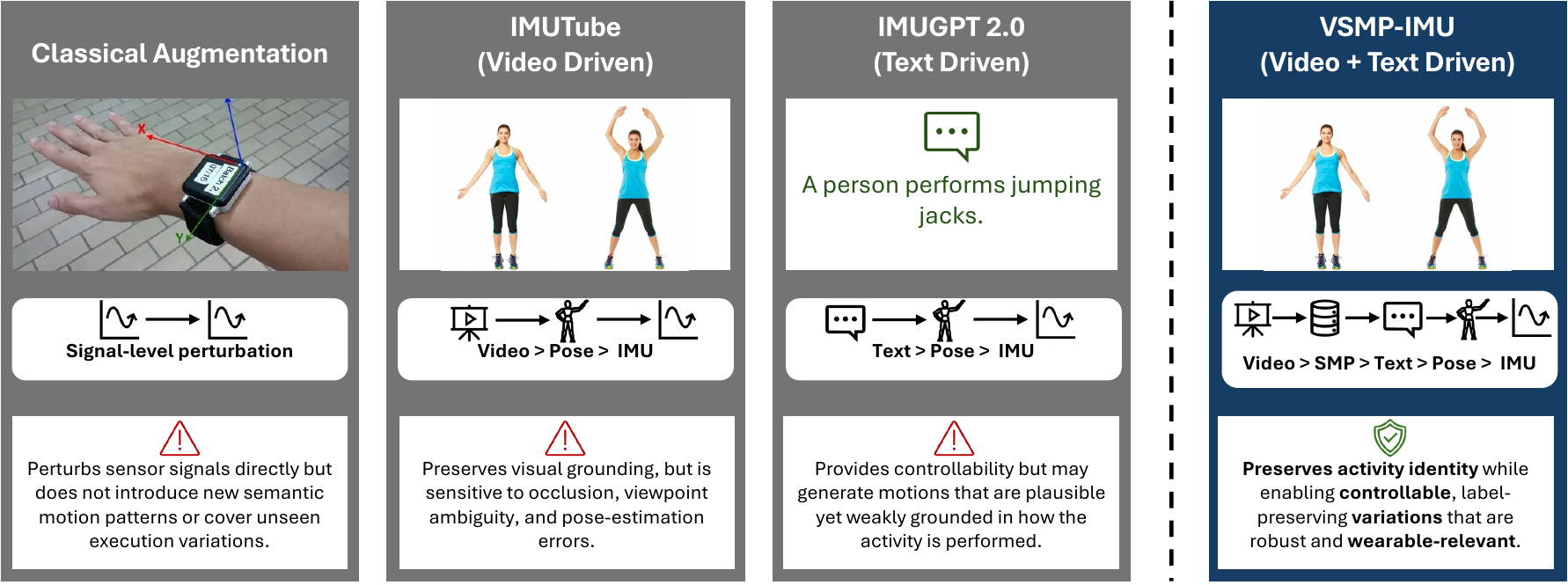}
    \caption{Compared to alternative synthetic IMU data generation methods, VSMP-IMU better preserves activity-defining invariants while controlling label-retention execution and sensing variations through Semantic Motion Programs (SMPs) as a structured intermediate representation.} %Classical augmentation perturbs existing IMU signals but does not introduce new semantic motion structure. Video-driven synthetic IMU methods such as IMUTube are grounded in observed videos but remain sensitive to pose-estimation errors and viewpoint ambiguity. Text-driven methods such as IMUGPT~2.0 provide greater controllability but are weakly grounded in how the activity is actually performed.
    \label{fig:teaser}
\end{teaserfigure}
%%
%% The abstract is a short summary of the work to be presented in the
%% article.
\begin{abstract}
Wearable human activity recognition (HAR) is often limited by the scarcity of labeled sensor data, especially in low-resource, class-imbalanced, and subject-generalization settings. 
Synthetic IMU generation can reduce this dependency and enhance HAR machine learning model's performance, but existing approaches face a trade-off without addressing all factors: video-driven methods are visually grounded but sensitive to pose-estimation errors, while text-driven methods are controllable but often weakly grounded in how activities are actually performed.
We present VSMP-IMU, a video-grounded framework for controllable synthetic IMU generation based on a structured Semantic Motion Program (SMP), which separates activity-defining semantics from label-preserving variation.
Given an input video, VSMP-IMU extracts and augments an SMP, uses it to synthesize motion, converts the motion into virtual IMU signals, and grounds the resulting signals to the target wearable domain.
We evaluate VSMP-IMU against state-of-the-art synthetic data generation methods on five public IMU-HAR datasets under leave-one-person-out evaluation.
VSMP-IMU achieves an average Macro-F1 of 78.33\%, improving over real-only training by 9.77\% and over the strongest prior synthetic baseline by 4.04\%. 
\revise{A separate end-to-end controllability study on 12 UTD-MHAD activities shows strong motion-level control but attribute-dependent attenuation after sensor grounding. Requested tempo correlates with generated motion at $\rho=0.92$ and with grounded IMU at $\rho=0.61$,
while $90.6\%$ of generated motions reproduce the requested repetition count within one repetition. Amplitude remains comparatively well preserved after grounding ($\rho=0.75$), whereas primitive-duration control exhibits greater sensor-level error.}
In low-resource settings with reduced training data-samples, it improves over real-only training by 18.54\% and over the strongest prior synthetic baselines by more than 6\% on average. Under long-tail evaluation in imbalanced datasets, it improves tail-class Macro-F1 by 19.86\% over Real-only training and by 4.76\% over SOTA. These results show that structured video-grounded semantics provide a practical foundation for controllable, wearable-relevant synthetic sensor data generation.
\end{abstract}

%%
%% The code below is generated by the tool at http://dl.acm.org/ccs.cfm.
%% Please copy and paste the code instead of the example below.
%%


%%
%% Keywords. The author(s) should pick words that accurately describe
%% the work being presented. Separate the keywords with commas.

%%
%% This command processes the author and affiliation and title
%% information and builds the first part of the formatted document.
\maketitle

\section{Introduction}

Wearable human activity recognition (HAR) is a major field of study in ubiquitous computing, supporting applications such as fitness tracking, rehabilitation, elder care, workplace monitoring, and context-aware assistance~\cite{haresamudram2025past}. Despite progress in deep learning, practical HAR systems still depend heavily on labeled inertial measurement unit (IMU) data~\cite{kwon2020imutube,leng2024imugpt}. This dependence on labeled wearable data remains a major barrier to scalable deployment beyond controlled benchmarks, especially in low-resource, class-imbalanced, and subject-generalization settings~\cite{zhou2024autoaughar,hong2024crosshar}.

The challenge is not only that labeled IMU data is scarce, but also that real-world variability is difficult to capture. Wearable data collection requires repeated instrumentation, careful protocols, and reliable annotation across many users and sessions~\cite{kwon2020imutube}. Activity execution varies across people, sensor placement and orientation differ across wearers, and rare activities are difficult to collect at scale ~\cite{su2022learning,kang2022augmented,hong2024crosshar}. As a result, training sets often underrepresent the variations that matter most when HAR models are evaluated on unseen users or deployed in new settings.

Synthetic IMU generation methods have emerged to reduce this data bottleneck and enhance model performance through data augmentation to complement real sensor data. 
Existing approaches ~\cite{kwon2020imutube,kwon2021approaching,santhalingam2023synthetic}, however, only partially address this goal. Classical signal-level augmentation methods such as jittering, scaling, rotation, and time warping can improve robustness, but they operate directly on existing sensor recordings. They do not introduce new semantic executions or physically plausible variations of an activity label.
Video-driven methods estimate human motion from video and convert the recovered trajectories into simulated IMU signals~\cite{santhalingam2023synthetic}. These methods benefit from visual grounding but remain sensitive to pose-estimation errors, occlusion, viewpoint ambiguity, and motion-recovery noise. Text-driven methods generate motion from activity descriptions before converting it into IMU signals~\cite{leng2023generating,leng2024imugpt}. They offer greater controllability but are often weakly coupled to dataset-specific spatial-temporal motion patterns~\cite{leng2024imugpt,haresamudram2025limitations}. Neither paradigm explicitly separates \emph{activity identity} from \emph{execution variation}, although wearable HAR requires label-preserving variability in both motion style and sensor realization.

We argue that synthetic IMU generation needs a structured intermediate representation that preserves the semantic concept of an activity while exposing controllable sources of variation~\cite{su2022learning,leng2024imugpt}. 
%Rather than directly transferring noisy frame-level pose from video or relying on generic unconstrained free-form text where the whole motion description is generated from a single label name, such a representation should describe both what motion is being performed, how it is organized over time, which body parts are involved, and which execution attributes can be modified to enrich the data variations of the same activity label.
Rather than transferring noisy frame-level pose estimates from video, or generating motion from unconstrained text prompts derived only from an activity label, synthetic IMU generation needs a more structured representation. Such a representation should capture what motion is being performed, how the motion unfolds over time, which body parts are involved, and which execution attributes can be safely varied while preserving the activity label.

Motivated by this observation, we propose \textbf{VSMP-IMU}, a framework for \textbf{V}ideo-grounded \textbf{S}emantic \textbf{M}otion \textbf{P}rograms for synthetic \textbf{IMU} generation. 
VSMP-IMU introduces  a structured intermediate representation \emph{Semantic Motion Program} (SMP) between video understanding and inertial synthesis.
An SMP describes an activity in terms of its motion primitives, body-part involvement, temporal organization, execution attributes, contact cues, augmentation constraints, and uncertainty. It preserves activity-defining structure extracted from video while exposing controllable, label-preserving sources of variation.
Given an activity video, VSMP-IMU extracts and refines an SMP, samples constrained variants, synthesizes 3D motion, converts the motion into virtual IMU signals, and grounds the signals to the target wearable domain. 
In this way, video provides evidence of how the activity is performed, the SMP provides control over valid execution variations, and sensor-aware grounding adapts the generated signals to the target IMU dataset.

The main contributions of this paper are:
\begin{itemize}
    \item \revise{We introduce Semantic Motion Programs as a structured intermediate representation that separates activity-defining invariants from label-preserving execution attributes. We further quantitatively evaluate whether requested changes in tempo, amplitude, repetition count, primitive duration, and symmetry are realized in generated motion and remain observable after virtual IMU synthesis and target-domain grounding.}
    \item We propose VSMP-IMU, a video-grounded synthetic IMU generation framework for enhancing HAR models. VSMP-IMU extracts SMPs from activity videos, augments them under semantic constraints, synthesizes 3D motion, and converts motion into virtual IMU signals. 
    \item We evaluate VSMP-IMU across five public HAR datasets under leave-one-person-out, low-resource, and long-tail with many underrepresented classes on HAR settings on paired video--IMU datasets and IMU-only HAR benchmarks~\cite{zhou2024autoaughar,hong2024crosshar}, demonstrating consistent improvements over real-only training, classical signal augmentation, video-driven generation, and text-driven generation. 

\end{itemize}

\section{Related Work}

\subsection{Synthetic IMU generation from video}

Synthetic or virtual inertial data generation has been explored as a way to reduce the need for costly labeled IMU collection in wearable HAR. A common approach estimates human motion from video and converts the recovered kinematics into inertial measurements. Early work showed that monocular RGB videos can provide training data for motion-sensor-based activity recognition~\cite{rey2019let}, and later work used residual convolutional models to translate videos into synthetic accelerometer and gyroscope signals~\cite{fortes2021translating}.

\textit{IMUTube} introduced an automated pipeline for extracting virtual on-body accelerometry from activity videos~\cite{kwon2020imutube}. Subsequent work used such virtual IMU data for more realistic HAR settings, including free-weight exercises, weakly labeled video sources, and larger HAR models trained with limited real calibration data~\cite{kwon2021approaching,kwon2021complex}. More recent systems extend video-to-IMU generation to broader sensing contexts: \textit{Video2IMU} generates realistic IMU signals and features from monocular videos~\cite{lamsa2022video2imu}, \textit{Vi2IMU} targets smartwatch-based ASL sensing~\cite{santhalingam2023synthetic}, and \textit{SignRing} uses online sign-language videos to synthesize IMU data for ring-based ASL recognition~\cite{li2023signring}.

Despite their scalability, video-driven pipelines remain sensitive to pose estimation, motion reconstruction, and sensor simulation errors. Occlusion, viewpoint ambiguity, depth uncertainty, body-shape variation, and pose noise can propagate into synthesized inertial signals, especially for subtle or fine-grained activities~\cite{jain2022effectiveness,leng2023utility}. Moreover, direct video-to-IMU generation is often tied to the observed trajectory and provides limited control over label-preserving variations such as tempo, amplitude, body-part emphasis, sensor placement, and sensing conditions.

\subsection{Language-driven inertial synthesis}

Another line of work uses language as the source modality for motion or inertial synthesis. \textit{IMUGPT} combined large language models with text-to-motion models to generate virtual accelerometer data from textual activity descriptions~\cite{leng2023generating}. \textit{IMUGPT 2.0} scaled this idea and introduced motion filtering and diversity-aware generation control~\cite{leng2024imugpt}. \textit{Text2IMU} further explored text-driven synthesis by generating acceleration and gyroscope signals from textual prompts, human surface models, and multiple virtual IMU placements~\cite{haeusler2025text2imu}.

Language-driven methods provide greater controllability than direct video-to-pose transfer and can generate diverse activity descriptions without source videos. However, their prompts are often coarse or weakly grounded in how activities are actually performed. As a result, generated motions may match the activity label while missing important wearable-relevant details such as temporal organization, body-part involvement, object interaction, repetition structure, or execution style. Recent comparisons suggest that video-based and text-based virtual IMU pipelines are complementary: video preserves concrete motion evidence, while text contributes semantic and contextual diversity~\cite{leng2025scaling}.

VSMP-IMU uses HY-Motion \cite{wen2025hy} as the text-conditioned 3D human-motion generator. HY-Motion maps natural-language motion descriptions to temporally ordered body-motion sequences, which can then be converted into virtual IMU measurements. We use HY-Motion as an off-the-shelf motion synthesis backend rather than as a contribution of this work.

\subsection{Data augmentation and domain generalization for wearable HAR}

Wearable HAR has also relied on signal-domain augmentation to improve robustness under limited labeled data. Common transformations include jittering, scaling, rotation, cropping, permutation, magnitude warping, and time warping. Prior studies show that such augmentations can improve recognition under small-data and noisy-label conditions~\cite{um2017data,jeong2021sensor}, while more recent work explores automated and generative augmentation strategies~\cite{wang2024data,zhou2024autoaughar}.

These methods are simple and effective, but they operate mainly at the signal level and do not explicitly model which motion properties define the activity class and which can vary across subjects, executions, or sensor setups. This distinction is important in wearable HAR, where subject identity, morphology, placement, orientation, and execution style can change the IMU signal without changing the activity label.

Domain adaptation and generalization methods address related sources of variation, including subject shift, device differences, sensor-placement changes, and dataset heterogeneity. Examples include adaptive spatial-temporal transfer learning~\cite{qin2019cross}, semantic-discriminative mixup~\cite{lu2022semantic}, augmented adversarial learning for partial sensor sets~\cite{kang2022augmented}, and cross-dataset or foundation-style HAR models based on self-supervised or language-guided pretraining~\cite{hong2024crosshar,miao2024goat}. Self-supervised methods such as contrastive predictive coding further reduce label dependence by learning temporal structure from unlabeled sensor streams~\cite{haresamudram2021contrastive,haresamudram2022assessing}. However, these approaches generally do not provide a structured semantic interface for controllable synthetic IMU generation.

\subsection{Structured intermediate representations for motion generation}

Recent work in video understanding and motion generation has shown the value of structured intermediate representations for human actions. Fine-grained datasets such as \textit{FineGym} represent activities hierarchically as events, elements, and temporally ordered sub-actions~\cite{shao2020finegym}. Motion-primitive approaches similarly model actions as compositions of reusable units, including contact-centered therbligs~\cite{dessalene2023therbligs} and neuro-symbolic motion programs~\cite{kulal2021hierarchical}.
Program-like structures have also been used for reasoning and generation. Modular motion programs support interpretable motion question answering~\cite{endo2023motion}, \textit{LEAP} generates egocentric action programs from video using large language models~\cite{dessalene2023leap}, and scene-affordance representations bridge language grounding and scene-conditioned motion synthesis~\cite{wang2024move}. These works suggest that structured representations can make motion generation more interpretable, compositional, and controllable than raw pose trajectories or free-form text alone.

This idea is particularly relevant for wearable HAR: variations in tempo, amplitude, repetition count, dominant side, body-part involvement, and object interaction can substantially change inertial signatures while preserving the activity label. VSMP-IMU builds on this insight by introducing an SMP that captures activity-relevant primitives, temporal structure, body-part involvement, execution attributes, object interaction, and uncertainty, enabling controlled label-preserving variation before synthetic IMU generation.

\section{Approach}

\begin{figure*}[htbp]
    \centering
    \includegraphics[width=0.8\textwidth]{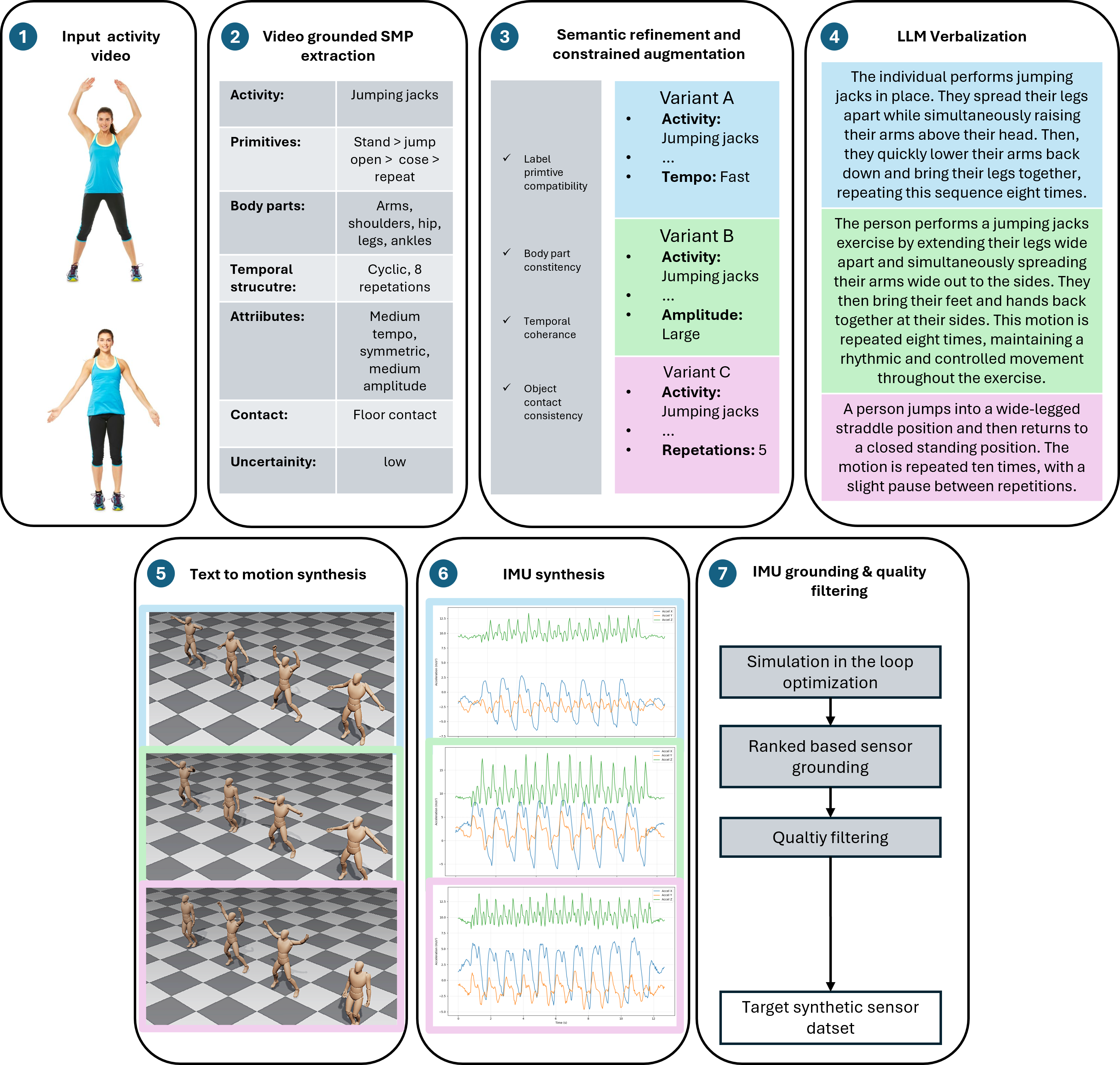}
    \caption{\textbf{Overview of VSMP-IMU.} VSMP-IMU extracts a video-grounded SMP from an activity video, refines and augments it under activity-preserving constraints, verbalizes each variant for HY-Motion synthesis, converts the generated motion into virtual IMU signals, and grounds and filters the signals to produce the final synthetic sensor dataset.}
    \label{fig:overview_pipeline}
\end{figure*}

\subsection{Overview}

VSMP-IMU generates synthetic IMU sequences that preserve activity identity while introducing realistic variation in motion execution and wearable sensing. Given an activity video $V$, the framework maps the video to a structured Semantic Motion Program (SMP), rather than directly converting noisy frame-level pose estimates or relying on a coarse text prompt.

As shown in Fig.~\ref{fig:overview_pipeline}, a pretrained video-capable VLM extracts an initial SMP $S_0$ describing the activity label, primitives, body parts, temporal structure, execution attributes, contact cues, and uncertainty. The SMP is refined and augmented under activity-preserving constraints to produce $K$ variants $\{S^{(k)}\}_{k=1}^{K}$. Each variant is verbalized into a HY-Motion-compatible prompt $Z^{(k)}$, synthesized into a 3D motion sequence $M^{(k)}$, converted into virtual IMU signals, adapted to the target wearable domain, and filtered to form the final synthetic set.

Target-domain adaptation uses two disjoint sources of training-fold statistics:
\begin{equation}
D_{\text{target}}=\{D_{\text{stat}},D_{\text{rank}}\},
\end{equation}
where $D_{\text{stat}}$ contains temporal and multivariate IMU statistics for simulation-in-the-loop optimization, and $D_{\text{rank}}$ contains empirical per-channel distributions for rank-based grounding. Both are estimated only from the real IMU training split within each LOPO fold.

Formally, VSMP-IMU extracts and refines an SMP as
\begin{equation}
S_0 = E_{\text{VLM}}(V), \qquad S = R(S_0),
\end{equation}
where $E_{\text{VLM}}(\cdot)$ is the video-grounded SMP extractor and $R(\cdot)$ resolves inconsistencies, validates constraints, and assigns uncertainty. The augmentation module samples activity-preserving variants:
\begin{equation}
S^{(k)} \sim A(S), \quad k = 1,\dots,K.
\end{equation}

Each variant is verbalized and synthesized into motion:
\begin{equation}
Z^{(k)} = T_{\text{HY}}(S^{(k)}), \qquad
M^{(k)} = G_{\text{HY}}(Z^{(k)}),
\end{equation}
where $T_{\text{HY}}(\cdot)$ denotes SMP-to-text verbalization and $G_{\text{HY}}(\cdot)$ denotes the HY-Motion generator.

The generated motion is converted into simulated IMU signals:
\begin{equation}
X_{\text{sim}}^{(k)}(\eta)
=
G_{\text{IMU}}(M^{(k)};\theta^{(k)},\eta),
\end{equation}
where $\theta^{(k)}$ contains the nominal virtual sensor configuration, including the target body segment, attachment point, nominal mounting transform, and nominal accelerometer/gyroscope bias and noise terms. The residual simulator parameter $\eta$ contains only bounded physical corrections to this nominal simulator, such as small mounting-orientation perturbations and bias/noise corrections. It does not change the target body segment, channel order, sign convention, unit convention, or gravity convention.

Simulation-in-the-loop optimization estimates residual simulator parameters using $D_{\text{stat}}$:
\begin{equation}
\eta^{(k)\star}
=
\arg\min_{\eta}
\mathcal{L}_{\text{sim}}
\left(
X_{\text{sim}}^{(k)}(\eta),
D_{\text{stat}}
\right),
\end{equation}
yielding
\begin{equation}
X_{\text{opt}}^{(k)}
=
X_{\text{sim}}^{(k)}(\eta^{(k)\star}).
\end{equation}
If this optimization is disabled, we set $\eta^{(k)\star}=0$.

Rank-based grounding then maps the optimized signal to the target wearable domain:
\begin{equation}
\widehat{X}^{(k)}_{\text{imu}}
=
G_{\text{rank}}
\left(
X_{\text{opt}}^{(k)},
\gamma^{(k)},
D_{\text{rank}}
\right),
\end{equation}
where $\gamma^{(k)}$ specifies post-simulation target-format conventions, such as axis order, sign flips, gravity inclusion or removal, unit conversion, and channel mapping. Unlike $\theta^{(k)}$ and $\eta$, $\gamma^{(k)}$ does not modify physical simulator parameters such as body segment, attachment point, mounting offset, bias, or noise. This separation keeps simulator-side physical calibration distinct from post-simulation convention alignment and final per-channel marginal grounding.

A final quality filter removes implausible, semantically inconsistent, or redundant samples:
\begin{equation}
Q(\widehat{X}^{(k)}_{\text{imu}}, M^{(k)}, S^{(k)}) \in \{0,1\}.
\end{equation}
Samples that pass the filter form $\mathcal{X}_{\text{synth}}$.

\begin{algorithm}[htbp]
\caption{VSMP-IMU}
\label{alg:VSMP_imu}
\begin{algorithmic}[1]
\Require Activity video $V$, number of variants $K$, sensor prior $p(\theta)$, target-format grounding specification $\Gamma$, target statistics $D_{\text{stat}}$, target rank distributions $D_{\text{rank}}$
\Ensure Filtered synthetic IMU set $\mathcal{X}_{\text{synth}}$

\State Extract and refine SMP: $S_0 \leftarrow E_{\text{VLM}}(V)$, $S \leftarrow R(S_0)$
\State Initialize $\mathcal{X}_{\text{synth}} \leftarrow \emptyset$

\For{$k = 1$ to $K$}
    \State Sample semantic variant: $S^{(k)} \sim A(S)$
    \State Verbalize and synthesize motion:
    $Z^{(k)} \leftarrow T_{\text{HY}}(S^{(k)})$,
    $M^{(k)} \leftarrow G_{\text{HY}}(Z^{(k)})$
    \State Sample nominal sensor parameters: $\theta^{(k)} \sim p(\theta)$
    \State Optimize bounded simulator residuals:
    \[
    \eta^{(k)\star}
    \leftarrow
    \arg\min_{\eta}
    \mathcal{L}_{\text{sim}}
    \left(
    G_{\text{IMU}}(M^{(k)};\theta^{(k)},\eta),
    D_{\text{stat}}
    \right)
    \]
    \State Generate optimized IMU:
    \[
    X_{\text{opt}}^{(k)}
    \leftarrow
    G_{\text{IMU}}(M^{(k)};\theta^{(k)},\eta^{(k)\star})
    \]
    \State Set or select target-format grounding parameters: $\gamma^{(k)} \leftarrow \Gamma$
    \State Ground synthetic IMU:
    \[
    \widehat{X}^{(k)}_{\text{imu}}
    \leftarrow
    G_{\text{rank}}
    \left(
    X_{\text{opt}}^{(k)},
    \gamma^{(k)},
    D_{\text{rank}}
    \right)
    \]
    \If{$Q(\widehat{X}^{(k)}_{\text{imu}}, M^{(k)}, S^{(k)}) = 1$}
        \State $\mathcal{X}_{\text{synth}} \leftarrow \mathcal{X}_{\text{synth}} \cup \{\widehat{X}^{(k)}_{\text{imu}}\}$
    \EndIf
\EndFor

\State \Return $\mathcal{X}_{\text{synth}}$
\end{algorithmic}
\end{algorithm}

\subsection{Semantic Motion Program}

The core representation in VSMP-IMU is the SMP, a structured intermediate representation between video understanding and inertial synthesis. The SMP is designed to be more structured and controllable than free-form text, while being less brittle than direct frame-level pose transfer. Its key role is to encode both \emph{what} activity is being performed and \emph{which aspects of that activity may vary} without changing the class label.

We define an SMP as
\begin{equation}
S = \{y, P, D, B, T, A, O, C, U\},
\end{equation}
where $y$ denotes the activity label, $P$ is a sequence of motion primitives, $D$ contains primitive-level duration information, $B$ encodes body-part involvement, $T$ specifies temporal organization, $A$ stores execution attributes, $O$ captures object and contact information, $C$ specifies augmentation constraints, and $U$ stores field-level uncertainty.

Table~\ref{tab:smp_fields} summarizes the SMP schema used in VSMP-IMU.

\begin{table*}[htbp]
\centering
\footnotesize
\caption{Fields of the SMP.}
\label{tab:smp_fields}
\begin{tabular}{llll}
\toprule
\textbf{Field} & \textbf{Type} & \textbf{Augmentable} & \textbf{Role in generation} \\
\midrule
Activity label $y$ & categorical & no & Defines target activity identity \\
Primitive sequence $P$ & ordered list & no & Defines core motion composition \\
Primitive durations $D$ & numeric sequence & yes & Controls local timing of motion phases \\
Body-part involvement $B$ & set / categorical & limited & Constrains dominant moving segments \\
Temporal structure $T$ & structured attributes & limited & Encodes ordering, repetition, and transitions \\
Tempo & scalar / categorical & yes & Controls overall motion speed \\
Amplitude & scalar / categorical & yes & Controls movement extent \\
Dominant side & categorical & conditional & Controls lateralized execution style \\
Symmetry & categorical & conditional & Controls symmetric or asymmetric execution style \\
Global displacement & scalar / categorical & conditional & Controls stationary or traveling motion \\
Object/contact $O$ & structured attributes & no & Preserves interaction semantics \\
Constraints $C$ & rule set & no & Specifies safe augmentation rules \\
Uncertainty $U$ & field-wise confidence & no & Restricts perturbations of unreliable fields \\
\bottomrule
\end{tabular}
\end{table*}

The central design principle is to split SMP fields into \emph{invariant} and \emph{variant} subsets. Invariant fields preserve class identity and remain fixed during augmentation. Variant fields change style, timing, or execution details while preserving the underlying activity. This factorization is especially important for wearable HAR, where useful augmentation should vary execution conditions without introducing semantic drift.

A concrete example for a jumping-jack activity is:
\begin{quote}
\textbf{Activity:} jumping jacks \\
\textbf{Primitives:} [stand $\rightarrow$ jump open $\rightarrow$ close $\rightarrow$ repeat] \\
\textbf{Body parts:} arms, shoulders, hips, legs, ankles \\
\textbf{Temporal structure:} cyclic, 8 repetitions \\
\textbf{Attributes:} medium tempo, symmetric, medium amplitude \\
\textbf{Contact:} floor contact \\
\textbf{Invariants:} jumping-jack identity, open/close cycle, bilateral arm and leg involvement, floor contact \\
\textbf{Variants:} tempo, amplitude, repetition count \\
\textbf{Uncertainty:} low
\end{quote}

This representation is more useful than a free-form caption because it preserves compositional structure and exposes exactly which factors may be modified safely during synthesis.
\subsection{Video-Grounded SMP Extraction with a Pretrained VLM}

Given an input activity video clip $V$, VSMP-IMU extracts an initial SMP using GPT-5.3, a pretrained video-capable vision-language model. In our implementation, the extractor operates at the clip level rather than the full untrimmed video level. Each clip is sampled into a fixed number of frames and paired with a structured prompting template that asks GPT-5.3 to output fields corresponding to the SMP schema.

Specifically, the extractor is prompted to predict:
(1) the activity label,
(2) the motion primitive sequence,
(3) body-part involvement,
(4) temporal organization,
(5) execution attributes such as tempo, amplitude, and symmetry,
(6) object/contact information when present, and
(7) uncertainty for ambiguous fields.
The prompt is structured rather than open-ended so that the model is encouraged to emit a JSON-like or slot-filled representation instead of free-form narrative text. A shortened prompt template is:
\begin{quote}
\small
Identify the activity shown in this video clip and describe it using the following fields:
activity label, motion primitives in temporal order, body parts involved, repetition pattern, tempo, amplitude, dominant side or symmetry, object/contact, and uncertainty for ambiguous fields.
Return the output in structured key-value form.
\end{quote}

The raw output is then parsed into the SMP schema. If the extractor returns malformed or partially missing fields, we apply a deterministic repair procedure: missing optional attributes are filled with \texttt{unknown}, malformed field names are mapped to the closest valid schema key, and incompatible values are rejected and reinitialized using conservative defaults.
%A longer prompt example and output format can be included in the appendix.

\paragraph{Extraction reliability audit.}
\revise{We separately evaluate whether the video-capable VLM extracts
the SMP fields reliably before they are used for augmentation. The audit
uses 200 manually annotated source clips spanning all five datasets,
activity types, motion scales, body-region focuses, temporal structures,
and object-interaction conditions. We compare the raw VLM responses,
the deterministically aggregated SMP, and the final SMP after semantic
refinement. The audit measures field-level accuracy, malformed and
missing outputs, repair frequency, residual semantic errors, and the
relationship between predicted uncertainty and extraction failure.}

\subsection{Semantic Refinement and Constrained Augmentation}

Because one-shot extraction may be incomplete, internally inconsistent, or hallucinated, VSMP-IMU refines the initial SMP before augmentation. The semantic refinement module applies four consistency checks, matching the pipeline in Fig.~\ref{fig:overview_pipeline}:
\begin{itemize}
    \item Label primitive compatibility: verifies that the predicted primitive sequence is plausible for the inferred activity label.
    \item Body part consistency: verifies that claimed body-part involvement matches the primitive descriptions.
    \item Temporal coherence: verifies that ordering, repetition count, and periodicity are mutually consistent.
    \item Object/contact consistency: verifies that contact patterns do not contradict the activity semantics.
\end{itemize}

Fields that fail hard consistency checks are corrected when possible or marked uncertain otherwise. Let $u_f \in [0,1]$ denote the uncertainty assigned to field $f$, where $u_f=0$ indicates high confidence and $u_f=1$ indicates maximum uncertainty. These uncertainty values are later used to control how aggressively each field may be perturbed during augmentation.

To increase diversity while preserving activity semantics, VSMP-IMU augments the refined SMP under explicit constraints. Let $\mathcal{I}(S)$ denote the invariant subset of fields and $\mathcal{V}(S)$ the variant subset. We sample an augmented program $S'$ according to
\begin{equation}
S' \sim p(S' \mid S), \qquad \text{s.t. } S'_{\mathcal{I}} = S_{\mathcal{I}},
\label{eq:safe_aug}
\end{equation}
so that activity-defining structure remains fixed.

In practice, augmentation is uncertainty-conditioned. For each variant field $f \in \mathcal{V}(S)$, we sample
\begin{equation}
f' \sim p_f(\cdot \mid f, u_f),
\end{equation}
where uncertainty directly controls the allowable perturbation magnitude. For continuous or ordinal fields, such as tempo, amplitude, repetition count, primitive duration, or displacement extent, we define a field-specific maximum perturbation radius $\Delta_f^{\max}$ and scale it by the confidence of the extracted field:
\begin{equation}
\Delta_f(u_f) = (1-u_f)\Delta_f^{\max}.
\label{eq:uncertainty_radius}
\end{equation}
The augmented value is then sampled from the bounded interval
\begin{equation}
f' \sim \mathrm{Uniform}\big[f-\Delta_f(u_f),\, f+\Delta_f(u_f)\big],
\label{eq:continuous_unc_aug}
\end{equation}
with clipping to the valid range of field $f$. Thus, low-uncertainty fields can be perturbed more broadly, while high-uncertainty fields are changed only slightly or left unchanged.

For categorical fields, such as dominant side or symmetry, uncertainty controls the probability of applying a valid categorical change:
\begin{equation}
P(f' \neq f) = (1-u_f)\rho_f,
\label{eq:categorical_unc_aug}
\end{equation}
where $\rho_f$ is the maximum allowed perturbation probability for field $f$. If a change is sampled, the replacement value is drawn only from the set of semantically valid alternatives specified by the activity constraints $C$. For example, a left/right dominant-side swap is allowed only for side-agnostic activities, and asymmetric execution is allowed only when it does not change the activity identity.

This design treats uncertainty conservatively. If the VLM confidently identifies tempo, amplitude, repetition count, or dominant side, VSMP-IMU can vary those fields to increase synthetic diversity. If the VLM is uncertain about a field, the system avoids making large changes based on potentially unreliable evidence, reducing the risk of semantic drift.

Sampling may be uniform within a valid range or prior-weighted using dataset-specific statistics, depending on the field. To prevent semantic drift, VSMP-IMU never modifies the activity label, core primitive order, mandatory body-part set, or required object/contact pattern. This differs from generic signal augmentation, which perturbs measurements directly without access to explicit semantic constraints.

For example, for a jumping-jack SMP, constrained variants may increase the tempo, increase the amplitude, or modify the repetition count while preserving the open/close primitive cycle and bilateral arm leg coordination.

\paragraph{Deterministic interventions for controllability evaluation.}
\revise{For the controllability study, we replace stochastic multi-attribute
augmentation with deterministic one-factor-at-a-time interventions.
Given a refined SMP $S$, intervention $I_a^\ell(S)$ modifies only
attribute $a$ to requested level $\ell$:}
\begin{equation}
S_{a,\ell}=I_a^\ell(S),
\qquad
S_{a,\ell}^{\setminus a}=S^{\setminus a}.
\end{equation}
\revise{The activity label, primitive sequence, mandatory body parts,
object/contact structure, and all non-intervened execution attributes
remain unchanged.}

\revise{We evaluate tempo and amplitude using three ordered levels, repetition
count using three requested counts, primitive duration using three
duration scales, and symmetry using two categorical conditions.
Interventions are applied only when they preserve the original
UTD-MHAD activity label.}

\subsection{SMP Verbalization}

Each augmented SMP is converted into a natural-language prompt before motion synthesis. This step is necessary because HY-Motion is text-conditioned and does not consume the structured SMP directly. In our implementation, the SMP-to-text module is designed specifically for HY-Motion rather than for generic caption generation. The verbalized prompt is passed to HY-Motion's native text-conditioning stack, which uses the HYTextModel encoder with a Qwen3-based LLM branch and an auxiliary CLIP ViT-L/14 sentence-embedding branch. Thus, the verbalization stage is optimized to produce prompts that match the linguistic style and motion-description format expected by HY-Motion's text encoder.

Given an SMP $S^{(k)}$, the verbalizer produces a HY-Motion-compatible text prompt:
\begin{equation}
Z^{(k)} = T_{\text{HY}}(S^{(k)}),
\end{equation}
where $T_{\text{HY}}(\cdot)$ denotes the SMP-to-text verbalization function. The function maps structured SMP fields into a concise motion description that can be encoded by HY-Motion's pretrained and motion-aligned text encoder. The goal is not to generate an open-ended video caption, but to produce a text-conditioned motion-generation prompt that remains faithful to the structured SMP while being compatible with the prompt distribution used by HY-Motion.

The verbalizer preserves the activity label, primitive order, temporal structure, body-part involvement, contact information, and controllable execution attributes. In particular, it expresses motion primitives as ordered action phrases, converts repetition and tempo fields into explicit temporal descriptions, and maps amplitude, symmetry, and dominant-side attributes into natural-language modifiers. This keeps the prompt semantically constrained while making it readable by HY-Motion's Qwen3/CLIP-based text encoder.

Representative HY-Motion-compatible verbalizations for jumping-jack variants include:
\begin{quote}
\small
\textbf{High tempo:} The individual performs jumping jacks in place. They spread their legs apart while simultaneously raising their arms above their head. Then, they quickly lower their arms back down and bring their legs together, repeating this sequence eight times.
\end{quote}
\begin{quote}
\small
\textbf{High amplitude:} The person performs a jumping jacks exercise by extending their legs wide apart and simultaneously spreading their arms wide out to the sides. They then bring their feet and hands back together at their sides. This motion is repeated eight times, maintaining a rhythmic and controlled movement throughout the exercise.
\end{quote}
\begin{quote}
\small
\textbf{More repetitions:} A person jumps into a wide-legged straddle position and then returns to a closed standing position. The motion is repeated ten times, with a slight pause between repetitions.
\end{quote}

The verbalizer is constrained to include only fields supported by the SMP. This prevents the prompt from introducing new objects, unsupported body parts, or activity changes during natural-language conversion. As a result, HY-Motion receives a prompt that is both semantically grounded in the extracted SMP and formatted for its motion-specific text encoder, rather than a generic free-form caption.

\paragraph{Matched intervention prompts.}
\revise{For the controllability experiment, prompts within each intervention
set use identical deterministic templates and differ only in the phrase
representing the tested attribute. The activity label, primitive
descriptions, body-part references, contact information, temporal
ordering, and all non-intervened attributes remain unchanged.}

\revise{For example, tempo prompts differ only through the phrases
``at a deliberately slow pace,'' ``at a moderate pace,'' and
``at a deliberately fast pace.'' Amplitude prompts similarly differ
only through ``small,'' ``medium,'' and ``large range of motion.''
This matched design prevents unrelated lexical variation from
confounding the control evaluation.}

\subsection{Text-to-Motion Synthesis with HY-Motion}

HY-Motion \cite{wen2025hy} is an external pretrained text-to-motion model used as the motion synthesis backend in VSMP-IMU. We do not train or fine-tune HY-Motion. Given a text prompt produced by the SMP verbalizer, HY-Motion generates a 3D human-motion sequence represented as joint positions, rotations, or equivalent kinematic body states, depending on the released checkpoint format. VSMP-IMU then converts this generated motion into segment-level trajectories and orientations required for virtual IMU synthesis.

The verbalized motion description $Z^{(k)}$ conditions HY-Motion to generate a temporally ordered 3D body-motion sequence:
\begin{equation}
M^{(k)} = G_{\text{HY}}(Z^{(k)}) = \{q_t^{(k)}\}_{t=1}^{T_m},
\end{equation}
where $q_t^{(k)}$ denotes the body state at time $t$. Depending on the HY-Motion output format, $q_t^{(k)}$ may represent skeleton joint positions, joint rotations, or other kinematic motion parameters. In our setting, the generated output is converted into segment-level trajectories and orientations required for IMU synthesis.

Multiple motion realizations may be generated from the same verbalized SMP by varying the generation seed or decoding configuration of HY-Motion. This allows VSMP-IMU to capture both semantic-level variability through SMP augmentation and realization-level variability through stochastic text-to-motion synthesis.

\revise{For each base SMP, all intervention conditions are generated using the
same HY-Motion seed and identical decoding configuration:}
\begin{equation}
M_{a,\ell,s}
=
G_{\mathrm{HY}}
\left(
T_{\mathrm{HY}}(I_a^\ell(S));s
\right),
\end{equation}
\revise{where $a$ denotes the tested attribute, $\ell$ denotes the requested
condition, and $s$ denotes one of five fixed generation seeds.
This paired-seed design reduces stochastic motion-generation variation
when comparing requested control levels.}

\subsection{IMU Synthesis}

The generated motion sequence is converted into virtual IMU data using an IMU synthesis module implemented with IMUSim. Given a motion sequence $M^{(k)}$, the simulator attaches one or more virtual sensors to selected body segments and produces accelerometer and gyroscope measurements from the underlying segment kinematics:
\begin{equation}
\tilde{X}^{(k)}_{\text{imu}}
=
G_{\text{IMU}}(M^{(k)}, \theta^{(k)}),
\end{equation}
where $\theta^{(k)}$ denotes the nominal virtual sensor configuration, including the target body segment, attachment point, nominal sensor-to-segment mounting transform, and nominal accelerometer/gyroscope bias and noise terms. Bounded residual corrections to these nominal simulator parameters are optimized later through $\eta$ in the IMU grounding stage.

The initial simulated sequence $\tilde{X}^{(k)}_{\text{imu}}$ is physically meaningful because it is derived from generated body kinematics. However, raw simulator outputs do not necessarily match the target wearable dataset's temporal statistics, cross-channel structure, coordinate-frame convention, sign convention, gravity convention, unit convention, channel layout, or magnitude statistics. VSMP-IMU handles these differences in the IMU grounding and quality-filtering stage: simulation-in-the-loop optimization adjusts bounded physical residuals through $\eta$, while target-format convention alignment and rank-based grounding use $\gamma$ and $D_{\text{rank}}$ to align the representation and marginal channel distributions.

\revise{For the controllability evaluation, outputs are retained at three
checkpoints: generated motion $M_{a,\ell,s}$, raw simulated IMU
$\widetilde{X}_{a,\ell,s}$, and final grounded IMU
$\widehat{X}_{a,\ell,s}$. Within each matched intervention set, the
virtual sensor segment, mounting transform, orientation, bias, noise
realization, channel convention, and grounding configuration remain
fixed.}

\revise{This design identifies whether a requested SMP control is lost during
text-to-motion synthesis, motion-to-IMU conversion, or target-domain
grounding.}

\subsection{IMU Grounding and Quality Filtering}

The final stage adapts the raw synthetic IMU sequence to the target wearable domain and filters generated samples before they are mixed with real training data. This stage contains three separate components applied in sequence: simulation-in-the-loop optimization, target-format convention alignment with rank-based sensor grounding, and quality filtering. These components address different parts of the synthetic-to-real gap. Simulation-in-the-loop optimization tunes bounded physical simulator residuals using temporal and multivariate IMU statistics. Target-format convention alignment then maps the optimized signal to the coordinate, sign, gravity, unit, and channel conventions of the target dataset. Rank-based grounding, inspired by the rank-based virtual-to-real calibration strategy used in IMUTube, performs the final non-parametric channel-wise marginal distribution alignment. Quality filtering removes implausible, semantically inconsistent, or redundant samples.

We decompose the target-domain information into two disjoint parts:
\begin{equation}
D_{\text{target}}
=
\{D_{\text{stat}}, D_{\text{rank}}\},
\end{equation}
where $D_{\text{stat}}$ contains temporal and multivariate statistics used by simulation-in-the-loop optimization, and $D_{\text{rank}}$ contains empirical per-channel cumulative distribution functions used only by rank-based grounding. Both are estimated exclusively from the real IMU training split within each LOPO fold.

\paragraph{Simulation-in-the-loop optimization.}
The raw synthetic IMU signal is generated from a motion sequence $M$ using a virtual IMU simulator. Although the generated signal is physically meaningful, it may depend on uncertain simulator choices such as body-segment attachment, attachment point, nominal sensor mounting orientation, bias, and noise level. We therefore first optimize bounded simulator-side residuals before applying any target-format convention alignment or rank-based marginal transformation.

Let $\theta$ denote the nominal virtual sensor configuration:
\begin{equation}
\theta =
\{j_{\text{target}}, r_j, R_{s \leftarrow j}^{0}, b_a^{0}, b_\omega^{0}, \sigma_a^{0}, \sigma_\omega^{0}\},
\end{equation}
where $j_{\text{target}}$ denotes the target body segment, $r_j$ is the attachment point on that segment, $R_{s \leftarrow j}^{0}$ is the nominal sensor-to-segment mounting transform, $b_a^{0}$ and $b_\omega^{0}$ are nominal accelerometer and gyroscope biases, and $\sigma_a^{0}$ and $\sigma_\omega^{0}$ parameterize nominal channel noise scales.

We introduce a bounded residual simulator parameter $\eta$ that adjusts only physical corrections to this nominal simulator:
\begin{equation}
\eta =
\{\Delta R_s, \Delta b_a, \Delta b_\omega, \kappa_a, \kappa_\omega\},
\end{equation}
where $\Delta R_s$ is a small mounting-orientation perturbation, $\Delta b_a$ and $\Delta b_\omega$ are bias corrections, and $\kappa_a$ and $\kappa_\omega$ scale the nominal accelerometer and gyroscope noise. The residual parameter $\eta$ does not change the target body segment, attachment point, channel order, sign convention, unit convention, or gravity convention. The simulator output is
\begin{equation}
X_{\text{sim}}(\eta)
=
G_{\text{IMU}}(M;\theta,\eta).
\end{equation}

The residual parameters are estimated by minimizing a simulation-level objective:
\begin{equation}
\eta^\star
=
\arg\min_{\eta}
\mathcal{L}_{\text{sim}}
\left(
X_{\text{sim}}(\eta),
D_{\text{stat}}
\right),
\label{eq:sim_loop_opt}
\end{equation}
where $D_{\text{stat}}$ contains temporal and multivariate statistics estimated only from the real IMU training split. To keep this step separate from rank-based grounding, $\mathcal{L}_{\text{sim}}$ does not match the per-channel empirical cumulative distributions. Instead, it compares temporal and multivariate statistics that are not captured by channel-wise rank transformation:
\begin{equation}
\mathcal{L}_{\text{sim}}
=
\lambda_{\Sigma}
\left\|
\Sigma(X_{\text{sim}}) - \Sigma(D_{\text{stat}})
\right\|_F
+
\lambda_{\text{psd}}
\left\|
\mathrm{PSD}(X_{\text{sim}}) - \mathrm{PSD}(D_{\text{stat}})
\right\|_2
+
\lambda_{r}
\left\|
r(X_{\text{sim}}) - r(D_{\text{stat}})
\right\|_2.
\label{eq:sim_loss}
\end{equation}
Here, $\Sigma(\cdot)$ denotes cross-channel covariance, $\mathrm{PSD}(\cdot)$ denotes the spectral energy distribution, and $r(\cdot)$ contains vector-norm and autocorrelation statistics. This objective encourages the simulated IMU to have realistic cross-axis structure and temporal dynamics while leaving final per-channel marginal alignment to the rank-based grounding step.

The optimized synthetic IMU sequence is then
\begin{equation}
X_{\text{opt}}
=
X_{\text{sim}}(\eta^\star).
\label{eq:optimized_imu}
\end{equation}
If simulation-in-the-loop optimization is disabled, we set $\eta^\star=0$ and use $X_{\text{opt}}=G_{\text{IMU}}(M;\theta,0)$ directly.

\paragraph{Rank-based sensor grounding.}
After simulation-in-the-loop optimization, VSMP-IMU applies target-format convention alignment followed by rank-based sensor grounding. Convention alignment maps the optimized synthetic IMU into the channel representation expected by the target dataset. Rank-based grounding then aligns the marginal channel distributions of the convention-aligned synthetic IMU with the target real-sensor domain. This step is inspired by IMUTube's virtual-to-real distribution matching, but is applied here to synthetic IMU generated from SMP variants and text-to-motion synthesis.

Before rank transformation, we apply target-format convention alignment using
\begin{equation}
\gamma =
\{P_{\text{axis}}, c_{\text{sign}}, g_{\text{conv}}, s_{\text{unit}}, C_{\text{map}}\},
\end{equation}
where $P_{\text{axis}}$ is the target dataset's axis order, $c_{\text{sign}}$ specifies sign flips, $g_{\text{conv}}$ specifies gravity inclusion or removal according to the target dataset convention, $s_{\text{unit}}$ specifies unit conversion, and $C_{\text{map}}$ specifies the target channel layout. Unlike $\theta$ and $\eta$, $\gamma$ is a post-simulation representation transform; it does not alter physical simulator parameters such as body segment, attachment point, mounting offset, bias, or noise. This produces a convention-aligned sequence
\begin{equation}
X_{\text{align}}
=
A_{\gamma}(X_{\text{opt}}),
\end{equation}
where $A_{\gamma}(\cdot)$ applies axis, sign, gravity, unit, and channel-layout alignment.

Rank-based grounding then performs channel-wise quantile matching. Let $x_v$ denote a scalar sample from a convention-aligned synthetic IMU channel and let $x_r$ denote the corresponding grounded sample in the target real-sensor domain. For each channel independently, we estimate the empirical cumulative distribution function of the convention-aligned synthetic samples, $F_v(\cdot)$, and the empirical cumulative distribution function of the real target-domain training samples, $G_r(\cdot)$. The grounded value is obtained by matching quantile ranks:
\begin{equation}
x_r
=
G_r^{-1}
\left(
F_v(x_v)
\right).
\label{eq:rank_grounding}
\end{equation}

Equivalently, each synthetic sample is first mapped to its percentile under the synthetic distribution and is then replaced by the real-domain value at the same percentile. This preserves the temporal ordering and relative motion pattern of the generated sequence while aligning its per-channel marginal distribution with the target wearable sensor domain.

For a multi-channel IMU sequence, the transformation is applied independently to each channel:
\begin{equation}
\widehat{X}_{\text{imu}}^{(c)}
=
G_{r,c}^{-1}
\left(
F_{v,c}
\left(
X_{\text{align}}^{(c)}
\right)
\right),
\qquad
c \in \{a_x,a_y,a_z,\omega_x,\omega_y,\omega_z\},
\label{eq:channel_rank_grounding}
\end{equation}
where $a_x,a_y,a_z$ are accelerometer channels and $\omega_x,\omega_y,\omega_z$ are gyroscope channels. The empirical real-domain distributions $G_{r,c}$ are estimated only from $D_{\text{rank}}$, which is constructed from real IMU training data within each training fold, preventing access to the held-out test subject.

The complete grounding operation is written as
\begin{equation}
\widehat{X}_{\text{imu}}
=
G_{\text{rank}}
\left(
X_{\text{opt}},
\gamma,
D_{\text{rank}}
\right),
\label{eq:grounding_operator}
\end{equation}
where $D_{\text{rank}}$ contains empirical per-channel distributions estimated from the real training split.

This ordering keeps the two adaptation steps mutually separate. Simulation-in-the-loop optimization adjusts bounded physical simulator residuals using multivariate and temporal statistics from $D_{\text{stat}}$, while target-format convention alignment and rank-based grounding perform representation alignment and final non-parametric marginal matching using $\gamma$ and $D_{\text{rank}}$. The optimization loss therefore does not rematch the same per-channel empirical distributions handled by the rank transform.

Together, these steps model practical wearable variability, including strap rotation, placement inconsistency, coordinate-frame mismatch, gravity-convention differences, unit or channel-layout differences, device-specific noise, and magnitude differences across sensors.

\paragraph{Quality filtering.}
Not all generated samples are useful for downstream HAR. We therefore apply a quality filter before mixing synthetic data with real training samples. The filter contains three components with different implementations: a rule-based motion plausibility filter, a supervised class-faithfulness filter, and an embedding-based redundancy filter. All trainable components are fit only on the real IMU training split within each LOPO fold; no held-out subject data is used.

First, the motion plausibility filter is rule-based and does not require training. It rejects motion sequences and synthetic IMU samples whose kinematics or inertial values violate simple physical sanity checks, such as abrupt frame-to-frame discontinuities, unrealistically large acceleration or angular-velocity magnitudes, invalid joint trajectories, or missing/degenerate motion segments:
\begin{equation}
Q_{\text{motion}}(M,\widehat{X}_{\text{imu}}) \in \{0,1\}.
\end{equation}
In practice, $Q_{\text{motion}}(M,\widehat{X}_{\text{imu}})=0$ if any motion or inertial statistic exceeds a predefined physically plausible range estimated from the training split.

Second, the class-faithfulness filter is implemented using a lightweight HAR validator trained only on real IMU training data from the current fold. This validator is separate from the final downstream HAR model used for evaluation. Its purpose is not to improve recognition directly, but to reject synthetic samples whose generated IMU pattern is inconsistent with the intended activity label $y$. Given a grounded synthetic sequence $\widehat{X}_{\text{imu}}$, the validator outputs class probabilities $p(c \mid \widehat{X}_{\text{imu}})$. Synthetic samples with confidence below a threshold $\tau_{\text{cls}}$ or with a predicted label different from $y$ are discarded:
\begin{equation}
Q_{\text{class}}(\widehat{X}_{\text{imu}}, y) =
\mathbb{I}\left[
\max_c p(c \mid \widehat{X}_{\text{imu}}) \geq \tau_{\text{cls}}
\ \wedge\
\arg\max_c p(c \mid \widehat{X}_{\text{imu}}) = y
\right].
\end{equation}
The threshold $\tau_{\text{cls}}$ is selected using the validation split within the training subjects, preventing leakage from the held-out test subject.

Third, the redundancy filter uses an embedding representation rather than a separately trained generative model. We use the penultimate-layer feature representation of the trained HAR validator as the feature encoder $\phi(\cdot)$. A generated sample is rejected if it is too close to an already retained synthetic sample in this feature space:
\begin{equation}
Q_{\text{red}}(\widehat{X}_{\text{imu}}) =
\mathbb{I}\left[
\min_{X' \in \mathcal{X}_{\text{synth}}}
\|\phi(\widehat{X}_{\text{imu}}) - \phi(X')\|_2
> \tau_{\text{nn}}
\right].
\end{equation}
This step prevents the synthetic set from being dominated by near-duplicate motion realizations. The nearest-neighbor threshold $\tau_{\text{nn}}$ is chosen on the training/validation split and is fixed before testing.

The final quality decision is
\begin{equation}
Q(\widehat{X}_{\text{imu}}, M, S) =
Q_{\text{motion}}(M,\widehat{X}_{\text{imu}})
\cdot
Q_{\text{class}}(\widehat{X}_{\text{imu}}, y)
\cdot
Q_{\text{red}}(\widehat{X}_{\text{imu}}).
\end{equation}

Only samples that pass all three filters are retained. The retained set $\mathcal{X}_{\text{synth}}$ forms the target synthetic sensor dataset used for downstream HAR training.

\section{Results}

\subsection{Experimental Setup}

\subsubsection{Datasets}

We evaluate VSMP-IMU on paired video--IMU benchmarks and IMU-only HAR benchmarks. Paired datasets enable direct evaluation of video-grounded synthesis, while IMU-only datasets test whether external public videos can generate useful synthetic inertial data without paired target-domain video. All experiments use leave-one-person-out (LOPO) evaluation and report results averaged over all held-out subjects and five random seeds.

The evaluated datasets are:

\begin{itemize}
    \item \textbf{MM-Fit} \cite{stromback2020mm}: a paired video--IMU dataset with 10 activity classes and 5 sensor positions, focused on structured fitness exercises such as squats, lunges, sit-ups, and bicep curls. It evaluates cyclic motion structure, body-part coordination, and repetition-level variation.

    \item \textbf{UTD-MHAD} \cite{chen2015utd}: a paired multimodal dataset with 27 activity classes and 2 sensor positions, covering gestures, sports-like actions, locomotion, sitting/standing transitions, lunges, and squats. It tests generalization across diverse motion types.

    \item \textbf{MMAct} \cite{kong2019mmact}: a paired video--IMU dataset with 37 activity classes and 3 sensor positions, covering daily, abnormal, and desk-work actions across multiple scenes. It evaluates scalability to larger and more heterogeneous activity vocabularies.

    \item \textbf{PAMAP2} \cite{reiss2012introducing}: an IMU-only benchmark with 18 activity classes and 3 sensor positions, including physical activities and activities of daily living such as walking, running, cycling, stair ascent/descent, ironing, vacuum cleaning, and rope jumping.

    \item \textbf{HAD-AW} \cite{ashry2018lstm}: an IMU-only smartwatch benchmark with 31 activity classes, including ambulatory and fine-grained wrist-centric daily activities such as eating and brushing teeth. It provides a challenging setting for low-resource and subject-generalizable HAR.
\end{itemize}

\subsubsection{Source Video Collection for IMU-Only Target Datasets}

For IMU-only datasets, synthetic inertial sequences are generated from external public videos. For each activity class, we retrieve YouTube candidates using queries such as \textit{``[activity name]''}, \textit{``person [activity name]''}, and \textit{``[activity name] exercise / daily activity''}. We collect five clips per class to capture variation in execution style and duration.

To reduce label mismatch, clips are filtered in two stages. A video-language model first checks agreement between the retrieved label and clip-level semantics. Ambiguous clips are then removed using manual verification and confidence-threshold filtering. Clips are discarded when the target activity is not dominant, when multiple competing actions are present, or when the activity semantics do not match the target label. All video-based methods use the same retained source video pool.

\subsubsection{Protocols}

We evaluate three LOPO settings. In each fold, one subject is held out for testing, while the remaining subjects are used for training and validation. All preprocessing statistics, grounding distributions, validation thresholds, and trainable filtering components are estimated only from non-test subjects.

For paired datasets, all videos and IMU windows from the held-out subject are excluded from synthetic generation, VLM extraction, grounding, validation, and filtering.

\textbf{Standard LOPO:}
Models are trained on the full training split and evaluated on the held-out subject, measuring subject-generalization performance under full-data conditions.

\textbf{Low-resource LOPO:}
For all five datasets, we retain only 1\%, 5\%, 10\%, 25\%, or 50\% of labeled training data from non-test subjects. Synthetic IMU is added to this reduced training set, and all methods are evaluated on the same held-out subject. For paired video–IMU datasets, synthetic data is generated only from videos belonging to non-test subjects; for IMU-only datasets, synthetic data is generated from the filtered external public video pool described above.

\textbf{Long-tail LOPO:}
To evaluate class imbalance, we subsample each training split into a long-tailed distribution while keeping the held-out test subject unchanged. Classes are ranked by the number of real IMU windows in the training split. For class rank $r$, the retained sample count is
\begin{equation}
n_r = n_{\max}\rho^{-\frac{r-1}{C-1}},
\end{equation}
where $C$ is the number of classes, $n_{\max}$ is the count for the largest class, and $\rho=10$ is the maximum head-to-tail imbalance ratio. Classes in the upper half of the ranked list are reported as head classes, and those in the lower half as tail classes. Synthetic data is generated either for tail classes only or in proportion to class scarcity.

\subsubsection{Baselines}

We compare VSMP-IMU with four baselines:

\begin{itemize}
    \item \textbf{Real-only}: trains the HAR model only on real IMU data from training subjects.
    \item \textbf{Classical Aug}: applies signal-level augmentations, including jittering, scaling, rotation, and time warping.
    \item \textbf{IMUTube}: derives virtual IMU from source videos by estimating pose and converting motion to inertial signals.
    \item \textbf{IMUGPT 2.0}: generates synthetic IMU from textual activity prompts using text-conditioned motion synthesis followed by IMU conversion.
\end{itemize}

All generation-based methods use the same downstream HAR backbone, training schedule, synthetic sample budget, and source video pool when applicable.

\subsubsection{Evaluation Metrics}

We report \textbf{Macro-F1} as the primary metric because it gives equal weight to each activity class and is more informative than accuracy under low-resource and class-imbalanced conditions. Unless otherwise stated, results are reported as mean and standard deviation over five random seeds and all LOPO folds.
\subsubsection{Implementation Details}

VSMP-IMU extracts an initial SMP from each source video using GPT-5.3 as the video-capable VLM. The SMP is refined using rule-based checks for label--primitive compatibility, body-part consistency, temporal coherence, and object/contact consistency, then augmented under the uncertainty-conditioned rules described earlier.

Each SMP variant is converted into a HY-Motion-compatible natural-language prompt. HY-Motion is used as a frozen pretrained text-to-motion backend to synthesize the corresponding 3D human motion, which is converted into virtual inertial measurements using IMUSim.

After IMU synthesis, simulation-in-the-loop optimization adjusts bounded simulator-side residuals, including small mounting-orientation perturbations, bias corrections, and noise-scale corrections, using temporal and multivariate statistics from the real IMU training split. Target-format conventions such as axis order, sign convention, gravity inclusion or removal, unit conversion, and channel mapping are handled separately during convention alignment and rank-based grounding. Rank-based grounding is then applied to align per-channel marginal distributions with the target wearable domain. All statistics are estimated only from the training subjects in the current LOPO fold.

Synthetic and real samples are mixed using a synthetic-to-real ratio $\alpha \in [0.1,2.0]$. We evaluate four downstream HAR backbones: Random Forest with empirical cumulative distribution function features, DeepConvLSTM \cite{ordonez2016deep}, attention-based DeepConvLSTM \cite{singh2020deep}, and BiLSTM with Attention \cite{zhang2023attention}. Neural models use 2-second input windows, batch size 1024, AdamW with learning rate $1\times10^{-3}$, early stopping on validation Macro-F1, and up to $N=500$ epochs.

\subsubsection{End-to-End SMP Controllability Protocol}
\label{sec:control_protocol}
\revise{We evaluate end-to-end SMP controllability using 12 activities from
UTD-MHAD. The selected activities cover repetitive and discrete
actions, unilateral and bilateral motion, upper- and lower-body
movement, and both fine- and large-amplitude actions.}

\revise{The selected activities are right-hand wave, two-hand front clap,
right-arm throw, basketball shoot, right-hand draw circle clockwise,
front boxing, two-arm curl, two-hand push, jogging in place, walking
in place, forward lunge, and squat with arms extended.}

\revise{Tempo and amplitude are evaluated on all 12 activities. Repetition
count is evaluated on right-hand wave, two-hand front clap, right-hand
draw circle clockwise, front boxing, two-arm curl, jogging in place,
walking in place, and squat with arms extended. Primitive duration is
evaluated on right-arm throw, basketball shoot, two-hand push, and
forward lunge. Symmetry is evaluated on two-hand front clap, two-arm
curl, two-hand push, and squat with arms extended.}

\revise{Tempo uses slow, medium, and fast conditions. Amplitude uses small,
medium, and large conditions. Repetition count uses 3, 6, and 9
repetitions. Primitive duration uses scaling factors of 0.75, 1.00,
and 1.25. Symmetry uses symmetric and asymmetric conditions.}

\revise{Each intervention condition is generated using five fixed HY-Motion
seeds. Only one SMP field changes within each matched intervention set.
This produces 180 tempo motions, 180 amplitude motions, 120 repetition
motions, 60 primitive-duration motions, and 40 symmetry motions, for a
total of 580 generated motion sequences.}

\begin{table}[t]
\centering
\caption{UTD-MHAD end-to-end SMP controllability protocol. Only one SMP field is changed within each matched intervention set.}
\label{tab:control_protocol}
\begin{tabular}{lccc}
\toprule
\revise{Control} & \revise{Activities} & \revise{Conditions} & \revise{Motions} \\
\midrule
Tempo & 12 & slow/medium/fast & 180 \\
Amplitude & 12 & small/medium/large & 180 \\
Repetitions & 8 & 3/6/9 & 120 \\
Primitive duration & 4 & 0.75/1.00/1.25 & 60 \\
Symmetry & 4 & symmetric/asymmetric & 40 \\
\midrule
Total & -- & -- & 580 \\
\bottomrule
\end{tabular}
\end{table}

\revise{Tempo is measured using cycle frequency for repetitive actions and mean
principal-joint velocity for discrete actions. Amplitude is measured
using principal-joint range of motion. Repetition count is measured
using cycle detection. Primitive duration is measured from the duration
of the affected motion phase. Symmetry is measured using mirrored
left--right trajectory similarity.}

\revise{For raw and grounded IMU, tempo is measured using dominant frequency
and autocorrelation period, amplitude using gyroscope and acceleration
energy, repetition count using peak-based cycle detection, and primitive
duration using corresponding signal landmarks. Symmetry is evaluated
only at the motion level because UTD-MHAD does not provide bilateral
wearable sensors.}

\revise{Activity-label retention is measured through blinded inspection of the
generated motion sequences by two independent annotators. Each
annotator determines whether the generated sequence retains the
original UTD-MHAD activity identity without access to the requested
intervention condition. Disagreements are resolved through
adjudication. We report the proportion of sequences judged to retain
the intended label.}

\subsubsection{Strictly Matched Caption--SMP Ablation Protocol}
\label{sec:matched_caption_smp_protocol}

\revise{We use a $2\times2$ factorial ablation to separate the effect
of structured representation from the effect of explicit semantic
variation. The first factor is representation type
$\{\text{caption},\text{SMP}\}$, and the second is semantic variation
$\{\text{base},\text{augmented}\}$. This produces four conditions:
(C0) a generic video caption without explicit attribute variation;
(C1) the same caption supplemented with matched attribute
instructions; (S0) a refined but unaugmented SMP verbalization; and
(S1) the complete constrained augmented SMP used by VSMP-IMU.}

\revise{For each activity class, we use five source clips from the
current training fold and generate $K=4$ motion sequences per clip.
The four generation slots correspond to a reference execution, a tempo
intervention, an amplitude intervention, and one additional
activity-valid intervention selected from repetition count, primitive
duration, symmetry, dominant side, or displacement. Conditions C1 and
S1 receive exactly the same requested attribute values. The associated
natural-language attribute phrase is inserted verbatim into both
prompts.}

\revise{Conditions C0 and S0 receive no explicit semantic
intervention. Their unchanged base caption or SMP verbalization is
generated four times using the same four HY-Motion seeds used by C1
and S1. Thus, all conditions contain the same number of generation
slots and differ only in representation type and whether matched
semantic variation is applied.}

\revise{All four conditions use identical source clips, activity
labels, HY-Motion seeds and decoding parameters, motion duration
constraints, IMUSim settings, virtual sensor placements, sensor-domain
adaptation, quality-filter thresholds, synthetic-to-real mixing ratio,
window segmentation, and downstream HAR training protocol. We use
$\alpha=0.5$ for MM-Fit, UTD-MHAD, and MMAct, and $\alpha=0.25$ for
PAMAP2 and HAD-AW.}

\revise{To eliminate differences in retained sample quantity, filtering
uses matched-block replenishment. Outputs generated from the same
source clip, generation slot, and seed form a four-condition block.
A block enters downstream training only when all four outputs pass the
motion-plausibility, class-faithfulness, and redundancy filters. When
any output fails, the complete block is regenerated using the next seed
from a shared reserve-seed list. Generation continues until each
condition contains the same number of retained samples in every
activity class.}

\begin{table*}[htbp]
\centering
\scriptsize
\caption{\textbf{Strictly matched caption--SMP ablation protocol.}
Caption and SMP conditions use identical source clips, generation
seeds, attribute requests where applicable, sensor processing, filters,
retained sample counts, and downstream training settings.}
\label{tab:matched_caption_protocol}
\begin{tabular}{llllll}
\toprule
\revise{ID} &
\revise{Representation} &
\revise{Variation} &
\revise{Attribute requests} &
\revise{Prompt across $K=4$} &
\revise{Retained budget} \\
\midrule
C0 &
Generic caption &
None &
None &
Base caption with four matched seeds &
Equal \\

C1 &
Generic caption &
Matched &
Identical to S1 &
Caption plus matched attribute phrases &
Equal \\

S0 &
Refined SMP &
None &
None &
Base SMP with four matched seeds &
Equal \\

S1 &
Refined SMP &
Constrained &
Identical to C1 &
Four constrained SMP verbalizations &
Equal \\
\bottomrule
\end{tabular}
\end{table*}

\subsubsection{SMP Extraction Audit Protocol}
\label{sec:smp_audit_protocol}
\revise{We evaluate SMP extraction reliability using a manually
annotated audit set of 200 source videos, with 40 clips sampled from
each of MM-Fit, UTD-MHAD, MMAct, PAMAP2, and HAD-AW. The subset is
stratified to include periodic and non-periodic activities,
large- and small-amplitude motion, upper-body, lower-body, and
whole-body movement, unilateral and bilateral execution,
object-interaction and no-object activities, stationary and locomotion
activities, and both full-body and hand-centric actions.}

\revise{Two annotators independently inspect each full video clip and
construct a reference SMP using the same schema supplied to the VLM.
The annotation contains the activity label, ordered primitive sequence,
primitive duration or relative phase length, body-part involvement,
periodicity, repetition count, tempo, amplitude, dominant side,
symmetry, global displacement, object/contact relations, invariant
fields, and safely augmentable variant fields. Annotators do not see
the VLM outputs during annotation. Disagreements are resolved through
adjudication, and the adjudicated SMP is used as the reference.}

\revise{We preserve three outputs for every audited clip. The first
stage consists of the five raw VLM responses. The second is the
aggregated SMP obtained using majority voting for categorical fields,
median aggregation for numeric and ordinal fields, and maximum
pairwise agreement for the primitive sequence. The third is the
refined SMP obtained after deterministic schema repair and the four
semantic consistency checks. Evaluating all three stages allows us to
separate raw extraction errors from errors corrected through
aggregation or semantic refinement.}

\revise{Activity labels, periodicity, dominant side, symmetry, and
global displacement are evaluated using exact accuracy. Body-part,
object/contact, invariant, and variant sets are evaluated using
Macro-F1. Primitive sequences are evaluated using both exact-sequence
accuracy and normalized edit similarity. Repetition count is evaluated
using mean absolute error, while tempo and amplitude are evaluated
using ordinal accuracy.}

\revise{We additionally report malformed-JSON rate, missing-mandatory-
field rate, invalid-value rate, five-response disagreement, deterministic
schema-repair rate, semantic-consistency correction rate, clip-discard
rate, and residual refined-SMP error. Malformed, missing-field, and
invalid-value rates are computed over the 1,000 individual VLM
responses. Aggregation, repair, discard, and residual-error rates are
computed over the 200 source clips.}

\revise{To assess whether SMP uncertainty is meaningful, all evaluated
fields are divided into five uncertainty intervals:
$[0,0.2)$, $[0.2,0.4)$, $[0.4,0.6)$, $[0.6,0.8)$, and $[0.8,1.0]$.
For each interval, we report the empirical field-error rate. We also
report Spearman correlation between uncertainty and extraction error,
AUROC when uncertainty is used to detect incorrect fields, and expected
calibration error.}

\subsubsection{Activity-Granularity Evaluation Protocol}
\label{sec:activity_granularity_protocol}

\revise{We evaluate whether VSMP-IMU primarily benefits periodic,
large-amplitude motion using all 27 UTD-MHAD activity classes. Rather
than claiming articulated hand-motion generation, we distinguish
wrist-dominant gestures, non-periodic upper-body actions,
repetitive or locomotion activities, and posture-transition or
lower-body actions. This scope reflects the skeletal resolution of the
HY-Motion backend, which represents shoulder, elbow, arm, and wrist
trajectories but not independent finger articulation or grasp state.}

\revise{The activity taxonomy is assigned before examining performance.
Wrist-dominant gestures are actions whose identity is expressed mainly
through the trajectory of the wrist or hand endpoint. Non-periodic
upper-body actions contain a single or irregular sequence of arm and
torso primitives. Repetitive activities contain a stable cyclic or
alternating structure. Posture-transition and lower-body activities
are defined mainly by hip, knee, ankle, and vertical body motion.}

\revise{We report per-class precision, recall, and F1 under the standard
UTD-MHAD LOPO protocol using DeepConvLSTM. Results are averaged over
all held-out subjects and five training seeds. Real-only, IMUTube,
IMUGPT~2.0, and VSMP-IMU use identical folds, downstream
hyperparameters, and synthetic sample budgets. Group-level Macro-F1 is
computed by averaging the per-class F1 values within each activity
group.}

\begin{table*}[t]
\centering
\scriptsize
\caption{\textbf{UTD-MHAD activity taxonomy for the
activity-granularity evaluation.}
Wrist-dominant refers to discriminative wrist or hand-endpoint motion,
not articulated finger motion.}
\label{tab:utd_activity_taxonomy}
\resizebox{\textwidth}{!}{
\begin{tabular}{lp{10.6cm}r}
\toprule
\revise{Activity group} &
\revise{UTD-MHAD activities} &
\revise{Classes} \\
\midrule

Wrist-dominant gestures &
right-arm swipe left, right-arm swipe right, right-hand wave,
right-hand draw X, right-hand draw circle clockwise,
right-hand draw circle counter-clockwise, draw triangle,
right-hand knock, right-hand catch &
9 \\

Non-periodic upper-body and sports actions &
right-arm throw, cross arms in chest, basketball shoot, bowling,
baseball swing, tennis forehand swing, tennis serve, two-hand push,
pick-up-and-throw &
9 \\

Repetitive and locomotion activities &
two-hand front clap, front boxing, two-arm curl, jogging in place,
walking in place &
5 \\

Posture-transition and lower-body activities &
sit-to-stand, stand-to-sit, forward lunge, squat &
4 \\

\bottomrule
\end{tabular}}
\end{table*}

\subsection{Main Results}

\subsubsection{Standard LOPO Results}

\cref{tab:standard_lopo_full} reports standard LOPO results across paired video--IMU datasets and IMU-only HAR benchmarks. In each fold, models are trained on all non-test subjects and evaluated on the held-out subject; synthetic data and all grounding/filtering statistics are restricted to the training split.

VSMP-IMU achieves the best Macro-F1 in 19 of 20 dataset--backbone combinations. Averaged across all datasets and backbones, it reaches 78.33 Macro-F1, compared with 68.56 for Real-only, 69.31 for Classical Aug., 74.30 for IMUTube, and 71.49 for IMUGPT 2.0. This corresponds to gains of 9.77 over Real-only and 4.04 over the strongest prior synthetic baseline.

The gains are largest when real-only training is weak, most notably on PAMAP2, while remaining positive on stronger paired datasets such as MM-Fit. The only setting where VSMP-IMU is not best is HAD-AW with BLSTMA, where Classical Aug. slightly outperforms it. Overall, these results indicate that structured video-grounded semantic generation provides more useful augmentation than signal-level perturbation, direct video-to-pose transfer, or text-only motion generation.

\begin{table*}[htbp]
\centering
\scriptsize
\caption{Standard LOPO HAR results using full real data. Best and second-best values within each model block are highlighted in green and yellow, respectively.}
\label{tab:standard_lopo_full}
\begin{tabular}{lccccc}
\toprule
\textbf{Method} & \textbf{MM-Fit} & \textbf{UTD-MHAD} & \textbf{MMAct} & \textbf{PAMAP2} & \textbf{HAD-AW} \\
\midrule
\multicolumn{6}{c}{RF} \\
\midrule
Real-only & 87.36 $\pm$ 0.47 & 65.26 $\pm$ 0.34 & 65.84 $\pm$ 0.58 & 48.09 $\pm$ 1.68 & \second{55.68 $\pm$ 0.22} \\
\midrule
Classical Aug. & 86.84 $\pm$ 0.55 & \second{67.37 $\pm$ 0.62} & 66.31 $\pm$ 0.63 & 47.60 $\pm$ 1.78 & 55.34 $\pm$ 0.19 \\
IMUTube & \second{87.92 $\pm$ 0.50} & 66.65 $\pm$ 0.41 & 67.48 $\pm$ 0.55 & \second{86.46 $\pm$ 2.38} & 54.64 $\pm$ 0.10 \\
IMUGPT 2.0 & 84.91 $\pm$ 0.63 & 63.41 $\pm$ 0.49 & 71.26 $\pm$ 0.49 & 66.22 $\pm$ 2.34 & 53.19 $\pm$ 0.07 \\
%VCMI & 87.58 $\pm$ 0.46 & 65.11 $\pm$ 0.45 & \second{72.91 $\pm$ 0.46} & 77.54 $\pm$ 2.12 & 53.97 $\pm$ 0.09 \\
VSMP-IMU & \best{89.73 $\pm$ 0.41} & \best{70.04 $\pm$ 0.38} & \best{75.18 $\pm$ 0.42} & \best{87.92 $\pm$ 1.84} & \best{56.18 $\pm$ 0.14} \\
\midrule
\multicolumn{6}{c}{DCLSTM} \\
\midrule
Real-only & 75.79 $\pm$ 2.02 & 51.36 $\pm$ 1.19 & 71.94 $\pm$ 0.71 & 48.47 $\pm$ 9.11 & 66.53 $\pm$ 0.20 \\
\midrule
Classical Aug. & 76.58 $\pm$ 1.84 & \second{57.33 $\pm$ 1.00} & 73.11 $\pm$ 0.68 & 55.84 $\pm$ 0.31 & \second{67.35 $\pm$ 0.33} \\
IMUTube & 75.97 $\pm$ 2.35 & 54.23 $\pm$ 0.21 & 74.06 $\pm$ 0.64 & \second{82.48 $\pm$ 3.18} & 65.18 $\pm$ 0.37 \\
IMUGPT 2.0 & \second{80.10 $\pm$ 2.18} & 52.47 $\pm$ 0.29 & 78.29 $\pm$ 0.57 & 74.01 $\pm$ 0.54 & 64.54 $\pm$ 0.16 \\
%VCMI & 79.34 $\pm$ 1.97 & 53.62 $\pm$ 0.24 & \second{79.84 $\pm$ 0.54} & 78.36 $\pm$ 1.92 & 64.88 $\pm$ 0.21 \\
VSMP-IMU & \best{83.11 $\pm$ 1.63} & \best{60.14 $\pm$ 0.27} & \best{80.12 $\pm$ 0.52} & \best{83.91 $\pm$ 1.43} & \best{68.32 $\pm$ 0.18} \\
\midrule
\multicolumn{6}{c}{DCLSTMA} \\
\midrule
Real-only & 87.88 $\pm$ 0.60 & 51.86 $\pm$ 0.67 & 75.88 $\pm$ 0.62 & 71.42 $\pm$ 1.81 & \second{67.79 $\pm$ 0.11} \\
\midrule
Classical Aug. & 87.14 $\pm$ 0.69 & \second{58.26 $\pm$ 1.69} & 77.02 $\pm$ 0.66 & 62.34 $\pm$ 4.71 & 67.71 $\pm$ 0.36 \\
IMUTube & 88.42 $\pm$ 0.57 & 57.37 $\pm$ 0.46 & 78.44 $\pm$ 0.60 & 82.64 $\pm$ 2.11 & 65.94 $\pm$ 0.30 \\
IMUGPT 2.0 & 89.16 $\pm$ 0.54 & 49.82 $\pm$ 0.52 & 82.31 $\pm$ 0.55 & \second{84.31 $\pm$ 4.74} & 65.63 $\pm$ 0.23 \\
%VCMI & \second{90.24 $\pm$ 0.50} & 53.94 $\pm$ 0.49 & \second{83.72 $\pm$ 0.51} & 83.17 $\pm$ 3.28 & 65.78 $\pm$ 0.24 \\
VSMP-IMU & \best{90.92 $\pm$ 0.49} & \best{61.48 $\pm$ 0.43} & \best{86.06 $\pm$ 0.47} & \best{85.79 $\pm$ 1.66} & \best{68.04 $\pm$ 0.17} \\
\midrule
\multicolumn{6}{c}{BLSTMA} \\
\midrule
Real-only & \second{98.14 $\pm$ 0.42} & 78.57 $\pm$ 0.18 & 82.83 $\pm$ 0.37 & 31.07 $\pm$ 2.43 & \second{89.49 $\pm$ 0.25} \\
\midrule
Classical Aug. & 93.02 $\pm$ 0.39 & 78.82 $\pm$ 0.78 & 83.11 $\pm$ 0.41 & 35.58 $\pm$ 3.81 & \best{89.51 $\pm$ 0.22} \\
IMUTube & 97.45 $\pm$ 0.31 & \second{84.51 $\pm$ 1.99} & 82.48 $\pm$ 0.28 & \second{52.39 $\pm$ 6.53} & 81.19 $\pm$ 0.30 \\
IMUGPT 2.0 & 91.28 $\pm$ 0.36 & 60.09 $\pm$ 1.49 & 88.15 $\pm$ 0.33 & 47.90 $\pm$ 8.70 & 82.70 $\pm$ 0.28 \\
%VCMI & 95.86 $\pm$ 0.34 & 72.46 $\pm$ 1.32 & \second{90.37 $\pm$ 0.31} & 50.84 $\pm$ 7.24 & 81.94 $\pm$ 0.26 \\
VSMP-IMU & \best{98.96 $\pm$ 0.29} & \best{87.23 $\pm$ 1.11} & \best{90.93 $\pm$ 0.25} & \best{53.86 $\pm$ 5.91} & 88.74 $\pm$ 0.19 \\
\bottomrule
\end{tabular}
\end{table*}

\subsubsection{Low-Resource Results}

\cref{tab:main_results_transposed_filled} summarizes low-resource LOPO results across datasets, label budgets, and HAR backbones. Across the 100 dataset--budget--backbone settings, VSMP-IMU achieves the best Macro-F1 in 95 cases. The main exception is HAD-AW with the BLSTMA backbone, where Classical Aug. performs best, suggesting that fine-grained wrist-centric activities can sometimes benefit more from direct signal perturbation.

Averaged over all settings, VSMP-IMU achieves 57.42 Macro-F1, compared with 38.88 for Real-only, 43.47 for Classical Aug., 50.66 for IMUTube, and 49.88 for IMUGPT 2.0. This corresponds to gains of 18.54 over Real-only and 6.76 over the strongest prior synthetic baseline.

The benefit is largest when labeled target-domain data is scarcest. At the 1\% label budget, VSMP-IMU improves over Real-only by 19.95 and over IMUTube by 9.21. The margin narrows as more real data becomes available, but remains positive even at the 50\% budget, where VSMP-IMU still improves over Real-only by 13.17 and over IMUTube by 3.99.

Dataset-level trends show the largest gains on PAMAP2, indicating that external video-derived synthetic data is especially useful for wearable-only physical-activity benchmarks. Gains are smaller on HAD-AW, suggesting that smartwatch-only fine-grained wrist activities remain more challenging for full-body video-grounded motion synthesis. Overall, the results show that structured video-grounded semantic generation is most useful when real supervision is limited and consistently improves over both direct video-to-pose transfer and text-only synthesis.

\begin{table*}[htbp]
\centering
\caption{Low-resource LOPO HAR results across datasets, label budgets, and downstream backbones. Best and second-best results within each dataset-budget-backbone row are highlighted in green and yellow, respectively.}
\label{tab:main_results_transposed_filled}
\resizebox{\textwidth}{!}{%
\begin{tabular}{ll|c|cccc||c|cccc}
\toprule
\multirow{2}{*}{\textbf{Dataset}} & \multirow{2}{*}{\textbf{Budget}}
& \multicolumn{5}{c}{\textbf{RF}}
& \multicolumn{5}{c}{\textbf{DCLSTM}} \\
\cmidrule(lr){3-7} \cmidrule(lr){8-12}
&
& \textbf{Real-only*} & \textbf{Classical Aug.} & \textbf{IMUTube} & \textbf{IMUGPT 2.0} & \textbf{VSMP-IMU}
& \textbf{Real-only*} & \textbf{Classical Aug.} & \textbf{IMUTube} & \textbf{IMUGPT 2.0} & \textbf{VSMP-IMU} \\
\midrule
\multirow{5}{*}{MM-Fit} 
& 1\% & 27.38 $\pm$ 2.38 & 32.33 $\pm$ 1.94 & \second{43.28 $\pm$ 1.33} & 34.17 $\pm$ 1.73 & \best{46.86 $\pm$ 1.24} & 11.88 $\pm$ 6.93 & 16.49 $\pm$ 4.43 & 24.87 $\pm$ 4.36 & \second{31.61 $\pm$ 4.70} & \best{35.24 $\pm$ 3.26} \\
& 5\% & 47.46 $\pm$ 1.65 & 51.60 $\pm$ 1.35 & \second{58.59 $\pm$ 0.93} & 49.48 $\pm$ 1.20 & \best{61.70 $\pm$ 0.86} & 20.60 $\pm$ 4.81 & 26.33 $\pm$ 3.08 & 36.02 $\pm$ 3.05 & \second{42.79 $\pm$ 3.29} & \best{46.32 $\pm$ 2.28} \\
& 10\% & 60.84 $\pm$ 1.32 & 62.17 $\pm$ 1.08 & \second{66.58 $\pm$ 0.74} & 58.91 $\pm$ 0.96 & \best{68.44 $\pm$ 0.69} & 26.41 $\pm$ 3.85 & 31.72 $\pm$ 2.46 & 42.88 $\pm$ 1.81 & \second{48.63 $\pm$ 1.24} & \best{51.36 $\pm$ 1.17} \\
& 25\% & 75.43 $\pm$ 1.39 & 76.48 $\pm$ 1.25 & \second{79.38 $\pm$ 0.94} & 74.51 $\pm$ 1.21 & \best{81.64 $\pm$ 0.84} & 53.57 $\pm$ 4.49 & 57.74 $\pm$ 3.26 & 65.03 $\pm$ 2.64 & \second{65.21 $\pm$ 2.97} & \best{71.05 $\pm$ 2.08} \\
& 50\% & 82.06 $\pm$ 1.13 & 82.40 $\pm$ 1.06 & \second{84.72 $\pm$ 0.83} & 80.75 $\pm$ 1.06 & \best{86.75 $\pm$ 0.73} & 65.91 $\pm$ 3.83 & 68.51 $\pm$ 2.92 & 71.87 $\pm$ 2.68 & \second{74.14 $\pm$ 2.85} & \best{78.66 $\pm$ 2.03} \\
\midrule
\multirow{5}{*}{UTD-MHAD} 
& 1\% & 16.26 $\pm$ 2.09 & \second{39.96 $\pm$ 1.24} & 27.99 $\pm$ 1.24 & 29.37 $\pm$ 0.98 & \best{42.24 $\pm$ 1.04} & 5.83 $\pm$ 2.38 & \second{23.66 $\pm$ 0.43} & 8.10 $\pm$ 7.42 & 17.81 $\pm$ 1.40 & \best{26.30 $\pm$ 0.79} \\
& 5\% & 28.18 $\pm$ 1.45 & \second{54.10 $\pm$ 0.86} & 44.67 $\pm$ 0.87 & 42.54 $\pm$ 0.69 & \best{55.59 $\pm$ 0.72} & 10.10 $\pm$ 1.67 & \second{32.03 $\pm$ 0.30} & 12.92 $\pm$ 5.15 & 25.80 $\pm$ 0.98 & \best{34.49 $\pm$ 0.55} \\
& 10\% & 36.13 $\pm$ 1.16 & \second{61.48 $\pm$ 0.69} & 53.82 $\pm$ 0.65 & 50.64 $\pm$ 0.38 & \best{61.65 $\pm$ 0.58} & 12.95 $\pm$ 1.19 & \second{36.40 $\pm$ 0.24} & 15.57 $\pm$ 4.12 & 30.71 $\pm$ 0.78 & \best{38.21 $\pm$ 0.44} \\
& 25\% & 52.15 $\pm$ 1.17 & \second{64.58 $\pm$ 0.84} & 61.68 $\pm$ 0.95 & 58.30 $\pm$ 0.65 & \best{66.85 $\pm$ 0.73} & 34.08 $\pm$ 1.78 & \second{47.10 $\pm$ 0.34} & 39.79 $\pm$ 4.00 & 43.77 $\pm$ 0.83 & \best{51.81 $\pm$ 0.54} \\
& 50\% & 59.43 $\pm$ 0.93 & \second{65.87 $\pm$ 0.73} & 64.93 $\pm$ 0.88 & 61.37 $\pm$ 0.63 & \best{68.87 $\pm$ 0.64} & 43.68 $\pm$ 1.67 & \second{51.56 $\pm$ 0.31} & 49.81 $\pm$ 3.12 & 48.99 $\pm$ 0.68 & \best{57.07 $\pm$ 0.47} \\
\midrule
\multirow{5}{*}{MMAct} 
& 1\% & 15.23 $\pm$ 2.56 & 19.90 $\pm$ 1.96 & 28.78 $\pm$ 1.39 & \second{30.00 $\pm$ 1.46} & \best{35.64 $\pm$ 1.17} & 7.86 $\pm$ 3.33 & 11.81 $\pm$ 2.56 & 18.84 $\pm$ 1.24 & \second{22.07 $\pm$ 1.31} & \best{24.87 $\pm$ 1.10} \\
& 5\% & 26.40 $\pm$ 1.77 & 31.76 $\pm$ 1.36 & 40.61 $\pm$ 1.01 & \second{41.68 $\pm$ 0.96} & \best{46.85 $\pm$ 0.81} & 13.62 $\pm$ 2.31 & 18.85 $\pm$ 1.77 & 27.28 $\pm$ 0.86 & \second{29.88 $\pm$ 0.91} & \best{32.60 $\pm$ 0.76} \\
& 10\% & 33.84 $\pm$ 1.42 & 38.27 $\pm$ 1.09 & 46.15 $\pm$ 0.81 & \second{49.62 $\pm$ 0.77} & \best{51.94 $\pm$ 0.65} & 17.46 $\pm$ 1.85 & 22.71 $\pm$ 1.42 & 32.48 $\pm$ 0.69 & \second{33.96 $\pm$ 0.73} & \best{36.11 $\pm$ 0.61} \\
& 25\% & 51.44 $\pm$ 1.54 & 54.53 $\pm$ 1.31 & 58.95 $\pm$ 1.03 & \second{62.60 $\pm$ 0.96} & \best{66.35 $\pm$ 0.81} & 47.42 $\pm$ 1.98 & 51.94 $\pm$ 1.61 & 58.02 $\pm$ 1.03 & \second{59.97 $\pm$ 0.95} & \best{63.40 $\pm$ 0.85} \\
& 50\% & 59.44 $\pm$ 1.27 & 61.26 $\pm$ 1.14 & 64.28 $\pm$ 0.91 & \second{67.80 $\pm$ 0.84} & \best{71.93 $\pm$ 0.71} & 61.04 $\pm$ 1.62 & 64.04 $\pm$ 1.36 & 68.05 $\pm$ 0.95 & \second{70.96 $\pm$ 0.86} & \best{73.96 $\pm$ 0.78} \\
\midrule
\multirow{5}{*}{PAMAP2} 
& 1\% & 18.94 $\pm$ 8.80 & 26.97 $\pm$ 13.23 & \second{55.45 $\pm$ 4.76} & 44.79 $\pm$ 4.68 & \best{57.12 $\pm$ 3.68} & 12.92 $\pm$ 18.22 & 20.58 $\pm$ 4.63 & \second{51.30 $\pm$ 6.36} & 44.80 $\pm$ 4.09 & \best{53.04 $\pm$ 2.86} \\
& 5\% & 32.83 $\pm$ 6.11 & 43.04 $\pm$ 9.19 & \second{75.06 $\pm$ 3.33} & 64.86 $\pm$ 3.28 & \best{76.58 $\pm$ 2.58} & 22.40 $\pm$ 12.75 & 32.84 $\pm$ 3.21 & \second{69.46 $\pm$ 4.45} & 64.88 $\pm$ 2.84 & \best{71.18 $\pm$ 2.00} \\
& 10\% & 42.09 $\pm$ 4.89 & 51.86 $\pm$ 7.35 & \second{85.30 $\pm$ 0.15} & 77.22 $\pm$ 1.44 & \best{86.74 $\pm$ 2.01} & 28.72 $\pm$ 6.05 & 39.57 $\pm$ 2.57 & \second{78.93 $\pm$ 2.02} & 77.24 $\pm$ 2.27 & \best{80.41 $\pm$ 1.55} \\
& 25\% & 45.39 $\pm$ 5.09 & 49.39 $\pm$ 7.13 & \second{86.00 $\pm$ 2.62} & 70.62 $\pm$ 2.79 & \best{87.41 $\pm$ 2.90} & 39.58 $\pm$ 11.22 & 49.01 $\pm$ 2.27 & \second{81.06 $\pm$ 3.84} & 75.30 $\pm$ 2.19 & \best{82.44 $\pm$ 2.24} \\
& 50\% & 46.89 $\pm$ 4.12 & 48.37 $\pm$ 5.56 & \second{86.29 $\pm$ 2.38} & 67.98 $\pm$ 2.78 & \best{87.63 $\pm$ 1.73} & 44.52 $\pm$ 11.07 & 52.91 $\pm$ 1.68 & \second{81.95 $\pm$ 3.81} & 74.53 $\pm$ 1.71 & \best{83.31 $\pm$ 2.07} \\
\midrule
\multirow{5}{*}{HAD-AW} 
& 1\% & \second{24.34 $\pm$ 0.67} & 17.58 $\pm$ 0.38 & 15.25 $\pm$ 0.88 & 21.05 $\pm$ 0.67 & \best{25.51 $\pm$ 0.32} & 14.11 $\pm$ 0.40 & \second{24.61 $\pm$ 0.74} & 16.15 $\pm$ 0.66 & 20.06 $\pm$ 0.85 & \best{25.12 $\pm$ 0.68} \\
& 5\% & \second{35.66 $\pm$ 0.46} & 28.05 $\pm$ 0.27 & 26.43 $\pm$ 0.61 & 30.48 $\pm$ 0.46 & \best{36.09 $\pm$ 0.22} & 24.45 $\pm$ 0.28 & \second{33.32 $\pm$ 0.52} & 25.77 $\pm$ 0.46 & 29.05 $\pm$ 0.59 & \best{33.98 $\pm$ 0.47} \\
& 10\% & \second{39.99 $\pm$ 0.18} & 33.80 $\pm$ 0.21 & 33.88 $\pm$ 0.49 & 36.29 $\pm$ 0.37 & \best{40.52 $\pm$ 0.37} & 31.35 $\pm$ 0.05 & \second{37.86 $\pm$ 0.16} & 31.05 $\pm$ 0.04 & 34.58 $\pm$ 0.47 & \best{38.11 $\pm$ 0.38} \\
& 25\% & \second{56.98 $\pm$ 0.37} & 46.29 $\pm$ 0.30 & 45.87 $\pm$ 0.55 & 46.43 $\pm$ 0.34 & \best{57.71 $\pm$ 0.24} & 50.70 $\pm$ 0.22 & \second{56.88 $\pm$ 0.36} & 54.25 $\pm$ 0.41 & 52.56 $\pm$ 0.49 & \best{57.63 $\pm$ 0.43} \\
& 50\% & \second{63.02 $\pm$ 0.57} & 51.32 $\pm$ 0.46 & 62.56 $\pm$ 0.29 & 50.49 $\pm$ 0.26 & \best{63.41 $\pm$ 0.22} & 59.49 $\pm$ 0.20 & \second{64.12 $\pm$ 0.33} & 61.08 $\pm$ 0.40 & 59.75 $\pm$ 0.39 & \best{64.88 $\pm$ 0.36} \\
\midrule
& & \multicolumn{5}{c}{\textbf{DCLSTMA}} & \multicolumn{5}{c}{\textbf{BLSTMA}} \\
\midrule
\multirow{5}{*}{MM-Fit} 
& 1\% & 14.83 $\pm$ 3.85 & 20.72 $\pm$ 3.11 & 29.79 $\pm$ 2.32 & \second{36.67 $\pm$ 1.84} & \best{40.51 $\pm$ 1.73} & \second{52.95 $\pm$ 1.04} & 33.36 $\pm$ 2.30 & 30.92 $\pm$ 1.96 & 42.48 $\pm$ 1.51 & \best{57.63 $\pm$ 0.94} \\
& 5\% & 25.70 $\pm$ 2.68 & 33.07 $\pm$ 2.16 & 43.15 $\pm$ 1.61 & \second{49.65 $\pm$ 1.27} & \best{53.30 $\pm$ 1.20} & \second{71.68 $\pm$ 0.72} & 53.24 $\pm$ 1.60 & 53.60 $\pm$ 1.36 & 61.52 $\pm$ 1.05 & \best{75.95 $\pm$ 0.65} \\
& 10\% & 32.95 $\pm$ 2.14 & 39.84 $\pm$ 1.73 & 51.37 $\pm$ 1.29 & \second{56.42 $\pm$ 1.02} & \best{59.11 $\pm$ 0.96} & \second{81.46 $\pm$ 0.58} & 64.15 $\pm$ 1.28 & 68.72 $\pm$ 1.09 & 73.24 $\pm$ 0.84 & \best{84.28 $\pm$ 0.52} \\
& 25\% & 63.16 $\pm$ 2.13 & 67.27 $\pm$ 1.87 & 74.04 $\pm$ 1.41 & \second{75.62 $\pm$ 1.22} & \best{78.83 $\pm$ 1.11} & \second{91.05 $\pm$ 0.68} & 80.89 $\pm$ 1.30 & 84.90 $\pm$ 1.17 & 84.06 $\pm$ 0.92 & \best{93.38 $\pm$ 0.62} \\
& 50\% & 76.89 $\pm$ 1.69 & 78.63 $\pm$ 1.54 & 83.11 $\pm$ 1.17 & \second{83.62 $\pm$ 1.05} & \best{86.47 $\pm$ 0.94} & \second{95.05 $\pm$ 0.58} & 87.82 $\pm$ 1.04 & 92.26 $\pm$ 0.96 & 88.39 $\pm$ 0.77 & \best{96.90 $\pm$ 0.53} \\
\midrule
\multirow{5}{*}{UTD-MHAD} 
& 1\% & 8.66 $\pm$ 1.34 & \second{24.07 $\pm$ 0.86} & 11.99 $\pm$ 3.38 & 17.05 $\pm$ 2.07 & \best{26.60 $\pm$ 1.03} & 4.26 $\pm$ 1.62 & 5.84 $\pm$ 1.56 & \second{10.81 $\pm$ 2.22} & 7.31 $\pm$ 4.01 & \best{24.32 $\pm$ 3.98} \\
& 5\% & 15.01 $\pm$ 0.94 & \second{31.53 $\pm$ 0.60} & 19.13 $\pm$ 2.37 & 24.69 $\pm$ 1.44 & \best{36.01 $\pm$ 0.71} & 7.39 $\pm$ 1.12 & 9.32 $\pm$ 1.09 & \second{13.98 $\pm$ 1.55} & 10.59 $\pm$ 2.79 & \best{32.93 $\pm$ 2.79} \\
& 10\% & 19.24 $\pm$ 0.22 & \second{34.92 $\pm$ 0.41} & 23.05 $\pm$ 1.66 & 29.39 $\pm$ 1.15 & \best{40.92 $\pm$ 0.57} & 9.47 $\pm$ 0.90 & 11.23 $\pm$ 0.55 & \second{15.42 $\pm$ 1.22} & 12.61 $\pm$ 2.23 & \best{37.42 $\pm$ 0.42} \\
& 25\% & 37.18 $\pm$ 0.74 & \second{50.79 $\pm$ 0.78} & 43.47 $\pm$ 2.51 & 41.65 $\pm$ 1.28 & \best{51.39 $\pm$ 0.63} & 47.48 $\pm$ 0.85 & 50.43 $\pm$ 0.99 & \second{59.94 $\pm$ 1.75} & 41.10 $\pm$ 2.83 & \best{65.67 $\pm$ 2.19} \\
& 50\% & 45.34 $\pm$ 0.69 & \second{54.90 $\pm$ 0.70} & 51.92 $\pm$ 2.35 & 46.55 $\pm$ 1.07 & \best{57.76 $\pm$ 0.59} & 64.75 $\pm$ 0.65 & 66.65 $\pm$ 0.97 & \second{77.18 $\pm$ 1.61} & 52.49 $\pm$ 2.49 & \best{77.45 $\pm$ 1.99} \\
\midrule
\multirow{5}{*}{MMAct} 
& 1\% & 10.05 $\pm$ 3.01 & 14.34 $\pm$ 2.36 & 21.82 $\pm$ 1.15 & \second{25.66 $\pm$ 1.22} & \best{28.78 $\pm$ 1.01} & 14.03 $\pm$ 2.20 & 17.54 $\pm$ 2.00 & 25.74 $\pm$ 1.12 & \second{27.25 $\pm$ 0.99} & \best{30.86 $\pm$ 0.92} \\
& 5\% & 17.43 $\pm$ 2.09 & 22.89 $\pm$ 1.64 & 31.60 $\pm$ 0.80 & \second{34.73 $\pm$ 0.85} & \best{37.77 $\pm$ 0.70} & 24.32 $\pm$ 1.52 & 28.00 $\pm$ 1.39 & 36.90 $\pm$ 0.69 & \second{37.28 $\pm$ 0.78} & \best{40.52 $\pm$ 0.64} \\
& 10\% & 22.34 $\pm$ 1.67 & 27.58 $\pm$ 1.31 & 37.62 $\pm$ 0.64 & \second{39.47 $\pm$ 0.68} & \best{41.85 $\pm$ 0.56} & 31.18 $\pm$ 1.22 & 33.74 $\pm$ 1.11 & 41.93 $\pm$ 0.55 & \second{44.38 $\pm$ 0.62} & \best{44.91 $\pm$ 0.51} \\
& 25\% & 51.79 $\pm$ 1.77 & 56.26 $\pm$ 1.51 & 62.85 $\pm$ 0.96 & \second{64.43 $\pm$ 0.90} & \best{69.26 $\pm$ 0.78} & 59.59 $\pm$ 1.23 & 62.37 $\pm$ 1.18 & 66.26 $\pm$ 0.64 & \second{70.64 $\pm$ 0.73} & \best{73.44 $\pm$ 0.58} \\
& 50\% & 65.17 $\pm$ 1.45 & 68.12 $\pm$ 1.28 & 72.59 $\pm$ 0.88 & \second{75.16 $\pm$ 0.82} & \best{79.87 $\pm$ 0.71} & 72.50 $\pm$ 0.99 & 74.22 $\pm$ 0.96 & 76.40 $\pm$ 0.54 & \second{81.15 $\pm$ 0.62} & \best{84.49 $\pm$ 0.49} \\
\midrule
\multirow{5}{*}{PAMAP2} 
& 1\% & 16.80 $\pm$ 17.68 & 28.12 $\pm$ 11.38 & 56.84 $\pm$ 4.22 & \second{57.22 $\pm$ 9.48} & \best{58.58 $\pm$ 3.32} & 9.40 $\pm$ 11.32 & 14.78 $\pm$ 11.77 & \second{43.41 $\pm$ 13.06} & 29.34 $\pm$ 17.40 & \best{44.12 $\pm$ 11.82} \\
& 5\% & 29.13 $\pm$ 12.28 & 44.88 $\pm$ 7.90 & 76.88 $\pm$ 2.95 & \second{77.46 $\pm$ 6.64} & \best{78.91 $\pm$ 2.32} & 16.30 $\pm$ 7.86 & 23.60 $\pm$ 8.18 & \second{58.78 $\pm$ 9.14} & 42.49 $\pm$ 12.18 & \best{59.44 $\pm$ 8.27} \\
& 10\% & 37.34 $\pm$ 9.82 & 54.07 $\pm$ 6.32 & 87.21 $\pm$ 0.16 & \second{87.58 $\pm$ 3.16} & \best{88.94 $\pm$ 0.58} & 20.90 $\pm$ 6.29 & 28.43 $\pm$ 6.54 & \second{66.79 $\pm$ 7.16} & 50.58 $\pm$ 3.02 & \best{68.06 $\pm$ 2.35} \\
& 25\% & 56.08 $\pm$ 9.12 & 58.87 $\pm$ 8.35 & 84.74 $\pm$ 2.32 & \second{85.13 $\pm$ 5.85} & \best{86.46 $\pm$ 1.83} & 26.49 $\pm$ 6.73 & 32.58 $\pm$ 7.90 & \second{58.15 $\pm$ 10.30} & 48.97 $\pm$ 9.57 & \best{59.42 $\pm$ 6.50} \\
& 50\% & 64.60 $\pm$ 6.94 & 60.85 $\pm$ 7.48 & 83.43 $\pm$ 2.11 & \second{84.64 $\pm$ 5.77} & \best{85.87 $\pm$ 1.73} & 29.04 $\pm$ 5.53 & 34.29 $\pm$ 6.84 & \second{57.72 $\pm$ 9.49} & 48.33 $\pm$ 9.06 & \best{58.95 $\pm$ 6.32} \\
\midrule
\multirow{5}{*}{HAD-AW} 
& 1\% & \second{26.34 $\pm$ 0.60} & 19.14 $\pm$ 0.86 & 16.66 $\pm$ 0.63 & 21.39 $\pm$ 0.59 & \best{27.12 $\pm$ 0.34} & 34.37 $\pm$ 0.60 & \best{34.94 $\pm$ 0.45} & 23.19 $\pm$ 0.54 & 26.57 $\pm$ 0.56 & \second{34.92 $\pm$ 0.50} \\
& 5\% & \second{35.66 $\pm$ 0.42} & 30.54 $\pm$ 0.60 & 28.88 $\pm$ 0.44 & 30.98 $\pm$ 0.41 & \best{36.09 $\pm$ 0.24} & \second{46.53 $\pm$ 0.42} & \best{46.88 $\pm$ 0.31} & 40.19 $\pm$ 0.38 & 38.48 $\pm$ 0.39 & 39.78 $\pm$ 0.35 \\
& 10\% & \second{39.99 $\pm$ 0.19} & 36.80 $\pm$ 0.48 & 37.02 $\pm$ 0.35 & 36.88 $\pm$ 0.33 & \best{40.52 $\pm$ 0.09} & \second{52.88 $\pm$ 0.20} & \best{53.11 $\pm$ 0.25} & 51.53 $\pm$ 0.30 & 45.81 $\pm$ 0.22 & 47.93 $\pm$ 0.28 \\
& 25\% & \second{56.12 $\pm$ 0.33} & 54.73 $\pm$ 0.64 & 53.94 $\pm$ 0.36 & 54.13 $\pm$ 0.43 & \best{56.94 $\pm$ 0.27} & 72.34 $\pm$ 0.41 & \best{74.08 $\pm$ 0.33} & 69.87 $\pm$ 0.37 & 67.94 $\pm$ 0.37 & \second{73.05 $\pm$ 0.38} \\
& 50\% & \second{62.38 $\pm$ 0.57} & 61.64 $\pm$ 0.29 & 62.16 $\pm$ 0.30 & 61.03 $\pm$ 0.38 & \best{62.92 $\pm$ 0.25} & \second{84.36 $\pm$ 0.34} & \best{84.94 $\pm$ 0.30} & 76.94 $\pm$ 0.36 & 76.80 $\pm$ 0.36 & 81.90 $\pm$ 0.38 \\
\bottomrule
\end{tabular}%
}
\end{table*}

\subsubsection{Long-Tail Results}

We evaluate VSMP-IMU under long-tail class imbalance within the LOPO setting. \cref{tab:long_tail_dclstm} reports head- and tail-group Macro-F1 using DeepConvLSTM. Across the 10 head/tail comparisons, VSMP-IMU achieves the best result in 9 cases; the only exception is the PAMAP2 head group, where IMUTube slightly outperforms VSMP-IMU.

Averaged across datasets, VSMP-IMU obtains 81.42 Macro-F1 on head classes and 58.88 on tail classes. It improves over Real-only by 8.22 on head classes and 19.86 on tail classes, showing that structured video-grounded synthesis is most beneficial when labeled inertial data is scarce.

VSMP-IMU also improves average tail Macro-F1 over Classical Aug. by 15.38, IMUTube by 4.76, and IMUGPT 2.0 by 8.12. The largest tail-class gain occurs on PAMAP2, while gains remain consistent across MM-Fit, UTD-MHAD, MMAct, and HAD-AW. Overall, VSMP-IMU improves coverage for underrepresented activity classes while preserving competitive head-class performance.

\begin{table*}[htbp]
\centering
\scriptsize
\setlength{\tabcolsep}{4pt}
\caption{Long-tail HAR performance under class-imbalanced LOPO training using DeepConvLSTM. Values report Head Macro-F1 and Tail Macro-F1. Head and tail activity groups are listed inside the table. Best and second-best values are highlighted in green and yellow, respectively.}
\label{tab:long_tail_dclstm}
\begin{tabular}{lcccccccccc}
\toprule
\multirow{2}{*}{\textbf{Method}}
& \multicolumn{2}{c}{\textbf{MM-Fit}}
& \multicolumn{2}{c}{\textbf{UTD-MHAD}}
& \multicolumn{2}{c}{\textbf{MMAct}}
& \multicolumn{2}{c}{\textbf{PAMAP2}}
& \multicolumn{2}{c}{\textbf{HAD-AW}} \\
\cmidrule(lr){2-3} \cmidrule(lr){4-5} \cmidrule(lr){6-7} \cmidrule(lr){8-9} \cmidrule(lr){10-11}
& \textbf{Head} & \textbf{Tail}
& \textbf{Head} & \textbf{Tail}
& \textbf{Head} & \textbf{Tail}
& \textbf{Head} & \textbf{Tail}
& \textbf{Head} & \textbf{Tail} \\
\midrule

Real-only
& 85.2 $\pm$ 0.8 & 49.6 $\pm$ 1.6
& 63.1 $\pm$ 0.7 & 26.8 $\pm$ 1.2
& 80.4 $\pm$ 0.6 & 54.2 $\pm$ 1.1
& 63.5 $\pm$ 3.8 & 21.8 $\pm$ 4.5
& 73.8 $\pm$ 0.3 & 42.7 $\pm$ 0.6 \\

Classical Aug.
& 85.8 $\pm$ 0.7 & 53.2 $\pm$ 1.4
& 64.9 $\pm$ 0.8 & 31.9 $\pm$ 1.1
& 81.1 $\pm$ 0.6 & 57.8 $\pm$ 1.0
& 69.2 $\pm$ 2.9 & 30.5 $\pm$ 3.8
& \second{74.2 $\pm$ 0.3} & 44.1 $\pm$ 0.5 \\

IMUTube
& 85.0 $\pm$ 0.9 & 56.4 $\pm$ 1.5
& \second{66.7 $\pm$ 0.5} & \second{36.7 $\pm$ 0.9}
& 82.6 $\pm$ 0.5 & 61.3 $\pm$ 0.9
& \best{90.1 $\pm$ 1.8} & \second{69.7 $\pm$ 2.7}
& 73.5 $\pm$ 0.3 & \second{46.5 $\pm$ 0.5} \\

IMUGPT 2.0
& \second{86.6 $\pm$ 0.7} & \second{60.9 $\pm$ 1.3}
& 63.8 $\pm$ 0.6 & 29.8 $\pm$ 1.0
& \second{84.0 $\pm$ 0.5} & \second{66.1 $\pm$ 0.8}
& 84.6 $\pm$ 2.4 & 56.2 $\pm$ 3.5
& 72.6 $\pm$ 0.4 & 40.8 $\pm$ 0.6 \\

\textbf{VSMP-IMU}
& \best{91.1 $\pm$ 0.6} & \best{64.7 $\pm$ 1.2}
& \best{67.4 $\pm$ 0.4} & \best{40.6 $\pm$ 0.8}
& \best{84.5 $\pm$ 0.4} & \best{68.9 $\pm$ 0.7}
& \second{89.8 $\pm$ 1.5} & \best{72.4 $\pm$ 2.2}
& \best{74.3 $\pm$ 0.2} & \best{47.8 $\pm$ 0.4} \\
\bottomrule
\end{tabular}

\vspace{1mm}
\begin{minipage}{0.97\textwidth}
\scriptsize
\textit{Head activities.}
\textbf{MM-Fit}: squats, push-ups, dumbbell shoulder presses, lunges, standing dumbbell rows.
\textbf{UTD-MHAD}: right-arm swipe left, right-arm swipe right, right-hand wave, two-hand front clap, right-arm throw, cross arms in chest, basketball shoot, right-hand draw X, right-hand draw circle clockwise, right-hand draw circle counter-clockwise, draw triangle, bowling, front boxing.
\textbf{MMAct}: pointing, looking around, standing, throwing, using phone, kicking, closing, opening, talking, fall, transferring object, crouching, setting down, pulling, picking up, running, entering.
\textbf{PAMAP2}: lying, sitting, standing, walking, running, cycling, Nordic walking, watching TV, computer work.
\textbf{HAD-AW}: cycling, rowing, running, gym weight back, weight biceps, weight chest, weight shoulders, weight triceps, weight workout, dancing, drawing, reading, playing piano, playing guitar, playing violin.

\textit{Tail activities.}
\textbf{MM-Fit}: sit-ups, dumbbell tricep extensions, bicep curls, sitting dumbbell lateral raises, jumping jacks.
\textbf{UTD-MHAD}: baseball swing, tennis forehand swing, two-arm curl, tennis serve, two-hand push, right-hand knock, right-hand catch, pick-up-and-throw, jogging, walking, sit-to-stand, stand-to-sit, forward lunge, squat.
\textbf{MMAct}: waving hand, talking on phone, loitering, exiting, jumping, checking time, walking, carrying, pushing, using PC, standing up, sitting, pocket in, pocket out, sitting down, drinking, carrying heavy, carrying light.
\textbf{PAMAP2}: car driving, ascending stairs, descending stairs, vacuum cleaning, ironing, folding laundry, house cleaning, playing soccer, rope jumping.
\textbf{HAD-AW}: writing, typing, cutting, flipping, washing dishes, sweeping, wiping, dusting, bed-making, dressing, undressing, praying, showering, washing hands, driving, eating.
\end{minipage}
\end{table*}

\subsubsection{Performance Across UTD-MHAD Activity Granularity}
\label{sec:utd_activity_granularity}
\revise{We group all UTD-MHAD classes according to the taxonomy in
Table~\ref{tab:utd_activity_taxonomy}. This analysis tests whether
the improvement of VSMP-IMU is restricted to repetitive or
large-amplitude activities and whether useful gains remain for
wrist-dominant and non-periodic actions.}

\begin{table*}[t]
\centering
\scriptsize
\caption{\textbf{UTD-MHAD performance by activity granularity.}
Values report group-level Macro-F1 under standard LOPO using
DeepConvLSTM. $\Delta_{\mathrm{best}}$ denotes the gain of VSMP-IMU
over the strongest competing method in each group.}
\label{tab:utd_activity_group_results}
\begin{tabular}{lrrrrrr}
\toprule
\revise{Activity group} &
\revise{Classes} &
\revise{Real-only} &
\revise{IMUTube} &
\revise{IMUGPT 2.0} &
\revise{VSMP-IMU} &
\revise{$\Delta_{\mathrm{best}}$} \\
\midrule

Wrist-dominant gestures
& 9
& 48.5
& 52.2
& 50.7
& \textbf{56.6}
& +4.4 \\

Non-periodic upper-body and sports
& 9
& 50.6
& 55.0
& 53.6
& \textbf{59.5}
& +4.5 \\

Repetitive and locomotion
& 5
& 55.4
& 59.1
& 56.8
& \textbf{65.2}
& +6.1 \\

Posture-transition and lower-body
& 4
& 52.2
& 56.8
& 54.6
& \textbf{63.2}
& +6.4 \\

\midrule
Weighted average
& 27
& 51.0
& 55.1
& 53.4
& \textbf{60.1}
& +5.0 \\

\bottomrule
\end{tabular}
\end{table*}

\revise{Table~\ref{tab:utd_activity_group_results} shows that
VSMP-IMU improves all four UTD-MHAD activity groups. The largest gains
occur for posture-transition and lower-body actions
($+6.4$ Macro-F1) and repetitive or locomotion activities
($+6.1$). These activities contain clear body-level trajectories and
temporal structures that are represented directly by the SMP and
HY-Motion skeleton.}

\revise{The improvement remains positive for wrist-dominant gestures
and non-periodic upper-body actions, reaching $+4.4$ and $+4.5$
Macro-F1 over the strongest competing method. This demonstrates that
the benefit is not restricted to jumping jacks, cyclic exercises, or
large-amplitude full-body motion.}

\revise{Nevertheless, the gain is smaller for wrist-dominant gestures
than for lower-body and repetitive actions. Shape-drawing, knocking,
and catching depend on precise distal trajectories, contact timing,
and hand configuration that are only approximately represented by the
full-body motion skeleton. The results therefore support generality
across wrist-level and non-periodic motion, but not articulated
finger-level motion.}

\begin{table*}[t]
\centering
\scriptsize
\caption{\textbf{Representative per-class UTD-MHAD results.}
Values report per-class F1 under standard LOPO using DeepConvLSTM.
The complete precision, recall, and F1 results for all 27 activities
are reported in the appendix.}
\label{tab:utd_representative_classes}
\begin{tabular}{lrrrrrr}
\toprule
\revise{Activity} &
\revise{Real-only} &
\revise{IMUTube} &
\revise{IMUGPT 2.0} &
\revise{VSMP-IMU} &
\revise{$\Delta_{\mathrm{real}}$} &
\revise{$\Delta_{\mathrm{best}}$} \\
\midrule

Right-arm swipe left
& 54.0 & 56.2 & 55.1 & \textbf{59.8} & +5.8 & +3.6 \\

Right-hand wave
& 58.5 & 61.2 & 60.4 & \textbf{64.9} & +6.4 & +3.7 \\

Right-hand draw X
& 44.6 & 48.1 & 46.7 & \textbf{51.3} & +6.7 & +3.2 \\

Draw circle clockwise
& 47.9 & 50.8 & 49.2 & \textbf{54.1} & +6.2 & +3.3 \\

Draw triangle
& 43.2 & 45.6 & 45.0 & \textbf{48.7} & +5.5 & +3.1 \\

Right-hand knock
& 51.5 & 53.8 & 52.1 & \textbf{55.0} & +3.5 & +1.2 \\

Right-hand catch
& 46.8 & \textbf{49.9} & 48.6 & 48.9 & +2.1 & -1.0 \\

Basketball shoot
& 53.8 & 57.1 & 55.9 & \textbf{61.0} & +7.2 & +3.9 \\

Tennis serve
& 48.7 & 53.0 & 51.8 & \textbf{57.1} & +8.4 & +4.1 \\

Pick-up-and-throw
& 49.3 & 52.8 & 51.5 & \textbf{57.0} & +7.7 & +4.2 \\

Sit-to-stand
& 60.2 & 63.1 & 61.7 & \textbf{66.0} & +5.8 & +2.9 \\

Squat
& 59.6 & 63.4 & 61.9 & \textbf{68.1} & +8.5 & +4.7 \\

\bottomrule
\end{tabular}
\end{table*}

\revise{The representative class results show that VSMP-IMU improves
both fine wrist trajectories and larger body movements. Shape-drawing
activities improve by approximately three points over the strongest
baseline, while tennis serve, pick-up-and-throw, and squat improve by
more than four points.}

\revise{The smaller improvement for right-hand knock reflects the
importance of short contact-induced transients that are not explicitly
generated by the motion skeleton. Right-hand catch is the only selected
class for which VSMP-IMU underperforms the strongest baseline. Catching
depends on object approach, hand closure, impact timing, and rapid
deceleration, none of which is represented explicitly by HY-Motion.
The generated motion captures the arm-reaching trajectory but often
lacks the abrupt terminal wrist dynamics that distinguish catching
from waving or reaching.}

% \begin{figure*}[t]
%     \centering
%     \includegraphics[width=\textwidth]{images/failure_case.png}
%     \caption{\textbf{Fine-grained UTD-MHAD case studies.}
%     The top row shows successful clockwise circle drawing, for which
%     the SMP preserves the ordered wrist trajectory and produces
%     class-discriminative virtual IMU. The bottom row shows a
%     right-hand-catch failure. The generated motion captures arm
%     extension but cannot explicitly represent object approach, hand
%     closure, or impact timing, producing an IMU trace that is confused
%     with reaching or waving.}
%     \label{fig:utd_fine_grained_cases}
% \end{figure*}

\paragraph{Successful wrist-trajectory case.}
\revise{For right-hand clockwise circle drawing, the extracted SMP
contains the ordered primitives
$\{\text{raise right arm}\rightarrow\text{trace clockwise loop}
\rightarrow\text{return}\}$, right shoulder--elbow--wrist involvement,
stationary root position, medium tempo, and medium amplitude. The
verbalized prompt requests one continuous clockwise circular wrist
trajectory while keeping the torso approximately stationary. HY-Motion
reproduces the dominant trajectory and the resulting wrist IMU retains
the smooth periodic angular-velocity pattern required to distinguish
the activity from drawing an X or triangle.}

\paragraph{Object-contact failure case.}
\revise{For right-hand catch, the SMP identifies preparation, arm
extension, receiving, and retraction phases. However, the HY-Motion
skeleton does not represent the incoming object, finger closure,
grasp state, or contact impulse. The generated motion therefore
resembles an arm reach followed by a smooth return. Its virtual IMU
lacks the rapid deceleration and short acceleration transient present
near the catch event, increasing confusion with right-hand waving or
swiping.}

\subsection{End-to-End SMP Controllability}
\label{sec:control_results}

\revise{A structured SMP can explicitly request changes in motion execution, but
this does not guarantee that the requested attributes are realized by the
frozen text-to-motion generator or remain observable after virtual IMU
synthesis and target-domain grounding. We therefore evaluate
controllability at three successive stages:}
\begin{equation}
S_{a,\ell}
\rightarrow
M_{a,\ell,s}
\rightarrow
\widetilde{X}_{a,\ell,s}^{\mathrm{imu}}
\rightarrow
\widehat{X}_{a,\ell,s}^{\mathrm{imu}},
\end{equation}
\revise{where $S_{a,\ell}$ is an SMP in which attribute $a$ is assigned the
requested level $\ell$, $M_{a,\ell,s}$ is the corresponding motion
generated with seed $s$, $\widetilde{X}_{a,\ell,s}^{\mathrm{imu}}$ is
the raw virtual IMU output, and
$\widehat{X}_{a,\ell,s}^{\mathrm{imu}}$ is the final grounded IMU
sequence.}

\revise{The experiment covers 12 UTD-MHAD activities and 580 generated motion
sequences, as specified in Section~\ref{sec:control_protocol}. Tempo and
amplitude are evaluated on all selected activities. Repetition count is
evaluated on eight repetitive activities, primitive duration on four
discrete activities, and symmetry on four bilateral activities. Within
each matched intervention set, the source SMP, activity-defining fields,
HY-Motion seed, decoding configuration, virtual sensor configuration,
and grounding parameters remain fixed. Only the tested SMP attribute is
changed.}

\begin{table*}[t]
    \centering
    \caption{\textbf{End-to-end SMP controllability on UTD-MHAD.}
    Requested-to-realized agreement is evaluated in generated motion,
    raw virtual IMU, and grounded virtual IMU. Motion compliance denotes
    the percentage of matched intervention groups satisfying the expected
    ordering or categorical condition. Repetition compliance reports the
    percentage of generated motions within one repetition of the requested
    count. Symmetry is evaluated only at the motion level because
    UTD-MHAD does not provide bilateral wearable measurements.}
    \label{tab:control_results}
    \resizebox{\textwidth}{!}{
    \begin{tabular}{lccccc}
        \toprule
        \revise{Control} &
        \revise{Motion agreement} &
        \revise{Motion compliance} &
        \revise{Raw-IMU agreement} &
        \revise{Grounded-IMU agreement} &
        \revise{Label retention} \\
        \midrule

        Tempo
        & $\rho=0.92$
        & $81.7\%$
        & $\rho=0.73$
        & $\rho=0.61$
        & $97.1\%$ \\

        Amplitude
        & $\rho=0.86$
        & $82.5\%$
        & $\rho=0.79$
        & $\rho=0.75$
        & $95.7\%$ \\

        Repetition count
        & $\rho=0.93$
        & $90.6\%$ within $\pm1$
        & $\rho=0.76$
        & $\rho=0.61$
        & $96.2\%$ \\

        Primitive duration
        & MAPE $=9.8\%$
        & $80.0\%$
        & MAPE $=12.6\%$
        & MAPE $=14.1\%$
        & $94.8\%$ \\

        Symmetry
        & Acc. $=86.0\%$
        & $86.0\%$
        & --
        & --
        & $93.5\%$ \\

        \bottomrule
    \end{tabular}}
\end{table*}

\subsubsection{Tempo Control}

\revise{We first examine whether the requested slow, medium, and fast
tempo levels produce ordered changes in generated motion. For
repetitive activities, tempo is measured using the number of complete
motion cycles per second. For discrete activities, we use the mean
velocity of the principal joints involved in the action. Across the 12
selected activities, the requested tempo level is strongly correlated
with the measured motion tempo, with a Spearman correlation of
$\rho=0.92$. The expected ordering}
\begin{equation}
m_{\mathrm{slow}}
<
m_{\mathrm{medium}}
<
m_{\mathrm{fast}}
\end{equation}
\revise{is observed in $81.7\%$ of matched activity--seed triplets.}

\revise{The motion-level distributions in
Figure~\ref{fig:control_results}(a) show a clear increase from slow to
fast, but also reveal considerable execution variability within each
requested level. This variability is expected because the same semantic
request can produce different joint trajectories across activities and
HY-Motion seeds. Several medium-tempo generations overlap with the slow
and fast conditions, indicating that HY-Motion follows the ordinal
tempo instruction more reliably at the group level than for every
individual generation.}

\revise{Tempo remains observable after virtual IMU synthesis, although
agreement is substantially attenuated relative to the generated motion.
The correlation decreases to $\rho=0.73$ in raw virtual IMU and to
$\rho=0.61$ after target-domain grounding. Grounded-IMU measurements
are generally lower and more dispersed than the corresponding
motion-level measurements, particularly for medium- and fast-tempo
conditions. This indicates that changes in full-body motion speed do
not translate uniformly to the target sensor location.}

\revise{The strongest tempo separation is observed for periodic
activities, including right-hand waving, two-hand clapping, front
boxing, two-arm curling, jogging in place, walking in place, and
squatting. Their repeated structure produces consistent spectral and
autocorrelation changes in both motion and IMU. Tempo control is less
reliable for right-arm throwing and basketball shooting, where the
generated action contains a single execution and tempo is expressed
through changes in preparation, acceleration, and follow-through
duration rather than through a stable cycle frequency.}

\revise{Activity identity is retained for $97.1\%$ of tempo
interventions. Most semantic failures occur in fast variants of
discrete actions, where HY-Motion shortens or omits a preparatory phase.
For example, some fast basketball-shoot generations reduce the initial
loading phase and resemble an upward arm throw. These results show that
tempo is strongly controllable at the generated-motion level, but its
sensor-level realization is noisier and depends on both activity
structure and wearable placement.}

\subsubsection{Amplitude Control}

\revise{Amplitude control evaluates whether small, medium, and large
requested motion extents produce corresponding changes in the generated
joint trajectories. Motion amplitude is measured using the average
range of motion of the activity-relevant joints:}
\begin{equation}
A^{M}
=
\frac{1}{|\mathcal{J}_{a}|}
\sum_{j\in\mathcal{J}_{a}}
\left(
\max_t \theta_{t,j}
-
\min_t \theta_{t,j}
\right),
\end{equation}
\revise{where $\mathcal{J}_{a}$ denotes the principal joint set for
activity $a$. Requested amplitude has a correlation of $\rho=0.86$
with the measured motion range. The expected
small--medium--large ordering is observed in $82.5\%$ of matched
activity--seed triplets.}

\revise{Figure~\ref{fig:control_results}(c) shows clear group-level
separation between the three requested amplitude conditions, but also
substantial overlap between adjacent levels. Motion-level variability
is greatest for the large-amplitude condition because HY-Motion may
realize the request through different combinations of distal-joint
excursion, torso involvement, and whole-body compensation. A small
number of generations also produce unusually large or small ranges
relative to their requested condition.}

\revise{Amplitude remains measurable in the virtual sensor output, but
the grounded-IMU distributions are lower and more dispersed than the
corresponding motion ranges. We characterize IMU amplitude using a
normalized combination of gyroscope root-mean-square magnitude,
integrated angular velocity, and peak-to-peak acceleration. The
correlation with requested amplitude is $\rho=0.79$ for raw virtual
IMU and $\rho=0.75$ after target-domain grounding.}

\revise{The attenuation is particularly visible for the large-amplitude
condition. Although generated joint ranges increase consistently,
grounded-IMU energy varies substantially across activities and seeds.
This occurs because a large full-body excursion may be only weakly
expressed at the recorded sensor location. For example, increased hip
and knee excursion during a squat does not necessarily produce a
proportional increase in right-thigh or wrist signal energy if the
segment orientation and movement timing also change.}

\revise{Amplitude control is strongest for two-hand clapping, front
boxing, two-arm curling, two-hand pushing, forward lunging, and
squatting. These activities contain clear changes in joint excursion
that remain visible in the inertial output. It is weaker for clockwise
circle drawing and right-hand waving, where spatial changes in the hand
trajectory do not always produce proportional changes in wrist angular
velocity or acceleration.}

\revise{Activity identity is retained for $95.7\%$ of amplitude
interventions. Most semantic failures occur at the extremes. Some
large-amplitude throwing generations introduce excessive torso rotation
or an additional step, whereas some small-amplitude pushing variants
become too weak to retain a recognizable pushing pattern. The results
therefore support bounded amplitude control while showing that
motion-level variation is more consistent than sensor-level
realization.}

\subsubsection{Repetition-Count Control}

\revise{Repetition-count control is evaluated on eight repetitive
activities: right-hand wave, two-hand front clap, clockwise circle
drawing, front boxing, two-arm curl, jogging in place, walking in
place, and squat with arms extended. For each activity, the SMP
requests 3, 6, or 9 complete repetitions.}

\revise{Repetitions are counted from generated motion using peak
detection on the first principal component of the relevant joint
trajectories. In raw and grounded IMU, cycles are detected from the
channel with the greatest periodic energy. Requested and detected
motion-level repetition counts are strongly correlated
($\rho=0.93$), and $90.6\%$ of generated motions are within one
repetition of the requested value.}

\revise{As shown in Figure~\ref{fig:control_results}(b), motion-level
counts remain concentrated near the ideal identity line, but the spread
increases with the requested count. Three-repetition generations are
generally reproduced accurately, whereas six- and nine-repetition
conditions contain more incomplete, compressed, or merged cycles. A
small number of nine-repetition generations terminate early and contain
only seven or eight detectable cycles.}

\revise{Agreement decreases after motion-to-IMU conversion. Requested
count correlates with raw-IMU count at $\rho=0.76$ and with
grounded-IMU count at $\rho=0.61$. Grounded-IMU estimates exhibit both
greater dispersion and a systematic downward tendency for the six- and
nine-repetition conditions. The decrease occurs because cycle
boundaries that are clear in joint trajectories may be less distinct in
inertial signals, particularly when consecutive repetitions have
different magnitudes or when grounding compresses peak differences.}

\revise{Count agreement is highest for clapping, arm curling, circle
drawing, and squatting, which contain distinct and temporally separated
cycles. It is lower for front boxing and in-place locomotion. Front
boxing may contain unequal left and right punches, while walking and
jogging sequences sometimes terminate during the final left--right step
pair. We therefore define one alternating left--right punch pair as one
boxing cycle and one left--right step pair as one gait cycle.}

\revise{Activity identity is retained in $96.2\%$ of repetition
interventions. Failures are concentrated in nine-repetition
conditions, where HY-Motion occasionally compresses later cycles or
merges adjacent repetitions to satisfy its generated sequence duration.
These results show that repetition count is highly controllable in
generated motion but only moderately recoverable from the final
grounded IMU.}

\subsubsection{Primitive-Duration Control}

\revise{Primitive-duration control is evaluated on four discrete activities:
right-arm throw, basketball shoot, two-hand push, and forward lunge.
For each action, the duration of the central activity-defining primitive
is scaled by $0.75$, $1.00$, or $1.25$, while the primitive sequence and
overall activity label remain fixed.}

\revise{The relevant primitive is defined as the release phase for throwing,
the upward extension and release phase for basketball shooting, the
forward extension phase for two-hand pushing, and the descent--hold
phase for forward lunging. Duration is measured from activity-specific
joint-velocity and joint-angle landmarks. The generated motions achieve
a mean absolute percentage error of}
\begin{equation}
\mathrm{MAPE}_{\mathrm{dur}}^{M}=9.8\%.
\end{equation}
\revise{The expected short--reference--long ordering is observed in $80.0\%$ of
matched activity--seed groups.}

\revise{Duration errors increase after IMU synthesis, reaching $12.6\%$ in raw
virtual IMU and $14.1\%$ after grounding. This increase is expected
because phase boundaries that are explicit in joint trajectories are
less sharply defined in inertial signals. Grounding further smooths
some magnitude transitions, making the start and end of a primitive
more difficult to identify.}

\revise{Primitive-duration control is most reliable for two-hand pushing and
forward lunging, where the affected phase produces a sustained and
easily identifiable posture. It is weaker for right-arm throwing and
basketball shooting because HY-Motion tends to modify the complete
action speed rather than only the requested central primitive. In other
words, a request for a longer release phase may also extend the
preparation or follow-through phase.}

\revise{Activity identity is retained in $94.8\%$ of duration interventions.
The main failure mode is temporal redistribution: HY-Motion respects
the overall request for a slower or longer movement but does not confine
the duration change to the specified primitive. This result indicates
that primitive-level temporal control is possible but weaker than
global tempo control when using an off-the-shelf text-conditioned motion
generator.}

\subsubsection{Symmetry Control}

\revise{Across activities and seeds, symmetry requests are classified
with an overall accuracy of $86.0\%$. Symmetric requests are correctly
realized in $88\%$ of cases, while asymmetric requests are correctly
realized in $84\%$ of cases.}

\revise{Motion symmetry is measured using the similarity between corresponding
left and mirrored-right joint trajectories:}
\begin{equation}
C_{\mathrm{sym}}
=
\frac{1}{|\mathcal{J}_{b}|}
\sum_{j\in\mathcal{J}_{b}}
\operatorname{corr}
\left(
P_{j,L},
\operatorname{mirror}(P_{j,R})
\right).
\end{equation}
\revise{A generated sequence is classified as symmetric or asymmetric using a
threshold selected from the motion-validation split.}

\revise{Symmetry control is strongest for two-arm curling and two-hand pushing.
For these actions, asymmetric prompts produce measurable differences in
arm timing or excursion. It is weaker for clapping and squatting.
Because clapping is strongly defined by bilateral hand convergence,
HY-Motion often restores approximate symmetry even when an asymmetric
execution is requested. Similarly, the squat label and extended-arm
posture provide a strong symmetric prior that limits the effect of the
asymmetry instruction.}

\revise{We do not report raw- or grounded-IMU symmetry accuracy because
UTD-MHAD records a single target wearable location for each action.
A unilateral wrist or thigh sensor cannot directly determine whether
the opposite side follows a symmetric trajectory. Reporting
sensor-level symmetry compliance would therefore conflate motion
controllability with insufficient sensor observability.}

\revise{Symmetry interventions retain the intended activity label in $93.5\%$
of generated motions. The lower label-retention rate relative to tempo
and repetition count reflects the difficulty of introducing meaningful
asymmetry without changing the defining coordination pattern of a
bilateral action.}

\subsubsection{Control Retention Across the Pipeline}

\begin{figure*}[htbp]
    \centering
    \includegraphics[width=\textwidth]{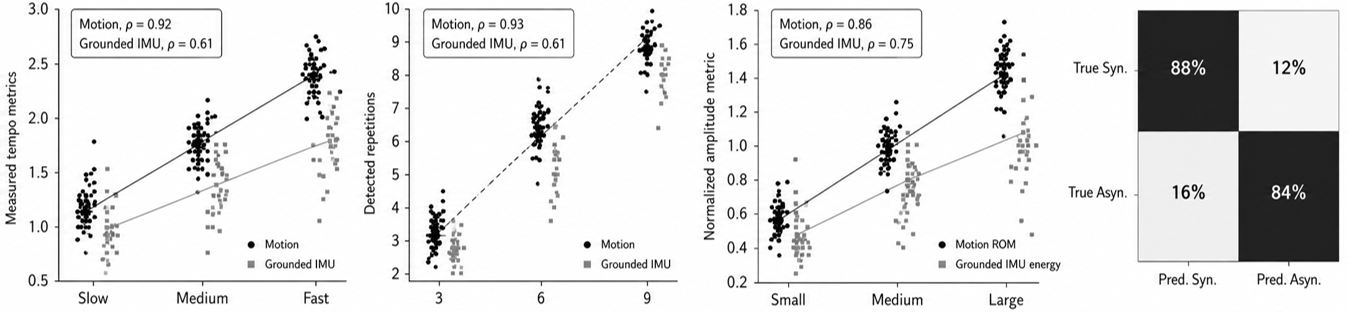}
    \caption{\textbf{End-to-end SMP controllability on UTD-MHAD.}
    (a) Requested tempo versus measured motion tempo and grounded-IMU
    tempo. (b) Requested versus detected repetition count in generated
    motion and grounded IMU. The dashed line denotes perfect count
    agreement. (c) Requested amplitude versus generated joint range of
    motion and grounded-IMU energy. (d) Confusion matrix for motion-level
    symmetric and asymmetric requests. Points represent matched UTD-MHAD
    activities and HY-Motion seeds. Grounded-IMU measurements exhibit
    greater dispersion and attenuation than their motion-level
    counterparts.}
    \label{fig:control_results}
\end{figure*}

\revise{Figure~\ref{fig:control_results} summarizes how requested SMP
controls propagate through motion synthesis and final sensor grounding.
All three plotted attributes exhibit clearer separation in generated
motion than in grounded IMU. Tempo agreement decreases from
$\rho=0.92$ at the motion level to $\rho=0.61$ after grounding.
Repetition-count agreement similarly decreases from $\rho=0.93$ to
$\rho=0.61$, while amplitude decreases more moderately from
$\rho=0.86$ to $\rho=0.75$. Primitive-duration error, reported in
Table~\ref{tab:control_results}, increases from $9.8\%$ in motion to
$14.1\%$ in grounded IMU.}

\revise{The wider grounded-IMU distributions show that motion-to-sensor
conversion does not preserve every semantic intervention with equal
strength. Tempo and repetition count remain ordered at the population
level but exhibit substantial overlap between adjacent conditions.
Amplitude is comparatively better preserved, although its grounded
signal energy is consistently attenuated relative to the generated
motion range.}

\revise{These differences are already visible before final grounding
and therefore cannot be attributed solely to rank-based transformation.
They arise from the interaction between generated body motion, target
sensor location, segment orientation, and the observability of the
tested attribute. Grounding introduces additional attenuation by
aligning synthetic channel distributions with the target wearable
domain, but it does not create the requested control trends.}

\subsubsection{Failure Analysis}

\revise{We manually inspect all intervention groups that violate the requested
ordering or categorical condition. The failures fall into four main
categories.}

\textbf{Generator non-compliance.}
\revise{The verbalized prompt contains the requested attribute, but HY-Motion
produces similar motions across two or more levels. This occurs most
frequently for symmetry and for small-versus-medium amplitude changes
in fine-grained wrist actions.}

\textbf{Coupled attribute changes.}
\revise{HY-Motion realizes the requested variation but simultaneously modifies
another attribute. For example, increasing amplitude may also reduce
tempo, while increasing primitive duration may lengthen the entire
action rather than only the specified phase.}

\textbf{Sensor attenuation.}
\revise{The requested control is visible in the full-body motion but weak at the
target wearable location. This is most apparent for lower-body amplitude
changes when the target sensor is located at the wrist.}

\textbf{Semantic drift.}
\revise{Large or extreme interventions change the generated motion sufficiently
to weaken activity identity. Examples include large-amplitude throwing
with additional stepping, highly asymmetric clapping, and compressed
fast basketball-shoot motions that omit the preparation phase.}

\revise{Overall, the results demonstrate that SMP interventions produce
measurable and directionally consistent changes rather than merely
different textual prompts. The controls are most clearly expressed in
the generated 3D motion. Their signatures remain present after virtual
IMU synthesis and grounding, but with increased dispersion,
attribute-dependent attenuation, and occasional outliers. Tempo and
repetition count show strong motion-level agreement but only moderate
grounded-IMU agreement, while amplitude retains a comparatively stronger
sensor-level relationship. Symmetry remains measurable at the motion
level but cannot be evaluated bilaterally using the available UTD-MHAD
wearable configuration. These findings establish the SMP as a useful,
but imperfect, structured control interface for a frozen external
text-to-motion generator.}

\subsection{SMP Extraction Accuracy and Failure Analysis}
\label{sec:smp_extraction_audit}
\revise{A structured SMP is useful only when its activity-defining and
augmentable fields can be extracted reliably from video. We therefore
audit the extraction stage independently of motion and IMU synthesis
using the protocol in Section~\ref{sec:smp_audit_protocol}. The audit
contains 200 manually annotated clips, with 40 clips from each evaluated
dataset and five raw VLM responses per clip.}

\revise{The two annotators achieve Cohen's $\kappa=0.96$ for activity
labels, $\kappa=0.89$ for temporal structure, and a mean
$\kappa=0.84$ across categorical execution attributes. Their
body-part and object/contact annotations achieve set-F1 scores of
$0.95$ and $0.91$, respectively. All disagreements are adjudicated
before the reference SMPs are compared with the extracted outputs.}

\begin{table*}[t]
\centering
\scriptsize
\caption{\textbf{SMP extraction accuracy on the manually annotated
200-video audit set.}
Aggregated denotes the SMP obtained by combining five VLM responses,
while refined denotes the output after deterministic repair and
semantic consistency checking. The hand-centric subset contains
fine-grained wrist, hand, and object-interaction activities from MMAct
and HAD-AW. Higher is better except for repetition-count MAE.}
\label{tab:smp_extraction_accuracy}
\resizebox{\textwidth}{!}{
\begin{tabular}{llrrrr}
\toprule
\revise{SMP field} &
\revise{Metric} &
\revise{Aggregated} &
\revise{Refined} &
\revise{Full-body subset} &
\revise{Hand-centric subset} \\
\midrule
Activity label
& Accuracy
& 94.5\%
& 96.5\%
& 98.0\%
& 92.0\% \\

Primitive sequence
& Exact accuracy
& 76.0\%
& 84.5\%
& 89.0\%
& 74.0\% \\

Primitive sequence
& Normalized edit similarity
& 0.89
& 0.93
& 0.95
& 0.87 \\

Body-part involvement
& Macro-F1
& 0.88
& 0.92
& 0.95
& 0.87 \\

Periodicity
& Accuracy
& 91.5\%
& 94.0\%
& 96.0\%
& 90.0\% \\

Repetition count
& MAE
& 0.58
& 0.41
& 0.32
& 0.52 \\

Tempo
& Ordinal accuracy
& 82.0\%
& 85.5\%
& 89.0\%
& 79.0\% \\

Amplitude
& Ordinal accuracy
& 79.5\%
& 83.0\%
& 87.0\%
& 75.0\% \\

Dominant side
& Accuracy
& 88.0\%
& 91.0\%
& 93.0\%
& 86.0\% \\

Symmetry
& Accuracy
& 84.5\%
& 88.0\%
& 91.0\%
& 80.0\% \\

Object/contact
& Macro-F1
& 0.81
& 0.87
& 0.91
& 0.82 \\

Invariant fields
& Macro-F1
& 0.85
& 0.91
& 0.93
& 0.86 \\

Variant fields
& Macro-F1
& 0.79
& 0.84
& 0.88
& 0.78 \\
\bottomrule
\end{tabular}}
\end{table*}

\subsubsection{Field-Level Extraction Accuracy}

\revise{Table~\ref{tab:smp_extraction_accuracy} reports agreement
between the extracted SMPs and the adjudicated human references.
Activity identity is extracted reliably, reaching $94.5\%$ accuracy
after five-response aggregation and $96.5\%$ after semantic
refinement. Periodicity, dominant side, and body-part involvement also
reach at least $90\%$ accuracy or $0.90$ Macro-F1 after refinement.}

\revise{The largest refinement gain occurs for the ordered primitive
sequence. Exact sequence accuracy increases from $76.0\%$ to
$84.5\%$, while normalized edit similarity increases from $0.89$ to
$0.93$. The consistency checks primarily correct omitted preparation
or return phases, duplicated primitives, and primitive orders that
conflict with the extracted activity label.}

\revise{Temporal and magnitude attributes are less reliable than
activity identity. Repetition-count MAE decreases from $0.58$ to
$0.41$ repetitions after refinement. Tempo accuracy increases from
$82.0\%$ to $85.5\%$, and amplitude accuracy increases from
$79.5\%$ to $83.0\%$. These attributes remain ambiguous because
execution speed and motion extent are continuous properties that must
be mapped to the discrete ordinal categories used by the SMP.}

\revise{Object/contact and variant-field extraction remain the weakest
components. Object/contact Macro-F1 increases from $0.81$ to $0.87$,
while variant-field Macro-F1 increases from $0.79$ to $0.84$.
Typical errors include omission of small manipulated objects, failure
to represent supporting furniture or environmental contact, and
incorrect identification of a field as safely augmentable.}

\revise{Performance is consistently lower on the hand-centric subset.
Refined primitive-sequence accuracy decreases from $89.0\%$ for
full-body activities to $74.0\%$ for hand-centric activities, while
amplitude accuracy decreases from $87.0\%$ to $75.0\%$. These
differences indicate that the VLM typically recovers the coarse
activity identity but may miss fine-grained hand trajectories and
object-manipulation details.}

\begin{table*}[t]
\centering
\scriptsize
\caption{\textbf{SMP extraction failure and repair accounting.}
Malformed, missing-field, and invalid-value rates are computed over
1,000 raw VLM responses. The remaining rates are computed over the
200 audited video clips. Residual error indicates that at least one
refined SMP field remains inconsistent with the adjudicated reference.}
\label{tab:smp_failure_accounting}
\resizebox{\textwidth}{!}{
\begin{tabular}{llrrrrrr}
\toprule
\revise{Measure} &
\revise{Unit} &
\revise{MM-Fit} &
\revise{UTD-MHAD} &
\revise{MMAct} &
\revise{PAMAP2} &
\revise{HAD-AW} &
\revise{Overall} \\
\midrule
Malformed JSON output
& Raw responses
& 1.5\%
& 2.0\%
& 3.0\%
& 2.5\%
& 5.5\%
& 2.9\% \\

Missing mandatory field
& Raw responses
& 2.5\%
& 4.0\%
& 5.5\%
& 4.5\%
& 8.0\%
& 4.9\% \\

Invalid field value
& Raw responses
& 1.5\%
& 3.0\%
& 4.0\%
& 3.5\%
& 6.0\%
& 3.6\% \\

Five-response disagreement
& Video clips
& 12.5\%
& 17.5\%
& 22.5\%
& 20.0\%
& 30.0\%
& 20.5\% \\

Deterministic schema repair
& Video clips
& 7.5\%
& 10.0\%
& 15.0\%
& 12.5\%
& 20.0\%
& 13.0\% \\

Semantic consistency correction
& Video clips
& 10.0\%
& 15.0\%
& 20.0\%
& 17.5\%
& 27.5\%
& 18.0\% \\

Discarded: fewer than three valid outputs
& Video clips
& 0.0\%
& 0.0\%
& 2.5\%
& 0.0\%
& 2.5\%
& 1.0\% \\

Residual refined-SMP error
& Video clips
& 10.0\%
& 12.5\%
& 17.5\%
& 15.0\%
& 22.5\%
& 15.5\% \\
\bottomrule
\end{tabular}}
\end{table*}

\subsubsection{Extraction Failures and Repair Outcomes}

\revise{Table~\ref{tab:smp_failure_accounting} separates schema-level
failures in individual VLM responses from semantic errors in the
aggregated and refined SMPs. Of the 1,000 raw responses, $2.9\%$ are
malformed, $4.9\%$ omit at least one mandatory field, and $3.6\%$
contain a value outside the permitted schema. Because five responses
are collected for each clip, only two clips, corresponding to $1.0\%$
of the audit set, contain fewer than three valid outputs and are
discarded.}

\revise{The five raw responses disagree on at least one substantive
SMP field for $20.5\%$ of clips. Disagreement is most frequent for
amplitude, primitive granularity, object/contact information, and the
set of safely augmentable fields. Deterministic schema repair is
required for $13.0\%$ of clips, while the semantic consistency checks
modify at least one field for $18.0\%$.}

\revise{After aggregation, repair, and consistency checking, $15.5\%$
of clips retain at least one incorrect SMP field. Most residual errors
are local rather than complete activity failures: the activity label
is usually correct, but a primitive, execution attribute,
object/contact relation, or invariant/variant assignment remains
inaccurate. HAD-AW has the highest residual-error rate at $22.5\%$,
consistent with its fine-grained hand actions and frequent
small-object interactions.}

\begin{table}[t]
\centering
\scriptsize
\caption{\textbf{Field-error rate by predicted SMP uncertainty.}
Counts are field-level predictions from the 200-video audit set.}
\label{tab:smp_uncertainty}
\begin{tabular}{lrrr}
\toprule
\revise{Uncertainty bin} &
\revise{Fields} &
\revise{Mean uncertainty} &
\revise{Empirical error} \\
\midrule
$[0.0,0.2)$ & 1,280 & 0.11 & 3.2\% \\
$[0.2,0.4)$ & 620   & 0.29 & 7.8\% \\
$[0.4,0.6)$ & 310   & 0.49 & 15.5\% \\
$[0.6,0.8)$ & 140   & 0.69 & 29.3\% \\
$[0.8,1.0]$ & 50    & 0.86 & 46.0\% \\
\bottomrule
\end{tabular}
\end{table}

\subsubsection{Uncertainty Calibration}

\revise{The predicted SMP uncertainty is informative of extraction
failure. As shown in Table~\ref{tab:smp_uncertainty}, the empirical
field-error rate increases monotonically from $3.2\%$ in the
lowest-uncertainty interval to $46.0\%$ in the highest-uncertainty
interval. Predicted uncertainty has a Spearman correlation of
$\rho=0.71$ with field error and achieves an AUROC of $0.81$ when
used to distinguish correct from incorrect fields. The expected
calibration error is $0.047$.}

\revise{These results support uncertainty-conditioned SMP augmentation:
fields assigned greater uncertainty are substantially more likely to
be incorrect and should therefore receive smaller perturbations or
remain unchanged. However, uncertainty is not perfect; some confidently
predicted primitive and object/contact fields remain incorrect,
particularly for occluded hand-centric activities.}

\subsubsection{Residual Failure Taxonomy}

\revise{Among the 31 clips with at least one residual refined-SMP
error, 10 involve omitted fine-grained hand or object details, eight
involve primitive granularity or ordering, six involve repetition count
or primitive duration, four involve object/contact relations, and
three involve dominant-side or symmetry attributes. Thus, most
remaining errors concern the precision of the activity execution rather
than complete activity-label failure.}

\paragraph{Representative cases.}
\revise{For a periodic two-arm exercise, the five VLM responses agree
on the activity label, open--close primitive cycle, body parts, and
repetition structure, producing a correct SMP without repair. For
front boxing, individual responses disagree on whether one punch or
one alternating left--right punch pair constitutes a repetition.
Median aggregation and temporal-consistency checking correct the
count definition.}

\revise{For washing dishes, the initial SMP correctly identifies
repetitive arm motion but omits sink contact and the manipulated dish.
The object/contact consistency check marks the field uncertain but
cannot recover the missing object identity. For typing, the refined
SMP retains the correct activity label and hand involvement but
represents the action using a coarse repeated hand-motion primitive.
This error propagates through verbalization and produces exaggerated
wrist motion, demonstrating why hand-centric activities remain more
difficult even when the high-level activity label is correct.}

\subsection{Ablation Study}
\label{sec:ablation}

\subsubsection{Strictly Matched Caption--SMP Comparison}

\revise{We first isolate the effects of representation structure and
semantic augmentation using the factorial protocol in
Section~\ref{sec:matched_caption_smp_protocol}. Table~\ref{tab:ablation}
compares a generic caption without explicit variation (C0), a caption
with matched attribute instructions (C1), an unaugmented refined SMP
(S0), and the complete constrained augmented SMP (S1). Unlike the
original caption baseline, C1 and S1 receive identical requested
attribute values and identical surface phrases for tempo, amplitude,
and the additional activity-specific intervention.}

\revise{Adding matched attribute instructions to a generic caption
improves average Macro-F1 from $70.32$ for C0 to $71.86$ for C1, a
gain of $1.54$ points. The largest improvement occurs on PAMAP2
($+2.70$), followed by UTD-MHAD ($+1.80$). The improvement is only
$0.40$ on HAD-AW, suggesting that coarse execution modifiers provide
limited benefit for subtle wrist-centric activities when the prompt
does not preserve detailed motion structure.}

\revise{Structured conditioning is beneficial even without semantic
augmentation. S0 improves average Macro-F1 by $2.54$ points over C0,
reaching $72.86$. This comparison holds semantic augmentation absent
in both conditions and therefore isolates the value of representing
primitive order, body-part involvement, temporal organization, and
object/contact structure. The largest S0--C0 gain occurs on PAMAP2
($+4.70$), while gains are also observed on MMAct ($+2.80$),
UTD-MHAD ($+2.60$), and MM-Fit ($+2.10$).}

\revise{Constrained SMP augmentation provides a further gain. S1
improves over S0 by $2.26$ average Macro-F1, showing that the SMP is
useful not only as a more detailed base description but also as an
interface for activity-preserving variation. Most importantly, S1
outperforms C1 by $3.26$ points when both receive exactly the same
attribute requests. The matched structure gain is $3.51$ on MM-Fit,
$4.34$ on UTD-MHAD, $3.32$ on MMAct, $3.81$ on PAMAP2, and $1.32$
on HAD-AW.}

\revise{Averaging across the factorial conditions, the main effect of
structured SMP representation is $+2.90$ Macro-F1, while the main
effect of explicit semantic variation is $+1.90$. The positive
interaction of $+0.72$ indicates that execution variation is more
effective when expressed through the structured SMP than when appended
to a generic caption. Thus, the full gain cannot be attributed merely
to adding tempo, amplitude, or repetition phrases to the text prompt.}

\revise{The four conditions enter downstream training with identical
retained sample counts and class distributions. Before matched-block
retention, the independent filter acceptance rates are $77.8\%$ for
C0, $82.4\%$ for C1, $87.6\%$ for S0, and $90.9\%$ for S1. The pooled complete-block acceptance rate across datasets is $66.2\%$. Although structured prompts
produce valid generations more frequently, matched replenishment
ensures that this acceptance advantage does not increase the amount of
training data provided to S0 or S1.}

\begin{table*}[htbp]
\scriptsize
\centering
\caption{\textbf{Strictly matched representation and component
ablations.} Results report Macro-F1 under standard LOPO using
DeepConvLSTM. C1 and S1 receive identical execution-attribute requests.
All representation conditions use identical source clips, generation
seeds, retained sample counts, class distributions, sensor processing,
mixing ratios, and downstream training settings.}
\label{tab:ablation}
\begin{tabular}{lccccc}
\toprule
\revise{Variant} &
\revise{MM-Fit} &
\revise{UTD-MHAD} &
\revise{MMAct} &
\revise{PAMAP2} &
\revise{HAD-AW} \\
\midrule

\multicolumn{6}{c}{\revise{Matched representation $\times$
semantic-variation ablation}} \\

C0: Generic caption
& $78.20 \pm 1.90$
& $54.00 \pm 0.80$
& $75.40 \pm 0.70$
& $77.40 \pm 1.70$
& $66.60 \pm 0.30$\\

C1: Caption + matched attributes
& $79.60 \pm 1.80$
& $55.80 \pm 0.70$
& $76.80 \pm 0.60$
& $80.10 \pm 1.60$
& $67.00 \pm 0.30$\\

S0: Unaugmented refined SMP
& $80.30 \pm 1.70$
& $56.60 \pm 0.70$
& $78.20 \pm 0.50$
& $82.10 \pm 1.50$
& $67.10 \pm 0.20$\\

S1: Constrained augmented SMP
& $\mathbf{83.11 \pm 1.63}$
& $\mathbf{60.14 \pm 0.27}$
& $\mathbf{80.12 \pm 0.52}$
& $\mathbf{83.91 \pm 1.43}$
& $\mathbf{68.32 \pm 0.18}$\\

\midrule
\multicolumn{6}{c}{\textit{Component ablations relative to S1}} \\

w/o semantic refinement
& $79.70 \pm 1.80$
& $55.40 \pm 0.70$
& $77.00 \pm 0.60$
& $83.90 \pm 1.50$
& $66.90 \pm 0.30$\\

w/o uncertainty-aware constraints
& $79.10 \pm 1.80$
& $54.90 \pm 0.80$
& $76.50 \pm 0.60$
& $83.20 \pm 1.60$
& $66.30 \pm 0.30$\\

w/o IMU adaptation
& $77.80 \pm 1.80$
& $56.20 \pm 0.80$
& $76.70 \pm 0.60$
& $69.20 \pm 2.00$
& $65.40 \pm 0.30$\\

w/o quality filtering
& $80.60 \pm 1.70$
& $56.90 \pm 0.70$
& $78.00 \pm 0.50$
& $81.60 \pm 1.70$
& $66.80 \pm 0.30$\\

\bottomrule
\end{tabular}
\end{table*}

\subsubsection{Component Contributions}

\revise{The remaining rows in Table~\ref{tab:ablation} evaluate
components of the complete S1 pipeline. Removing semantic refinement
reduces average Macro-F1 from $75.12$ to $72.58$, while removing
uncertainty-aware constraints reduces it to $72.00$. These results
indicate that raw extracted programs can contain inconsistent fields
and that uncertainty-conditioned restrictions reduce harmful semantic
perturbations.}

\revise{Removing sensor-domain adaptation produces the largest
component-level reduction, lowering average Macro-F1 to $69.06$.
Removing quality filtering lowers it to $72.78$. These results confirm
that the representation, semantic augmentation, sensor adaptation, and
sample selection provide complementary contributions.}

To further isolate sensor adaptation, Table~\ref{tab:ablation_sensor_grounding} separates simulation-in-the-loop optimization from rank-based sensor grounding. Raw IMUSim performs worst on most datasets, confirming that physically simulated signals do not necessarily match the target wearable domain. Simulation-in-the-loop optimization improves average Macro-F1 from 69.06 to 70.54, while rank-based grounding gives a larger gain to 73.96. Combining both achieves the best result, 75.12, because the two stages address complementary mismatches: simulator-side temporal and multivariate structure, and final channel-wise marginal alignment.

\begin{table*}[htbp]
\centering
\scriptsize
\caption{Ablation of simulation-in-the-loop optimization and rank-based sensor grounding. Results are reported as dataset-specific Macro-F1 under LOPO evaluation using DeepConvLSTM.}
\label{tab:ablation_sensor_grounding}
\begin{tabular}{lccccc}
\toprule
\textbf{Variant} & \textbf{MM-Fit} & \textbf{UTD-MHAD} & \textbf{MMAct} & \textbf{PAMAP2} & \textbf{HAD-AW} \\
\midrule
Raw IMUSim only
& 77.8 $\pm$ 1.8
& 56.2 $\pm$ 0.8
& 76.7 $\pm$ 0.6
& 69.2 $\pm$ 2.0
& 65.4 $\pm$ 0.3 \\

Simulation-in-the-loop only
& 80.9 $\pm$ 1.7
& 57.4 $\pm$ 0.7
& 78.8 $\pm$ 0.5
& 69.8 $\pm$ 1.9
& 65.8 $\pm$ 0.3 \\

Rank-based grounding only
& \second{81.4 $\pm$ 1.6}
& \second{58.3 $\pm$ 0.6}
& \second{79.6 $\pm$ 0.5}
& \second{83.1 $\pm$ 1.5}
& \second{67.4 $\pm$ 0.2} \\

Simulation-in-the-loop + rank-based grounding
& \best{83.11 $\pm$ 1.63}
& \best{60.14 $\pm$ 0.27}
& \best{80.12 $\pm$ 0.52}
& \best{83.91 $\pm$ 1.43}
& \best{68.32 $\pm$ 0.18} \\
\bottomrule
\end{tabular}%
\end{table*}

We also analyze sensitivity to the synthetic-to-real mixing ratio $\alpha$. Table~\ref{tab:ablation_alpha_datasetwise} shows that moderate mixing performs best: $\alpha=0.5$ for MM-Fit, UTD-MHAD, and MMAct, and $\alpha=0.25$ for PAMAP2 and HAD-AW. Larger ratios reduce performance across all datasets, indicating that synthetic data is most effective as a controlled supplement rather than a replacement for real target-domain IMU.

\begin{table*}[htbp]
\scriptsize
\centering
\caption{Sensitivity to the synthetic-to-real mixing ratio $\alpha$ with the number of SMP variants fixed to $K=4$. Results are reported as dataset-specific Macro-F1 under LOPO evaluation using DeepConvLSTM. Best and second-best values are highlighted in green and yellow, respectively.}
\label{tab:ablation_alpha_datasetwise}
\begin{tabular}{lccccc}
\toprule
\textbf{Setting} & \textbf{MM-Fit} & \textbf{UTD-MHAD} & \textbf{MMAct} & \textbf{PAMAP2} & \textbf{HAD-AW} \\
\midrule
$\alpha = 0.1$ 
& 79.8 $\pm$ 1.7 
& 56.1 $\pm$ 0.8 
& 77.9 $\pm$ 0.6 
& 83.0 $\pm$ 1.6 
& \second{67.5 $\pm$ 0.2} \\

$\alpha = 0.25$ 
& \second{81.5 $\pm$ 1.6} 
& 58.0 $\pm$ 0.7 
& \second{79.7 $\pm$ 0.6} 
& \best{83.9 $\pm$ 1.4} 
& \best{68.3 $\pm$ 0.2} \\

$\alpha = 0.5$ 
& \best{83.1 $\pm$ 1.6} 
& \best{60.1 $\pm$ 0.6} 
& \best{80.1 $\pm$ 0.5} 
& \second{83.7 $\pm$ 1.4} 
& 68.0 $\pm$ 0.2 \\

$\alpha = 1.0$ 
& 81.3 $\pm$ 1.7 
& \second{58.4 $\pm$ 0.7} 
& 79.4 $\pm$ 0.6 
& 83.2 $\pm$ 1.5 
& 67.2 $\pm$ 0.2 \\

$\alpha = 2.0$ 
& 80.4 $\pm$ 1.8 
& 56.9 $\pm$ 0.8 
& 78.3 $\pm$ 0.7 
& 82.3 $\pm$ 1.8 
& 66.4 $\pm$ 0.3 \\
\bottomrule
\end{tabular}%
\end{table*}

\subsection{Distribution Analysis and Realism}

\begin{figure}[htbp]
    \centering

    \includegraphics[width=1\linewidth]{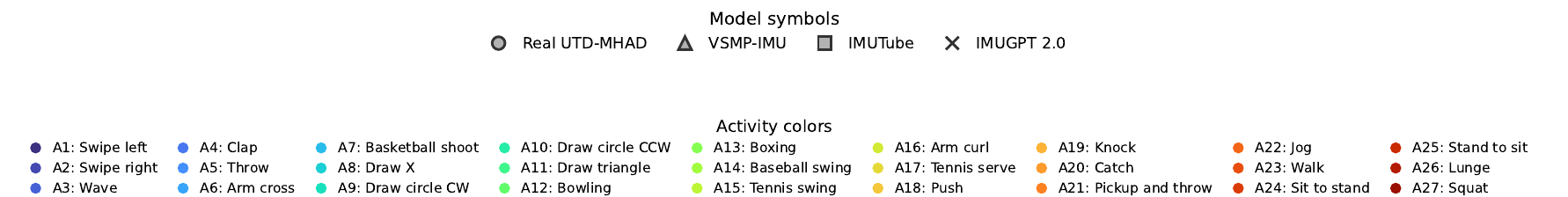}\\
    \includegraphics[width=1\linewidth]{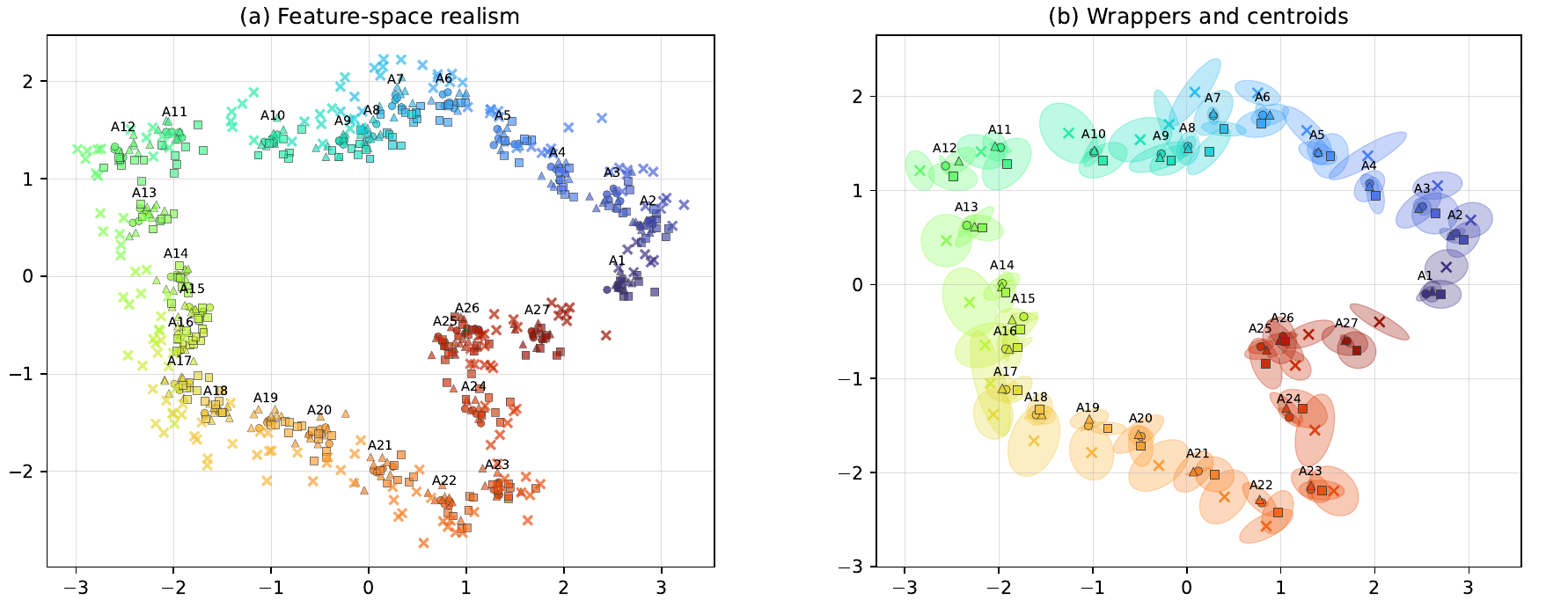}\\[-0.2em]
    \includegraphics[width=0.6\linewidth]{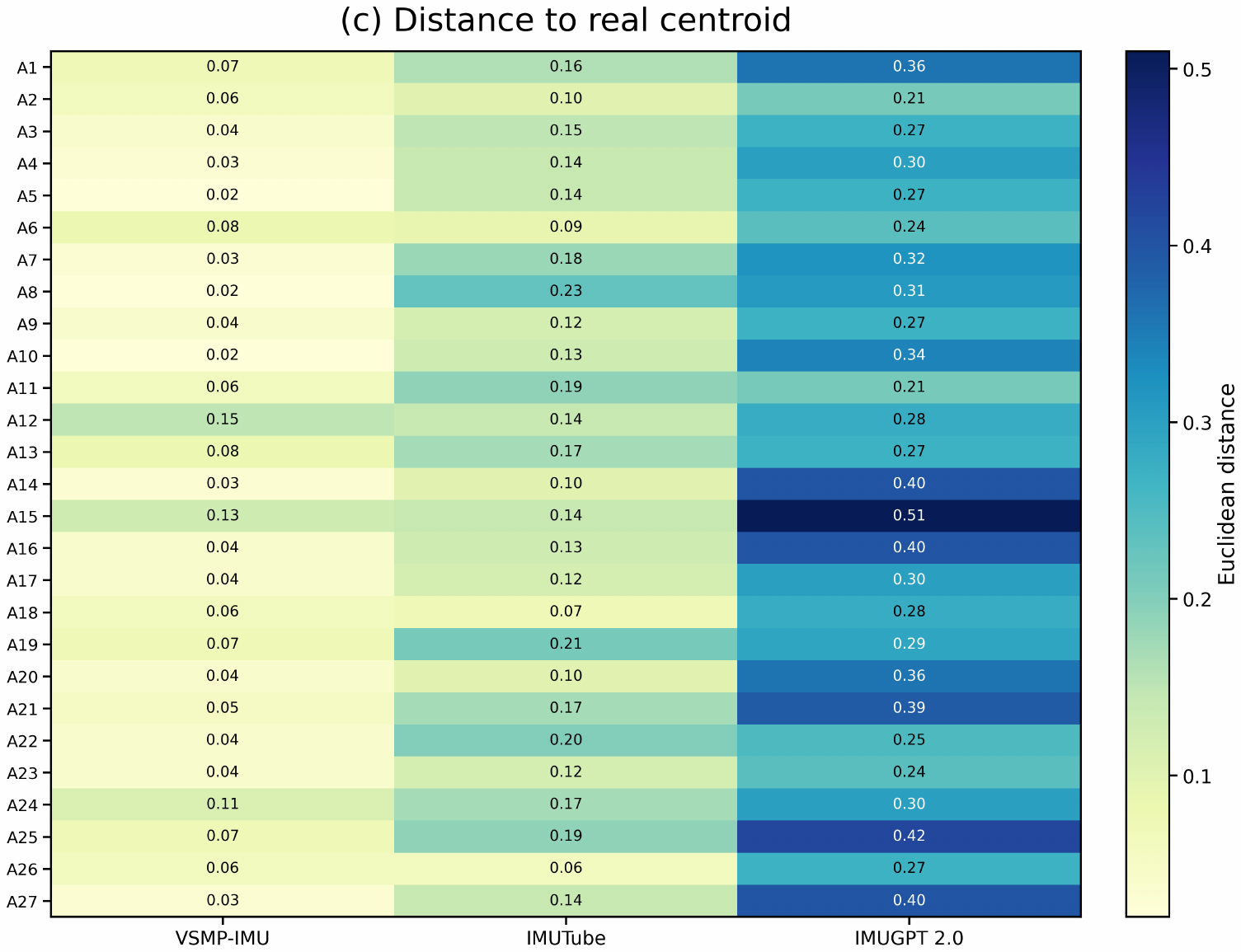}

    \caption{Feature-space realism analysis on the UTD-MHAD dataset. Real UTD-MHAD samples and synthetic samples generated by VSMP-IMU, IMUTube, and IMUGPT~2.0 are visualized in the embedding space of the trained HAR encoder. VSMP-IMU expands intra-class coverage while remaining closer to the corresponding real activity clusters, whereas baseline methods exhibit larger distributional drift. Activity labels A1--A27 correspond to the UTD-MHAD activity classes.}
    \label{fig:feature_space_realism}
\end{figure}

We next assess whether the synthetic sequences generated by VSMP-IMU remain close to the real IMU distribution while still improving downstream recognition. 
Feature-space realism is evaluated using the representation learned by the downstream
DeepConvLSTM encoder as visulaized in Figure~\ref{fig:feature_space_realism}. 
 Specifically, after training DeepConvLSTM on the corresponding real training split, we freeze the encoder and extract the hidden representation from the final convolutional feature layer for each 2-second IMU window.
  Thus, each real or synthetic window $x$ is mapped to an embedding
\begin{equation}
    z = f_{\theta}(x) \in \mathbb{R}^{d},
\end{equation}
where $f_{\theta}$ denotes the trained DeepConvLSTM encoder. For visualization, these high-dimensional embeddings are projected to two dimensions using PCA. The PCA projection is used only for qualitative visualization; all quantitative distance measurements are computed in the original encoder feature space.
To quantify realism, we compute the class-wise distance between the synthetic feature centroid and the real feature centroid. For activity class $a$, let $\mathcal{R}_a$ and $\mathcal{S}_a$ denote the sets of real and synthetic embeddings, respectively. We compute
\begin{equation}
\mu^{\mathrm{real}}_a
=
\frac{1}{|\mathcal{R}_a|}
\sum_{z_i \in \mathcal{R}_a} z_i,
\qquad
\mu^{\mathrm{syn}}_a
=
\frac{1}{|\mathcal{S}_a|}
\sum_{z_j \in \mathcal{S}_a} z_j,
\end{equation}
and report the Euclidean centroid distance
\begin{equation}
d_a
=
\left\|
\mu^{\mathrm{syn}}_a
-
\mu^{\mathrm{real}}_a
\right\|_2 .
\end{equation}

VSMP-IMU produces synthetic samples that expand intra-class coverage while remaining closely aligned with the corresponding real activity clusters. Averaged over all 27 activity classes, VSMP-IMU obtains the lowest mean distance to the real centroid, with an average distance of $0.056$, compared with $0.143$ for IMUTube and $0.313$ for IMUGPT~2.0. This corresponds to a $60.9\%$ reduction relative to IMUTube and an $82.2\%$ reduction relative to IMUGPT~2.0.

These results indicate that VSMP-IMU does not merely increase synthetic sample diversity, but does so while preserving class-consistent structure in the learned IMU feature space. In contrast, the larger centroid distances of IMUTube and IMUGPT~2.0 suggest greater distributional drift from the real IMU manifold, especially for classes where synthetic motion dynamics are harder to align with real inertial patterns.

\section{Discussion and Limitations}

\subsection{Discussion of Design Choices}

VSMP-IMU is designed to preserve activity-defining structure while varying execution- and sensing-level factors across users and deployments. The SMP representation abstracts away frame-level pose noise, provides more control than free-form captions, and introduces semantic, motion-level, and sensor-level diversity beyond direct signal augmentation.

The results suggest that this structured semantic layer is especially useful in low-resource and class-imbalanced settings, where semantically meaningful diversity is more valuable than perturbing existing inertial signals. By separating activity identity from motion instantiation, SMPs preserve core activity structure while allowing controlled changes in tempo, amplitude, repetition count, dominant side, and related execution attributes.

SMP verbalization provides a practical bridge to text-conditioned motion synthesis, retaining semantic control while remaining compatible with existing text-to-motion models. On the sensor side, VSMP-IMU separates simulator-side calibration from target-domain marginal alignment: simulation-in-the-loop optimization adjusts physically meaningful parameters using temporal and multivariate IMU statistics, while rank-based grounding performs final channel-wise distribution matching. This avoids an unconstrained synthetic-to-real mapping while improving compatibility with real wearable IMU data.

\revise{The strictly matched caption--SMP ablation clarifies that the
benefit of VSMP-IMU is not explained only by richer attribute
instructions or a larger retained sample budget. Appending matched
tempo, amplitude, and activity-specific variations to generic captions
improves average Macro-F1 by $1.54$, but the complete SMP still
outperforms this matched caption condition by $3.26$. Structured
conditioning is also beneficial without augmentation, and the positive
representation--augmentation interaction indicates that execution
variations are more useful when constrained by explicit primitives,
temporal organization, body-part involvement, and contact semantics.}

\revise{The controllability study distinguishes explicit control
specification from actual control realization. Tempo and repetition
count exhibit the strongest agreement at the generated-motion level,
reflecting their clear temporal structure. Their agreement decreases
more substantially after motion-to-IMU conversion and grounding.
Amplitude shows slightly weaker motion-level control but retains the
strongest relationship in grounded IMU. Primitive duration remains
measurable but exhibits increasing boundary error across the pipeline,
while symmetry is evaluated only at the motion level because UTD-MHAD
does not provide bilateral wearable observations.}

\revise{These results show that SMP controllability depends on both HY-Motion's
response to the verbalized intervention and the observability of the
resulting motion at the selected wearable location.}
\subsection{Computation--Performance Tradeoff}

VSMP-IMU adds offline generation cost relative to real-only training and classical signal augmentation. The cost comes from SMP extraction, constrained augmentation, motion synthesis, IMU simulation, sensor adaptation, and quality filtering. These steps are used only during synthetic data generation; at test time, the downstream HAR backbone is unchanged, so inference cost is identical to the corresponding real-only model.

On a single A6000 Ada GPU, classical signal augmentation takes less than 0.01 s per synthetic sample, IMUTube-style video-to-IMU generation takes about 2 s, and IMUGPT 2.0-style text-to-motion generation takes about 5 s. VSMP-IMU takes about 10 s per sample because it adds video-grounded SMP extraction, constrained semantic augmentation, simulation-in-the-loop optimization, rank-based grounding, and filtering.

VSMP-IMU has memory use comparable to IMUGPT 2.0 because both are dominated by text-to-motion generation and IMU simulation. Unlike IMUTube-style pipelines, VSMP-IMU does not require dense frame-level pose estimation or direct video-based motion recovery, so its peak memory footprint is expected to be similar to IMUGPT 2.0 and slightly lower than IMUTube under comparable batch-size and clip-duration settings.

Despite the higher offline cost, VSMP-IMU provides the largest performance gain. In standard LOPO, it improves average Macro-F1 by 9.77 over Real-only training, compared with 0.75 for Classical Aug., 5.74 for IMUTube, and 2.93 for IMUGPT 2.0. In low-resource LOPO, the corresponding gains are 18.54, 4.59, 11.78, and 11.00.

Thus, VSMP-IMU is most suitable when offline generation cost is acceptable and labeled IMU collection is expensive. The tradeoff is strongest in low-resource and long-tail regimes, where the gain over cheaper augmentation methods is largest.

\subsection{Limitations}

VSMP-IMU has several limitations.\revise{First, VSMP-IMU depends on the reliability of the VLM used to extract SMPs from video. The extraction audit shows that activity
identity is recovered reliably, but errors remain concentrated in
primitive granularity, fine-grained hand motion, object/contact
semantics, and identification of safely augmentable fields. Aggregation
and semantic refinement improve field-level accuracy and reduce schema
failures, but $15.5\%$ of audited clips retain at least one incorrect
field after refinement.}
\revise{These residual errors may propagate through verbalization into
motion and IMU synthesis even when the final activity label remains
correct. Uncertainty-conditioned augmentation reduces this risk by
restricting uncertain fields, but it cannot eliminate confidently
predicted semantic errors. Future work should incorporate denser
temporal video sampling and explicit hand-pose and object-interaction
models.}

\revise{Second, the UTD-MHAD activity-granularity analysis shows that
VSMP-IMU is not restricted to repetitive exercise motion. Positive
gains are observed for non-periodic sports actions and wrist-dominant
gestures such as swiping, waving, shape drawing, and knocking.
Nevertheless, these gains are smaller than those observed for
repetitive and lower-body activities.}

\revise{The HY-Motion backend represents full-body skeletal motion,
including shoulder, elbow, arm, and wrist trajectories, but does not
explicitly model articulated fingers, grasp type, object approach,
contact force, or object-state transitions. This particularly limits
classes such as right-hand catch and knock, where wearable signals
depend on short contact-induced transients. The current evaluation
should therefore be interpreted as evidence for wrist-level and
non-periodic motion generality within UTD-MHAD, not as validation of
articulated hand-motion synthesis.}

\revise{The present controllability evaluation is restricted to attributes that
can be evaluated meaningfully using UTD-MHAD. Dominant-side switching
is not tested because several UTD-MHAD activity labels explicitly
specify a right arm or right hand, and switching sides could change the
class definition. Global displacement is not tested because the
locomotion actions consist of walking and jogging in place.}

\revise{The controllability study distinguishes explicit control
specification from actual control realization. Tempo and repetition
count exhibit the strongest agreement at the generated-motion level,
reflecting their clear temporal structure. Their agreement decreases
more substantially after motion-to-IMU conversion and grounding.
Amplitude shows slightly weaker motion-level control but retains the
strongest relationship in grounded IMU. Primitive duration remains
measurable but exhibits increasing boundary error across the pipeline,
while symmetry is evaluated only at the motion level because UTD-MHAD
does not provide bilateral wearable observations.}

Finally, VSMP-IMU has higher offline cost than classical augmentation because it requires SMP extraction, text-to-motion synthesis, IMU simulation, and sensor adaptation. This cost is most justified when synthetic data can be reused or labeled IMU collection is expensive, but it may be less attractive for small datasets where simple signal augmentation is sufficient.

\section{Conclusion}

We presented VSMP-IMU, a video-grounded framework for synthetic IMU generation based on Semantic Motion Programs. VSMP-IMU introduces a structured intermediate representation that preserves activity-defining motion semantics while exposing controllable execution variations. \revise{The framework combines video grounding with measurable semantic
control. Tempo and repetition count exhibit the strongest realization
in generated motion, whereas amplitude is comparatively better
preserved in grounded IMU. Control becomes weaker for primitive-level
timing, subtle coordination changes, and attributes that are poorly
observed by the target wearable location.}

Across standard, low-resource, and long-tail LOPO evaluations, VSMP-IMU improves over real-only training, classical signal augmentation, direct video-to-IMU generation, and text-only synthesis. These results show that structured video-grounded semantics can generate more useful synthetic inertial data, especially when labeled IMU data is scarce or class-imbalanced. This suggests that structured multimodal representations and calibrated simulation pipelines are a promising direction for wearable data generation.

\revise{The activity-granularity analysis further shows that the
benefits extend beyond periodic exercises to UTD-MHAD wrist-dominant
gestures, non-periodic sports actions, and posture transitions.
However, gains are weaker for classes whose identity depends on
object contact, hand closure, or short distal-motion transients that
are not explicitly represented by the full-body motion backend.
VSMP-IMU should therefore be viewed as a structured supplement for
body- and wrist-level wearable data rather than a generator of
articulated finger motion.}

\bibliographystyle{ACM-Reference-Format}
\bibliography{refs}

\appendix
\section{Reproducibility Details}
\label{app:reproducibility}

This appendix reports implementation details required to reproduce the VSMP-IMU pipeline, including video preprocessing, VLM-based Semantic Motion Program (SMP) extraction, constrained SMP augmentation, text-to-motion synthesis, virtual IMU simulation, grounding, quality filtering, and downstream HAR training.

\subsection{Video Preprocessing and Frame Sampling}
\label{app:video_preprocessing}

All videos are processed at the clip level. For paired video--IMU datasets, clips are taken only from the training subjects in the current leave-one-person-out (LOPO) fold. The held-out subject's videos and IMU windows are excluded from VLM extraction, SMP augmentation, synthetic IMU generation, grounding, validation, filtering, and model training.

For IMU-only target datasets, source clips are external public videos collected independently of the target IMU dataset. The same retained source-video pool is used for all compared video-based synthetic-generation methods. Upon publication, we will provide the retained source-video URL list, filtering metadata, and dataset-to-source mapping to facilitate reproducibility, subject to platform availability and licensing constraints.

\begin{table}[h]
\centering
\scriptsize
\caption{Dataset-specific video preprocessing parameters for VLM-based SMP extraction.}
\label{tab:dataset_frame_sampling}
\begin{tabular}{lcccl}
\toprule
Dataset & Sampling strategy & $N_f$ & Resolution & Clip duration rule \\
\midrule
MM-Fit & Uniform temporal sampling & 24 & $512 \times 512$ & 2--12 s \\
UTD-MHAD & Uniform temporal sampling & 16 & $512 \times 512$ & 1--8 s \\
MMAct & Uniform temporal sampling & 32 & $512 \times 512$ & 2--17 s \\
PAMAP2 source videos & Uniform temporal sampling & 32 & $512 \times 512$ & 3--15 s \\
HAD-AW source videos & Uniform temporal sampling & 32 & $512 \times 512$ & 3--12 s \\
\bottomrule
\end{tabular}
\end{table}

Each video clip is sampled into a fixed set of frames before VLM processing. We use:
\begin{itemize}
    \item frame sampling strategy: uniform temporal sampling over the trimmed activity clip;
    \item number of sampled frames per clip: dataset-specific, as reported in Table~\ref{tab:dataset_frame_sampling};
    \item input frame resolution: $512 \times 512$;
    \item Clip duration: between 1 to 17 seconds depending on dataset and activity type.
\end{itemize}

If a clip contains fewer frames than $N_f$, frames are repeated using nearest-neighbor temporal padding. If a clip contains more frames than $N_f$, frames are uniformly sampled over the full clip duration. We do not use test-subject video frames in any fold.

\subsection{VLM-Based SMP Extraction}
\label{app:vlm_prompt}

We use GPT-5.3 as the video-capable vision-language model (VLM) for SMP extraction. The model is queried five times per source clip using the sampled frame sequence and a structured prompt. The VLM output is required to follow a JSON-like schema so that it can be parsed deterministically.

The prompt template is:

\begin{quote}
You are given a video clip of one human activity. Extract a Semantic Motion Program (SMP) that describes the activity for wearable IMU synthesis.

Return only valid JSON matching the schema below. Do not include extra commentary.

Identify:
(1) the activity label,
(2) the ordered motion primitives,
(3) primitive durations or relative phase lengths if visible,
(4) involved body parts,
(5) temporal structure, including repetition count, periodicity, ordering, and transitions,
(6) execution attributes, including tempo, amplitude, dominant side, symmetry, and global displacement,
(7) object or contact information,
(8) invariant fields that must not change during augmentation,
(9) variant fields that may change while preserving the activity label,
(10) field-wise uncertainty scores in $[0,1]$, where $0$ means high confidence and $1$ means maximum uncertainty.

Use conservative uncertainty values when the video is occluded, ambiguous, cropped, or contains multiple possible activities.
\end{quote}

The required JSON schema is:

\begin{verbatim}
{
  "activity_label": "string",

  "motion_primitives": [
    {
      "name": "string",
      "order_index": "integer",
      "description": "string",
      "body_parts": ["string"],
      "relative_duration": "float_or_unknown",
      "contact": ["string"]
    }
  ],

  "primitive_sequence": ["string"],

  "primitive_durations": {
    "type": "absolute|relative|unknown",
    "values": ["float_or_unknown"]
  },

  "body_parts": {
    "mandatory": ["string"],
    "optional": ["string"],
    "dominant_moving_segments": ["string"]
  },

  "temporal_structure": {
    "is_repetitive": "boolean",
    "repetition_count": "integer_or_unknown",
    "periodicity": "none|single|cyclic|unknown",
    "ordering": "ordered|weakly_ordered|unordered",
    "transition_type": "discrete|continuous|unknown"
  },

  "execution_attributes": {
    "tempo": "slow|medium|fast|unknown",
    "amplitude": "small|medium|large|unknown",
    "dominant_side": "left|right|bilateral|none|unknown",
    "symmetry": "symmetric|asymmetric|unknown",
    "global_displacement": "stationary|small|large|unknown"
  },

  "object_contact": {
    "objects": ["string"],
    "body_contact": ["string"],
    "environment_contact": ["string"],
    "required_contact": ["string"]
  },

  "constraints": {
    "invariant_fields": [
      "activity_label",
      "primitive_sequence",
      "mandatory_body_parts",
      "required_contact",
      "stationary_or_mobile"
    ],
    "variant_fields": [
      "tempo",
      "amplitude",
      "repetition_count",
      "primitive_duration",
      "dominant_side",
      "symmetry",
      "global_displacement"
    ],
    "rules": {
      "preserve_activity_label": true,
      "preserve_core_primitive_order": true,
      "preserve_mandatory_body_parts": true,
      "preserve_required_contact": true,
      "dominant_side_swap_allowed_only_if_side_agnostic": true,
      "symmetry_change_allowed_only_if_label_preserving": true,
      "global_displacement_change_must_preserve_stationary_or_mobile_semantics": true
    }
  },

  "uncertainty": {
    "activity_label": "float",
    "motion_primitives": "float",
    "primitive_duration": "float",
    "body_parts": "float",
    "temporal_structure": "float",
    "repetition_count": "float",
    "tempo": "float",
    "amplitude": "float",
    "dominant_side": "float",
    "symmetry": "float",
    "global_displacement": "float",
    "object_contact": "float"
  }
}
\end{verbatim}

We query the VLM five times per clip to reduce occasional schema-level parsing failures and API-side nondeterminism in video-conditioned responses. The final SMP is obtained by deterministic aggregation across the five parsed outputs. For categorical fields, we use majority voting. For ordered primitive sequences, we select the sequence with the highest average pairwise agreement. For numeric or ordinal fields such as repetition count, tempo, amplitude, and primitive duration, we use the median valid value. Field-wise uncertainty is computed as the mean reported uncertainty, increased when the five outputs disagree. If fewer than three valid JSON outputs are obtained, the clip is discarded from synthetic generation.

\paragraph{SMP audit annotation guide.}
\revise{Annotators viewed each complete source clip at its original
frame rate and could replay the clip without restriction. They recorded
the activity label using the target dataset taxonomy and decomposed the
activity into the smallest visually distinguishable ordered primitives.
Preparation and return phases were included only when they produced a
distinct body configuration or transition.}

\revise{Body-part annotations used the set
$\{$head, torso, left/right shoulder, left/right upper arm,
left/right forearm, left/right hand, pelvis, left/right thigh,
left/right lower leg, left/right foot$\}$. Tempo and amplitude were
annotated as low, medium, or high relative to typical executions of
the same activity class. Dominant side was annotated as left, right,
bilateral, or not applicable. Symmetry was annotated as symmetric,
asymmetric, or not applicable. Global displacement was annotated as
stationary, bounded local displacement, or travelling.}

\revise{Object/contact annotations included manipulated objects,
support surfaces, body--environment contact, and persistent
body--object contact. Invariant fields were defined as properties whose
modification would change the activity label, while variant fields were
properties that could change within the activity class without altering
its identity. Annotators completed their reference SMPs independently
before adjudication.}

\paragraph{SMP audit metrics.}
\revise{For a predicted primitive sequence $\widehat{P}$ and reference
sequence $P$, normalized edit similarity is}
\begin{equation}
\operatorname{Sim}_{P}
=
1-
\frac{
d_{\mathrm{edit}}(\widehat{P},P)
}{
\max(|\widehat{P}|,|P|)
}.
\end{equation}

\revise{Body-part, object/contact, invariant, and variant annotations
are evaluated as sets using Macro-F1. Repetition error is computed as}
\begin{equation}
\operatorname{MAE}_{\mathrm{rep}}
=
\frac{1}{N}
\sum_{i=1}^{N}
\left|
\widehat{r}_{i}-r_i
\right|.
\end{equation}

\revise{A clip is counted as having a residual refined-SMP error when
at least one non-unknown field differs from the adjudicated reference
after deterministic repair and semantic consistency checking.
Response-level schema errors are computed over all five VLM queries,
whereas aggregation, correction, discard, and residual-error rates are
computed once per source clip.}

\paragraph{Matched caption--SMP prompt construction.}
\revise{For source clip $V_i$, let $C_i$ denote its generic video
caption, $S_i$ its refined base SMP, and $S_i^{(k)}$ its constrained
variant for generation slot $k$. Let $A_i^{(k)}$ denote the
natural-language phrase expressing the requested execution attributes.
The four ablation prompts are}
\begin{align}
Z_{i,k}^{\mathrm{C0}} &= C_i, \\
Z_{i,k}^{\mathrm{C1}} &= C_i \oplus A_i^{(k)}, \\
Z_{i,k}^{\mathrm{S0}} &= T_{\mathrm{HY}}(S_i), \\
Z_{i,k}^{\mathrm{S1}} &= T_{\mathrm{HY}}(S_i^{(k)}).
\end{align}

\revise{The operator $\oplus$ appends the attribute instruction without
changing the original caption. For C1 and S1, $A_i^{(k)}$ is copied
verbatim between conditions. For example, both may receive the phrase
``at a fast pace, using a large range of motion, and repeating the
complete movement eight times.'' S1 additionally retains ordered
primitives, body-part involvement, temporal organization,
object/contact information, and activity-preserving constraints.}

\subsection{SMP Constraint Rules}
\label{app:smp_constraints}

SMP refinement applies four consistency checks before augmentation:
\begin{enumerate}
    \item \textbf{Label--primitive compatibility:} the primitive sequence must be plausible for the inferred activity label.
    \item \textbf{Body-part consistency:} the body parts listed in the SMP must match the motion primitives.
    \item \textbf{Temporal coherence:} primitive ordering, repetition count, periodicity, and transition type must be mutually consistent.
    \item \textbf{Object/contact consistency:} required object and environment contacts must not contradict the activity semantics.
\end{enumerate}

Fields are divided into invariant and variant subsets. The following fields are treated as invariant:
\begin{itemize}
    \item activity label;
    \item core primitive sequence;
    \item mandatory body-part set;
    \item required object/contact pattern;
    \item stationary-vs-mobile semantic category, while allowing bounded changes to displacement extent within that category.
\end{itemize}

The following fields may be augmented if allowed by the activity constraints:
\begin{itemize}
    \item tempo;
    \item amplitude;
    \item repetition count;
    \item primitive durations;
    \item dominant side;
    \item symmetry;
    \item global displacement extent.
\end{itemize}

For continuous or ordinal fields $f$, the perturbation radius is uncertainty-conditioned:
\[
\Delta_f(u_f) = (1-u_f)\Delta_f^{\max},
\]
where $u_f \in [0,1]$ is the VLM-estimated uncertainty. The augmented value is sampled from:
\[
f' \sim \mathrm{Uniform}(f-\Delta_f(u_f), f+\Delta_f(u_f)),
\]
with clipping to the valid range of field $f$.

For categorical fields, the probability of changing the field is:
\[
P(f' \neq f) = (1-u_f)\rho_f,
\]
where $\rho_f$ is the maximum allowed perturbation probability.

The perturbation ranges used in our experiments are:

\begin{table}[h]
\centering
\scriptsize
\caption{SMP perturbation ranges used for semantic augmentation.}
\label{tab:smp_perturbation_ranges}
\begin{tabular}{lll}
\toprule
Field & Range / probability & Constraint \\
\midrule
Tempo & $\Delta_{\mathrm{tempo}}^{\max} = 1$ level & Preserve primitive order and label \\
Amplitude & $\Delta_{\mathrm{amp}}^{\max} = 1$ level & Preserve motion type \\
Repetition count & $\Delta_{\mathrm{rep}}^{\max} = 5$ cycles & Repetitive activities only \\
Primitive duration & $\Delta_{\mathrm{dur}}^{\max} = 25\%$ & Preserve ordering and total coherence \\
Dominant side & $\rho_{\mathrm{side}} = 0.5$ & Side-agnostic activities only \\
Symmetry & $\rho_{\mathrm{sym}} = 0.3$ & Preserve class identity \\
Global displacement extent & $\Delta_{\mathrm{disp}}^{\max} = 25\%$ & Preserve stationary/mobile semantics \\
\bottomrule
\end{tabular}
\end{table}

\subsection{Simulation-in-the-Loop Optimization}
\label{app:simulation_loop}

Simulation-in-the-loop optimization adjusts bounded residual simulator parameters using only training-fold statistics. The residual parameter vector $\eta$ includes small mounting-orientation offsets, gravity-convention choices, bias corrections, and noise-scale corrections. It is not implemented as an unconstrained neural translator.

The optimization objective is:
\[
\mathcal{L}_{\mathrm{sim}}
=
\lambda_{\Sigma}\|\Sigma(X_{\mathrm{sim}})-\Sigma(D_{\mathrm{stat}})\|_F
+
\lambda_{\mathrm{psd}}\|\mathrm{PSD}(X_{\mathrm{sim}})-\mathrm{PSD}(D_{\mathrm{stat}})\|_2
+
\lambda_r\|r(X_{\mathrm{sim}})-r(D_{\mathrm{stat}})\|_2.
\]

This loss intentionally does not match per-channel empirical cumulative distributions, because per-channel marginal alignment is handled separately by the rank-based grounding step.

We use:
\begin{itemize}
    \item optimizer: \texttt{bounded cross-entropy method};
    \item number of optimization iterations: $30$;
    \item candidate samples per iteration: $64$;
    \item elite fraction: $0.20$;
    \item initial orientation-search standard deviation: $5^{\circ}$;
    \item initial accelerometer-bias search standard deviation: $0.25~\mathrm{m/s^2}$;
    \item initial gyroscope-bias search standard deviation: $0.02~\mathrm{rad/s}$;
    \item initial noise-scale search standard deviation: $0.10$;
    \item $\lambda_{\Sigma} = 1.0$;
    \item $\lambda_{\mathrm{psd}} = 0.5$;
    \item $\lambda_r = 0.5$;
    \item covariance statistic window: $2$ seconds;
    \item PSD frequency bands: $[0,0.5]$, $(0.5,1]$, $(1,2]$, $(2,4]$, $(4,8]$, and $(8,f_s/2]$ Hz;
    \item autocorrelation lags: $\{1,2,5,10,20,50\}$ samples after resampling to the dataset-specific HAR input rate $f_s$.
\end{itemize}

The bounded residual search ranges are:

\begin{table}[h]
\centering
\scriptsize
\caption{Residual simulator parameters optimized during simulation-in-the-loop calibration.}
\label{tab:residual_sim_params}
\begin{tabular}{ll}
\toprule
Parameter & Search range \\
\midrule
Roll offset & $[-10^{\circ}, 10^{\circ}]$ \\
Pitch offset & $[-10^{\circ}, 10^{\circ}]$ \\
Yaw offset & $[-15^{\circ}, 15^{\circ}]$ \\
Accelerometer bias & $[-0.5, 0.5]~\mathrm{m/s^2}$ per axis \\
Gyroscope bias & $[-0.05, 0.05]~\mathrm{rad/s}$ per axis \\
Accelerometer noise scale & $[0.5, 2.0]$ times nominal noise \\
Gyroscope noise scale & $[0.5, 2.0]$ times nominal noise \\
Gravity convention & $\{\mathrm{gravity\ included}, \mathrm{linear\ acceleration}\}$ \\
\bottomrule
\end{tabular}
\end{table}

All statistics in $D_{\mathrm{stat}}$ are estimated only from real IMU windows belonging to non-test subjects in the current LOPO fold. For each fold, $D_{\mathrm{stat}}$ and $D_{\mathrm{rank}}$ are constructed from disjoint windows sampled from the non-test subjects. We use a stratified split of the training-fold real IMU windows, assigning $50\%$ to $D_{\mathrm{stat}}$ and $50\%$ to $D_{\mathrm{rank}}$ within each activity class whenever enough windows are available. Classes with too few windows are split as evenly as possible. Held-out subject data is never used in either set.

\subsection{Rank-Based Sensor Grounding}
\label{app:rank_grounding}

After simulation-in-the-loop optimization, we apply sensor-frame alignment and rank-based marginal grounding. The grounding prior is:
\[
\gamma = \{j_{\mathrm{target}}, R_{\mathrm{target}}, P_{\mathrm{axis}}, c_{\mathrm{sign}}, g_{\mathrm{conv}}\}.
\]

The prior $p(\gamma)$ is:

\begin{table}[h]
\centering
\scriptsize
\caption{Grounding prior $p(\gamma)$ used for target-domain sensor alignment.}
\label{tab:grounding_prior}
\begin{tabular}{ll}
\toprule
Parameter & Prior / setting \\
\midrule
Target segment $j_{\mathrm{target}}$ &
Dataset-specific wearable location from the target dataset metadata \\
Target mounting transform $R_{\mathrm{target}}$ &
Nominal dataset mounting frame with bounded Euler perturbation \\
Mounting perturbation for wrist-worn sensors &
Roll/pitch/yaw sampled from $[-30^{\circ},30^{\circ}]$ \\
Mounting perturbation for non-wrist sensors &
Roll/pitch sampled from $[-15^{\circ},15^{\circ}]$, yaw from $[-30^{\circ},30^{\circ}]$ \\
Axis permutation $P_{\mathrm{axis}}$ &
All $6$ valid permutations of $(x,y,z)$ considered \\
Sign convention $c_{\mathrm{sign}}$ &
All $8$ sign-flip combinations in $\{-1,+1\}^{3}$ considered \\
Gravity convention $g_{\mathrm{conv}}$ &
$\{\mathrm{gravity\ included}, \mathrm{linear\ acceleration}\}$ \\
\bottomrule
\end{tabular}
\end{table}

For multi-sensor datasets, grounding is applied separately for each recorded wearable location. The target segment $j_{\mathrm{target}}$ is selected from the dataset-provided sensor placement metadata. For smartwatch-only datasets, $j_{\mathrm{target}}$ is set to the wrist segment. For IMU-only datasets with multiple sensor positions, synthetic IMU channels are generated and grounded separately for each target position before being concatenated in the same channel order as the real dataset.

Before rank transformation, the convention-aligned signal is:
\[
X_{\mathrm{align}} = A_{\gamma}(X_{\mathrm{opt}}),
\]
where $A_{\gamma}(\cdot)$ applies the target segment selection, mounting transform, axis permutation, sign convention, and gravity convention.

For each channel $c$, rank grounding maps a convention-aligned synthetic value $x_v$ to the real-domain value with the same empirical percentile:
\[
x_r = G^{-1}_{r,c}(F_{v,c}(x_v)).
\]

For a six-channel IMU sequence, this is applied independently to:
\[
c \in \{a_x,a_y,a_z,\omega_x,\omega_y,\omega_z\}.
\]

The empirical real-domain distributions $G_{r,c}$ are estimated only from $D_{\mathrm{rank}}$, which contains real IMU training windows from non-test subjects in the current LOPO fold. The empirical synthetic distributions $F_{v,c}$ are estimated from the convention-aligned synthetic samples generated for the same fold and target sensor location. Held-out subject data is never used for grounding.

This ordering keeps the two adaptation stages separate: simulation-in-the-loop optimization matches temporal and multivariate statistics using $D_{\mathrm{stat}}$, while rank-based grounding performs the final non-parametric per-channel marginal alignment using $D_{\mathrm{rank}}$.

\subsection{Quality Filtering}
\label{app:quality_filter}

Synthetic samples are filtered using three criteria: motion plausibility, class faithfulness, and redundancy. All thresholds are selected using only real IMU windows from non-test subjects in the current LOPO fold.

\paragraph{Motion plausibility.}
A rule-based filter rejects samples with invalid skeleton frames, missing body segments, degenerate motion, abrupt discontinuities, or inertial magnitudes outside physically plausible training-fold ranges. For each activity class $y$, we compute robust thresholds from the real training windows of that class. If a class has fewer than $100$ training windows, we fall back to thresholds computed from all real training windows in the fold.

Let $D_{\mathrm{train}}^{(y)}$ denote real training windows for class $y$. We use the following thresholds:

\begin{itemize}
    \item maximum acceleration magnitude:
    \[
    a_{\max}^{(y)}
    =
    \min \left(
    6g,\;
    Q_{0.999}\left(\|a\|_2;D_{\mathrm{train}}^{(y)}\right)
    +
    3\,\mathrm{MAD}\left(\|a\|_2;D_{\mathrm{train}}^{(y)}\right)
    \right),
    \]
    where $g=9.81~\mathrm{m/s^2}$, giving an absolute cap of $58.86~\mathrm{m/s^2}$.

    \item maximum angular velocity magnitude:
    \[
    \omega_{\max}^{(y)}
    =
    \min \left(
    35~\mathrm{rad/s},\;
    Q_{0.999}\left(\|\omega\|_2;D_{\mathrm{train}}^{(y)}\right)
    +
    3\,\mathrm{MAD}\left(\|\omega\|_2;D_{\mathrm{train}}^{(y)}\right)
    \right).
    \]

    \item maximum frame-to-frame joint displacement:
    \[
    d_{\max}=0.20h,
    \]
    where $h$ is the estimated body height of the generated skeleton. A motion is rejected if any joint moves more than $0.20h$ between adjacent generated frames.

    \item minimum motion energy:
    \[
    E_{\min}^{(y)}
    =
    \max\left(
    10^{-4},\;
    Q_{0.01}\left(E(X);D_{\mathrm{train}}^{(y)}\right)
    \right),
    \]
    where
    \[
    E(X)=
    \frac{1}{T}
    \sum_{t=1}^{T}
    \left(
    \|a_t-\bar{a}\|_2^2
    +
    0.1\|\omega_t-\bar{\omega}\|_2^2
    \right).
    \]
    This rejects nearly static or degenerate generated windows.

    \item allowed missing-frame fraction: $0\%$.
\end{itemize}

A generated sample is rejected by the motion-plausibility filter if:
\[
\max_t \|a_t\|_2 > a_{\max}^{(y)}
\quad \mathrm{or} \quad
\max_t \|\omega_t\|_2 > \omega_{\max}^{(y)}
\quad \mathrm{or} \quad
\max_{t,j} \|p_{t,j}-p_{t-1,j}\|_2 > d_{\max}
\quad \mathrm{or} \quad
E(X) < E_{\min}^{(y)}.
\]

\paragraph{Class faithfulness.}
A lightweight HAR validator is trained only on real IMU windows from the current training fold. This validator is separate from the final downstream HAR model. A synthetic sample is retained only if:
\[
\max_c p(c|X_{\mathrm{imu}}) \geq \tau_{\mathrm{cls}}
\quad \mathrm{and} \quad
\arg\max_c p(c|X_{\mathrm{imu}})=y.
\]

We use:
\[
\tau_{\mathrm{cls}} = 0.60.
\]

The threshold is selected from the candidate set
\[
\{0.50,0.55,0.60,0.65,0.70\}
\]
using validation Macro-F1 on the validation split within the training subjects. The selected value is fixed before evaluation on the held-out subject.

\begin{table*}[t]
\centering
\scriptsize
\caption{\textbf{Generation accounting for the strictly matched
caption--SMP ablation.}
A matched block contains one motion from each condition
(C0, C1, S0, and S1), generated using the same source clip,
generation slot, and HY-Motion seed. A block is accepted only when
all four motions pass the common quality filters. Counts are reported
per LOPO fold as the mean and standard deviation where applicable.}
\label{tab:matched_generation_accounting}
\resizebox{\textwidth}{!}{
\begin{tabular}{lrrrrrrr}
\toprule
\revise{Dataset} &
\revise{Classes} &
\revise{Source clips/fold} &
\revise{Target accepted blocks} &
\revise{Attempted blocks} &
\revise{Block acceptance} &
\revise{Retained motions/condition} &
\revise{Total retained motions} \\
\midrule

MM-Fit
& 10
& 50
& 200
& $279 \pm 12$
& 71.7\%
& 200
& 800 \\

UTD-MHAD
& 27
& 135
& 540
& $791 \pm 25$
& 68.3\%
& 540
& 2,160 \\

MMAct
& 35
& 175
& 700
& $1{,}094 \pm 38$
& 64.0\%
& 700
& 2,800 \\

PAMAP2
& 18
& 90
& 360
& $489 \pm 18$
& 73.6\%
& 360
& 1,440 \\

HAD-AW
& 31
& 155
& 620
& $1{,}005 \pm 44$
& 61.7\%
& 620
& 2,480 \\

\bottomrule
\end{tabular}}
\end{table*}

\revise{The retained-motion counts in
Table~\ref{tab:matched_generation_accounting} are distinct from the
number of fixed-duration windows supplied to the HAR model. Each
retained motion is converted into virtual IMU and segmented into
2-second windows. Because sequence durations vary, a motion may produce
multiple windows. After segmentation, we sample an identical number of
synthetic windows per activity class from C0, C1, S0, and S1 according
to the dataset-specific synthetic-to-real ratio. Thus, all four
conditions enter downstream training with identical window counts and
class distributions.}

\begin{table}[t]
\centering
\scriptsize
\caption{\textbf{Independent filter acceptance before matched-block
retention.} Final downstream counts remain identical across
conditions.}
\label{tab:matched_acceptance}
\begin{tabular}{lc}
\toprule
\revise{Condition} & \revise{Independent acceptance} \\
\midrule
C0: Generic caption & 77.8\% \\
C1: Caption + matched attributes & 82.4\% \\
S0: Unaugmented refined SMP & 87.6\% \\
S1: Constrained augmented SMP & 90.9\% \\
Complete matched block & 66.2\% \\
\bottomrule
\end{tabular}
\end{table}

\paragraph{Redundancy.}
The redundancy filter uses the penultimate-layer feature representation $\phi(\cdot)$ of the HAR validator. Features are $\ell_2$-normalized before nearest-neighbor comparison. A generated sample is retained only if its nearest-neighbor distance from already retained synthetic samples of the same class exceeds:
\[
\tau_{\mathrm{nn}}^{(y)}
=
Q_{0.05}
\left(
\min_{X_j \in D_{\mathrm{train}}^{(y)},\, j \neq i}
\|\phi(X_i)-\phi(X_j)\|_2
\right).
\]

Thus, $\tau_{\mathrm{nn}}^{(y)}$ is the $5$th percentile of within-class nearest-neighbor distances among real training windows. This avoids using an absolute feature-space threshold whose value would depend on the scale of the validator representation.

A generated sample is rejected as redundant if:
\[
\min_{X' \in \mathcal{X}_{\mathrm{synth}}^{(y)}}
\|\phi(X_{\mathrm{imu}})-\phi(X')\|_2
\leq
\tau_{\mathrm{nn}}^{(y)}.
\]

The nearest-neighbor threshold is computed using only real training windows from non-test subjects and is fixed before testing.

The final quality decision is:
\[
Q(X_{\mathrm{imu}},M,S)
=
Q_{\mathrm{motion}}(M,X_{\mathrm{imu}})
\cdot
Q_{\mathrm{class}}(X_{\mathrm{imu}},y)
\cdot
Q_{\mathrm{red}}(X_{\mathrm{imu}}).
\]
Only samples for which $Q=1$ are retained.

\subsection{Qualitative Analysis}

\begin{figure*}[htbp]
    \centering
    \includegraphics[width=\textwidth]{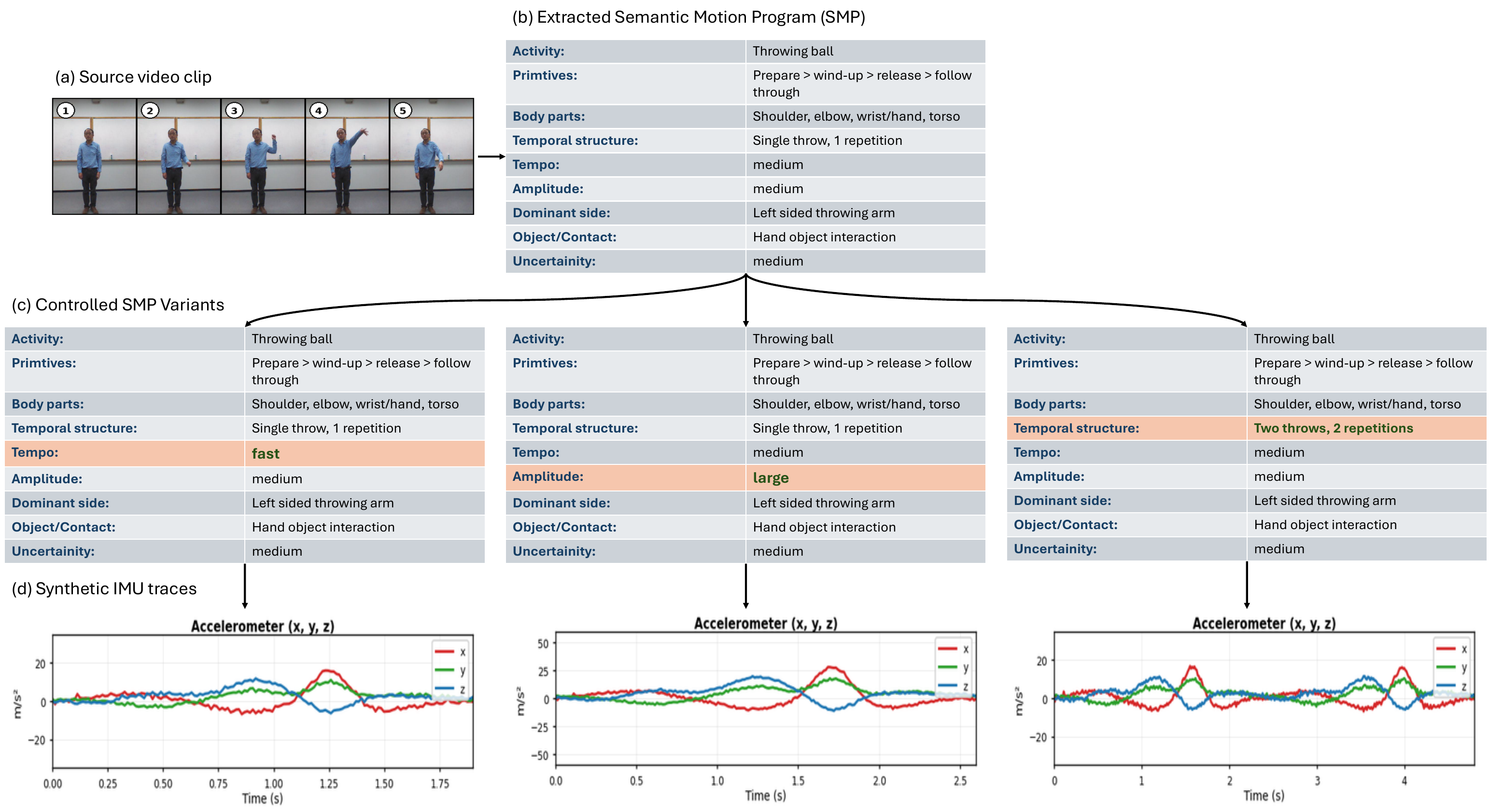}
    \caption{\textbf{Detailed case study for a single activity.} Starting from a source video clip, VSMP-IMU extracts an SMP and generates three controlled variants by modifying selected execution attributes while preserving activity identity and core primitive structure. The figure shows the source video frames, the extracted SMP, the three augmented variants, and the corresponding synthetic IMU traces. This qualitative case study complements the quantitative end-to-end
controllability analysis in Section~\ref{sec:control_results}.}
    \label{fig:case_study}
\end{figure*}

Figure~\ref{fig:case_study} illustrates representative examples from the proposed pipeline. Starting from an input video, VSMP-IMU extracts a SMP that summarizes activity structure in terms of motion primitives, temporal organization, body-part involvement, execution attributes, and object/contact cues. The extracted program is then augmented into multiple activity-preserving variants, which in turn produce diverse motion realizations and corresponding IMU traces. Compared with free-form captioning, the SMP provides finer control over which aspects of the activity remain fixed and which are allowed to vary.

\subsection{ECDF Feature Extraction for Random Forest}
\label{app:ecdf_rf}

The Random Forest (RF) backbone follows a feature-based HAR pipeline rather than operating directly on raw temporal IMU windows. For consistency with the neural backbones, we first segment both real and synthetic IMU sequences into the same fixed-duration windows used by the deep models. Specifically, each sample corresponds to a N-second IMU window. Let
\begin{equation}
X \in \mathbb{R}^{T \times C}    
\end{equation}

denote one window, where $T$ is the number of timesteps in the window and $C$ is the number of IMU channels.

For each channel $c \in \{1,\dots,C\}$, we compute the empirical cumulative distribution function (ECDF) over the values observed within the window:
\begin{equation}
\hat{F}_{c}(v)
=
\frac{1}{T}
\sum_{t=1}^{T}
\mathbf{1}(X_{t,c} \leq v),    
\end{equation}

where $\mathbf{1}(\cdot)$ is the indicator function. The ECDF summarizes the distribution of signal values in the window while discarding explicit temporal ordering. This makes it suitable for RF, which expects fixed-length feature vectors rather than sequential inputs.

To obtain a compact descriptor, we sample each channel-wise ECDF at $K=15$ fixed probability levels. Equivalently, this corresponds to extracting 15 quantile-style distributional features from each sensor channel. The resulting ECDF descriptor for channel $c$ is
\begin{equation}
\phi_c(X)
=
\left[
q_c(p_1), q_c(p_2), \dots, q_c(p_K)
\right],    
\end{equation}

where $p_k$ denotes the $k$-th fixed ECDF level and $q_c(p_k)$ is the corresponding empirical quantile of channel $c$ within the window. Features from all channels are concatenated to form the RF input vector:
\begin{equation}
\phi(X)
=
\left[
\phi_1(X), \phi_2(X), \dots, \phi_C(X)
\right]
\in \mathbb{R}^{KC}.    
\end{equation}

\section{Analysis Parameters}

\begin{table*}[t]
\centering
\scriptsize
\caption{\textbf{Activity-specific definitions used for controllability
measurement.} Principal joints are used for motion-level measurements.
For IMU-level measurements, the channel with the highest periodic or
transient energy is selected from the accelerometer and gyroscope
channels at the UTD-MHAD sensor location. Symmetry is evaluated only
at the motion level.}
\label{tab:activity_control_definitions}
\resizebox{\textwidth}{!}{
\begin{tabular}{lllll}
\toprule
\revise{Activity} &
\revise{Principal joints} &
\revise{Cycle or primitive definition} &
\revise{IMU location} &
\revise{Applicable controls} \\
\midrule

Right-hand wave &
right wrist, elbow, shoulder &
one complete lateral sweep and return &
right-wrist&
tempo, amplitude, repetitions \\

Two-hand front clap &
left/right wrists, elbows, shoulders &
hands separate $\rightarrow$ contact $\rightarrow$ separate &
right-wrist&
tempo, amplitude, repetitions, symmetry \\

Right-arm throw &
right wrist, elbow, shoulder, trunk &
preparation $\rightarrow$ forward acceleration $\rightarrow$ follow-through &
right-wrist&
tempo, amplitude, primitive duration \\

Basketball shoot &
wrists, elbows, shoulders, knees &
loading $\rightarrow$ upward extension $\rightarrow$ follow-through &
right-wrist&
tempo, amplitude, primitive duration \\

Right-hand draw circle&
right wrist, elbow, shoulder &
one complete closed clockwise hand trajectory &
right-wrist&
tempo, amplitude, repetitions \\

Front boxing &
left/right wrists, elbows, shoulders &
one alternating left--right punch pair &
right-wrist&
tempo, amplitude, repetitions \\

Two-arm curl &
left/right wrists, elbows, shoulders &
extension $\rightarrow$ bilateral flexion $\rightarrow$ extension &
right-wrist&
tempo, amplitude, repetitions, symmetry \\

Two-hand push &
left/right wrists, elbows, shoulders, trunk &
retracted $\rightarrow$ forward extension $\rightarrow$ retracted &
right-wrist&
tempo, amplitude, primitive duration, symmetry \\

Jogging in place &
hips, knees, ankles &
one left--right step pair &
right-thigh&
tempo, amplitude, repetitions \\

Walking in place &
hips, knees, ankles &
one left--right step pair &
right-thigh&
tempo, amplitude, repetitions \\

Forward lunge &
hips, knees, ankles, pelvis &
standing $\rightarrow$ forward descent/hold $\rightarrow$ return &
right-thigh&
tempo, amplitude, primitive duration \\

Extended squat&
hips, knees, ankles, shoulders &
standing $\rightarrow$ descent $\rightarrow$ lowest point $\rightarrow$ standing &
right-thigh&
tempo, amplitude, repetitions, symmetry \\

\bottomrule
\end{tabular}}
\end{table*}

\paragraph{Control-measurement parameters.}
\revise{Before cycle detection, motion and IMU trajectories are filtered using
a fourth-order zero-phase Butterworth low-pass filter with a cutoff of
$6$~Hz. Peaks are detected using a minimum prominence of
$0.40$ times the signal standard deviation and a minimum
distance of $0.35$~s. The minimum distance is selected to
remain below the shortest valid cycle duration across the slow,
medium, and fast conditions while preventing multiple detections within
a single movement cycle.}

\revise{The symmetry threshold is selected using only the motion-validation
split and is fixed before evaluating the test intervention sequences.
A sequence is classified as symmetric when its mean mirrored
left--right trajectory correlation is at least
$C_{\mathrm{sym}}=0.82$. The validation split contains
three pilot activities and two additional HY-Motion
generation seeds that are disjoint from the 12 activities and five
seeds used for the reported controllability results. The pilot
activities are bowling, tennis serve, and sit-to-stand, which provide
unilateral, coordinated, and lower-body motion patterns, respectively.}

\revise{Primitive boundaries are identified from activity-specific joint-angle
and velocity landmarks. A primitive starts when the relevant
joint-velocity magnitude exceeds $20\%$ of its sequence
maximum and ends when it falls below the same threshold after the
principal motion peak. To suppress short threshold crossings caused by
trajectory noise, the condition must remain satisfied for at least
$0.10$~s. The same filtering, peak-detection, symmetry, and
boundary-detection parameters are used for all intervention levels of
an activity.}

\end{document}